\documentclass[format=acmsmall, review=false, timestamp=false]{acmart}
\setcitestyle{numbers,sort&compress}
\usepackage{booktabs} 
\usepackage[ruled]{algorithm2e} 

\usepackage{amsmath}
\usepackage{array}
\usepackage{fixltx2e}
\usepackage{url}
\usepackage{hyperref}
\usepackage{enumitem}
\usepackage{subcaption}
\usepackage{pdflscape} 
\usepackage{multirow}
\usepackage{multicol}
\usepackage[flushleft,referable]{threeparttablex}
\usepackage{color}
\usepackage{tablefootnote}
\usepackage{enumitem} 
\usepackage{todonotes}
\usepackage[normalem]{ulem}
\usepackage{gensymb}

\newcommand{\ie}{{\it i.e.},~}
\newcommand{\eg}{{\it e.g.},~}
\newcommand{\etal}{{\it et al.}~}

\acmJournal{CSUR}
\acmVolume{0}
\acmNumber{0}
\acmArticle{0}
\acmYear{2018}
\acmMonth{0}

\setcopyright{acmcopyright}

\acmDOI{0000001.0000001}

\begin{document}
\title{Presentation Attack Detection for Iris Recognition: An Assessment of the State of the Art}
\titlenote{pre-print accepted for publication in ACM Computing Surveys on June 13, 2018}

\author{Adam Czajka}
\orcid{0000-0003-2379-2533}
\affiliation{%
  \institution{Research and Academic Computer Network (NASK)}
  \streetaddress{Kolska 12}
  \city{Warsaw}
  \postcode{01045}
  \country{Poland}}
\email{adam.czajka@nask.pl}
\author{Kevin W. Bowyer}
\orcid{0000-0002-7562-4390}
\affiliation{%
  \institution{University of Notre Dame}
  \streetaddress{384 Fitzpatrick Hall}
  \city{Notre Dame}
   \state{IN}
  \postcode{46556}
  \country{USA}}
\email{kwb@nd.edu}

\renewcommand\shortauthors{Czajka A. and Bowyer K. W.}

\begin{abstract}
Iris recognition is increasingly used in large-scale applications. As a result, presentation attack detection for iris recognition takes on fundamental importance. This survey covers the diverse research literature on this topic. Different categories of presentation attack are described and placed in an application-relevant framework, and the state of the art in detecting each category of attack is summarized. One conclusion from this is that presentation attack detection for iris recognition is not yet a solved problem. Datasets available for research are described, research directions for the near- and medium-term future {are outlined}, and a short list of recommended readings are suggested.
\end{abstract}

%
%

%
%

\maketitle

\section{Introduction}
\label{sec:Introduction}

{The} term {\em presentation attack} refers to making a presentation to the sensor with the goal of manipulating the system into an incorrect decision. The term {\em spoofing} is a related less formal term, and {\em liveness detection} can be considered as one of the countermeasures to detect a presentation attack. Commonly envisioned goals of a presentation attack are to impersonate a targeted identity or to evade recognition.

As iris recognition has become increasingly popular, {\em presentation attack detection} (PAD) has received substantial attention. Various early publications in this area reported near-perfect accuracy. But it is now recognized that these early efforts addressed idealized versions of the problem, especially when we analyze the results of independent iris PAD competitions presented in Section \ref{sec:Competitions} and the number of various attack instruments and ideas presented in Section \ref{sec:KnownVulnerabilities}, and possible instruments used in iris presentation attacks discussed in Section \ref{sec:PAI}. More recent works attempt to address {\em open set} versions of the problem. {T}hese more realistic works  (\eg {\cite{Sequeira_TSP_2016,Yambay_IJCB_2017,Raja_BIOSIG_2016,Sollinger_IET_2017}}) report accuracy figures lower than in the earlier works. However, we should expect that accuracy on the more realistic versions of the problem will improve as research progresses, and the amount of PAD-related data offered now by researchers is stunning, as summarized in Section \ref{sec:BenchmarkDatabases}.

It can be difficult to conceptually organize and evaluate the different technologies employed in presentation attacks and their detection. Therefore, in Section \ref{sec:PADmethods}, we present an organizing framework based on two main salient distinctions. PAD methods for iris recognition can be either {\it static} or {\it dynamic}, and also either {\it passive} or {\it active}.
A static method operates on a single sample, whereas a dynamic method operates on an image sequence to extract features related to dynamics of the observed object. A passive method makes the measurement without any stimulation beyond the normal visible-light or near-infrared illumination used to acquire an iris image, whereas an active method adds some additional element of stimulation to the eye / iris. This categorization is meant to clearly reflect the complexity of the sensor and of the image acquisition process. 

With this framework, relatively under-studied areas become easier to identify. Also, within each category, we can identify the current state-of-the-art. And it becomes easier to assess the value of currently available datasets to support the different categories of research. As a result, we suggest in Section \ref{sec:Conclusions} possible avenues for future research efforts, including datasets and algorithms.

We start this survey with basic terminology (Sec. \ref{sec:Terminology}) and make comments on visible-light vs near-infrared illumination in iris recognition, which has an influence on the PAD methodology (Sec. \ref{sec:Near_infrared}). Later in Section \ref{sec:EvaluationAndPerformance} we discuss also the evaluation of PAD, which is in principle different than evaluation of biometric recognition. Lastly, in Section \ref{sec:SuggestedReading} we provide a short list of ``recommended readings'' for those wanting to start a deeper dive into this area.

\section{Terminology}
\label{sec:Terminology}

The iris PAD literature has historically been  inconsistent in use of terminology. In this survey, we attempt to follow normative presentation attack vocabulary wherever appropriate, as recommended in {ISO/IEC 30107-3:2017}. 

Biometric characteristics (possibly a non-living sample) or artificial objects used in presentation attack are called {\bf presentation attack instruments (PAI)}. A presentation to the biometric sensor is either a {\em bona fide} presentation or an {\em attack} presentation. The following terms are used for basic error metrics:
\begin{itemize}
\item {\bf Attack Presentation Classification Error Rate (APCER):} proportion of {\it attack presentations} incorrectly classified as {\it bona fide presentations}.

\item {\bf Bona Fide Presentation Classification Error Rate (BPCER):} proportion of {\it bona fide presentations} incorrectly classified as {\it presentation attacks}.

{\item {\bf Correct Classification Rate (CCR):} sum of correctly classified {\it bona fide presentations} and correctly classified {\it presentation attacks} divided by the number of all presentations.}

\end{itemize}

{APCER and BPCER are functions of a decision threshold $\tau$.} When the threshold $\tau$ can be set so that APCER($\tau$) = BPCER($\tau$), then the equal error rate {EER = APCER($\tau$) = BPCER($\tau$) can be reported.}

The goal of an attacker is most often envisioned as either (a) impersonating some targeted identity, or (b) avoiding a match to the attacker's true identity. In this context, the following terms are used for the attacker's success rate:
\begin{itemize} 
\item {\bf Impostor Attack Presentation Match Rate (IAPMR):} proportion of impostor attack presentations that are successful; that is, in which the biometric reference for the targeted identity is matched. This error metric is analogous to false match rate (FMR) in identity verification.
\item {\bf Concealer Attack Presentation Non-Match Rate (CAPNMR):} proportion of concealer attack presentations that are successful; that is, in which the biometric reference of the concealer is not matched. This error metric is analogous to false non-match rate (FNMR) in identity verification.
\end{itemize}

{A}n attacker's intentions may be more nuanced than is envisioned by the current standard terminology. For example, {attackers} may seek to enroll multiple identities that do not match any existing person, but that they can match successfully to in the future. This could be seen as a more sophisticated form of concealer attack. The standard terminology may expand and evolve in the future to explicitly include such instances.

Meanings of additional acronyms, generally associated with texture features, used throughout the paper are:
BSIF: Binary Statistical Image Features \cite{Kannala_ICPR_2012},
CNN: Convolutional Neural Network \cite{Lecun_ProcIEEE_1998},
LBP: Local Binary Patterns \cite{Ojala_ICPR_1994},
SID: Shift-Invariant Descriptor \cite{Kokkinos_CVPR_2008}, and
SVM: Support Vector Machine \cite{Boser_CLT_1992}.

\section{Near-Infrared and Visible-Light Iris Recognition}
\label{sec:Near_infrared}

Essentially all commercial iris recognition systems operate using near-infrared illumination of the eye. This has been true since the early work by Daugman \cite{Daugman_patent_1994}. When using visible-light wavelengths, the melanin will absorb a significant amount of light, and the eyes with high concentration of melanin will appear ``dark'', sometimes making the pupil localization process difficult or impossible. Using near-infrared illumination, the texture of the iris surface can be imaged approximately equally well for all persons.

Iris recognition's reputation as a highly-accurate biometric {method} is thus established in the context of using near-infrared illumination. {According to the most recent IREX IX report \cite{IREX_IX}, the best-performing one-to-one iris matchers achieve a false non-match rate below one percent for a false match rate of $10^{-5}$ (1 in 100,000).}

Iris recognition using visible-light illumination has attracted attention in the past, \eg the UBIRIS effort \cite{UBIRIS_DB_URL} and more recently in the context of mobile devices {\cite{Sequeira_VISAPPb_2014,Trokielewicz_JTIT_2016}}. If the same accuracy could be achieved with ambient visible-light as with near-infrared illumination, it would be a major advance, enabling lower-cost operation and significant increases in flexibility. However, there is no evidence that visible-light iris recognition can, in similar conditions of use, achieve accuracy close to that of near-infrared iris recognition. General-purpose estimates of visible-light iris error rates from experimental data are problematic because they vary based on many factors, including whether the subjects in the study have ``dark'' or ``light'' eyes. {Also, comparing near-infrared iris images with visible-light iris images (cross-spectrum matching) is more challenging than comparing iris images illuminated by the same light (same-spectrum).} Nevertheless, visible-light iris recognition may find use in lower-security applications, perhaps enabled by ubiquitous use of mobile devices. Therefore, this survey covers research on PAD for visible-light iris as well as for near-infrared iris. {Interested readers can study the results of recent ``Cross-eyed 2017'' competition \cite{Sequeira_IJCB_2017}, and pursue their their own research using a cross-spectrum iris image database \cite{Sequeira_VISAPPb_2014}.}

\section{Known and Potential Vulnerabilities}
\label{sec:KnownVulnerabilities}

\subsection{Attack Goals: Impersonation vs Identity Concealment}
\label{sec:KnownVulnerabilities_Impersonation_vs_Concealment}

The goal of an ``Impostor Attack Presentation'' in the standard terminology is to impersonate some targeted identity. {T}his attack goal requires that the attacker gain access to an iris image, the enrolled iris code, or equivalent information for the targeted identity. However, there is a variant of this attack in which the attacker may simply want to match to any enrolled identity without caring which one. This can be the case in ``token-less'' biometric applications in which a probe sample is acquired and matched against all enrolled identities to identify the user. Matching any enrolled identity, rather than a specific targeted identity, is enough to authorize access. A real-world example where this attack could be relevant is the previous version of the NEXUS system operated jointly by the Canada Border Services Agency and the U.S. Customs and Border Protection\footnote{https://www.cbsa-asfc.gc.ca/prog/nexus/menu-eng.html; last accessed March 21, 2018}.

Another possible goal is the ``Concealer Attack Presentation''. This requires only that the attacker have some means of obscuring the useful texture information in their probe sample. For simpler iris recognition systems, even something as simple as eye drops to cause extreme pupil dilation could enable this type of attack.

A hybrid type of possible attack is that the attacker seeks to enroll an identity that does not correspond to any real person, and then use that identity in the future. This may be realized by generating and enrolling a synthetic iris pattern. {A} simpler approach may be to acquire an iris image with the sensor rotated upside-down. Since not all iris sensors detect the correct orientation, an upside-down image could generate a synthetic identity separate from that corresponding to the right-side-up sample \cite{Czajka_TIFS_2017}. This attack goal is not explicitly considered in the standard terminology.

\subsection{Creating Images to Match Iris Templates}
\label{sec:KnownVulnerabilities_Samples_Matching}

To accomplish an impersonation attack, one must present an image that, after processing by a biometric system, results in a match to a targeted iris template. An obvious approach is to take a photograph of one's eye that can be later printed and presented to the sensor. Due to the prevalence of near-infrared illumination in commercial systems, it is expected that, on average, samples acquired in near-infrared should have higher chances to support a successful presentation attack than visible-light images. However, especially for ``dark'' irises, the red channel of a visible-light iris image may result in a sample that shows enough iris texture to perform a successful print attack. 

There is a small body of work studying attacks in the context of the attacker not having an iris image of the targeted identity, but having the ability to compare a candidate probe image to the enrollment of the targeted identity and get a measure of the match quality. In principle, one can generate a large number of synthetic images that end up with a good match. However, these images may not look visually similar to the targeted iris, or even not similar to a human iris in general. The possibility of ``reversing'' an iris template is clear when we analyze this problem from an information theory point of view. Assume that a standard iris image has a resolution of $640\times480$ pixels and the gray levels are coded by 8 bits. The total number of images possible to be coded equals to $2^{640\times480\times8}=2^{1200\times2048}$. Assume that a typical iris code is composed of 2048 bits, hence the total number of different iris codes is $2^{2048}$. There are no formal limitations to have an iris code calculated for each possible grayscale image. Since $2^{1200\times2048} \gg 2^{2048}$, and there must exist a grayscale image for each iris code, the number of images (possibly not iris images) ending up with an identical binary code is $2^{1199\times2048} \gg 1$. The number of images that end up with a different iris code, but one close enough to generate a match, is even greater. Certainly, additional textural limitation must be added if the generated images should be visually similar to an iris. {Additionally, Rathgeb and Busch \cite{Rathgeb_IJCB_2017} showed how to generate a single iris code that will match with iris codes calculated for more than one distinct iris. Such morphed iris codes may be a starting point for preparation of synthetic iris images that, when presented to a sensor, would match more than one identity.}

Rathgeb and Uhl \cite{Rathgeb_ICPR_2010} present results for a hill-climbing approach to creat{e} an iris image that can be used to make a false match to a given iris enrollment. This method operates on a $512\times64$ normalized iris image, and assumes that an attempted match to an enrollment returns an indication of match quality rather than a match / non-match result. The method scans the initial {fake} image and adjusts the pixel value by a positive or negative increment in order to find a modification that improves the match quality. It iteratively scans and modifies the {fake} image in this hill-climbing manner to synthesize an image that matches the target enrollment. In the best case, about 1,400 iterations over the {fake} image are needed to obtain an acceptable match. Galbally \etal \cite{Galbally_CVIU_2013} use genetic algorithms to find synthetic irises that match their authentic counterparts. The authors suppressed block artifacts and applied Gaussian smoothing to give the synthetic samples a realistic appearance. Also, Venugopalan and Savvides \cite{Venugopalan_TIFS_2011} propose to blend a synthetic image based on someone's iris code with the image of a different subject's iris. This operation modifies an image in the frequency range used in matching, and leaves it almost unchanged in other frequency ranges. {Drozdowski \etal \cite{Drozdowski_BIOSIG_2017} propose a method to generate synthetic iris codes that have similar statistical properties as iris codes generated for authentic irises.}

\subsection{Creating Images of Artificial Identities}
\label{sec:KnownVulnerabilities_Samples_NonMatching}

There are also papers proposing either an iris image synthesis, or alteration of the authentic iris image to generate a new texture, possibly not matching to any existing identity. These techniques cannot be used directly in impersonation attack. However, they can be applied to produce fake irises indiscernible from living irises by a biometric sensor in identity concealment attacks, and in situations where having iris artifacts that resemble real irises is important. One should be careful with taking a claim of high visual realism of such samples for granted{.} {Also, visual realism} is less important than the assessment of ``authenticity'' done by an iris recognition sensor.

The first proposal we are aware of to render a synthetic iris texture (actually the entire eyeball) is by Lefohn \etal in 2003 \cite{Lefohn_CGA_2003}. They claim that their method ``can create patterns and colors that match existing human irises.'' They  follow the approach of composing the artificial eye with multiple, simple layers added incrementally, on top of the already-added layers, to end up with the desired pattern. Shah and Ross \cite{Shah_ICIP_2006} proposed to use Markov Random Fields to generate a background iris texture, and then iris-specific features such as crypts, radial and concentric furrows and collarette are embedded into the background. Zuo \etal \cite{Zuo_TIFS_2007} propose a model-based method of synthesizing iris textures. They start with generation of 3D fibers in {a} cylindrical coordinate system, which are then projected onto a plane to simulate a frontal view of an iris meshwork. The resulting 2D image is then blended with irregular edges, the collarette portion is brightened, the outer boundary between the ``iris'' and the ``sclera'' regions is blurred, and, finally, artificial eyelids and eyelashes are added. More recently, Thavalengal \etal \cite{Thavalengal_IJCB_2014} describe a means to alter an iris portion of a{n} {image} without destroying the photo-realistic features of the eye region. They propose a few simple techniques such as a) vertical flip of the iris portion, b) blurring of the iris texture in radial directions, c) swapping of iris texture sectors, and d) replacement of the entire iris portion with a real or synthetic image.

\section{Presentation Attack Instruments}
\label{sec:PAI} 

\subsection{Artifacts}
\label{sec:KnownVulnerabilities_Artifacts}

\begin{figure}[!htb]
    \centering
    \begin{subfigure}[t]{0.485\textwidth}
        \includegraphics[width=0.475\textwidth]{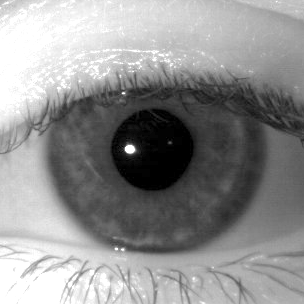}\hskip3mm
        \includegraphics[width=0.475\textwidth]{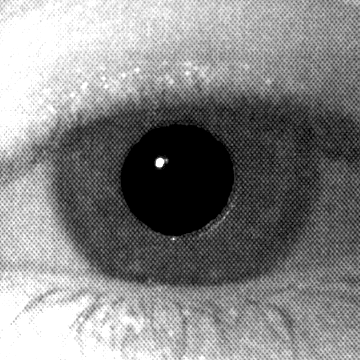}
        \caption{Illustration of a {\bf print attack}. An authentic iris (left) and the corresponding iris printout (right). Samples from a training partition of the {\sf LivDet-Iris Warsaw 2017} dataset; file IDs: 0319\_REAL\_L\_14 and 0319\_PRNT\_L\_1, correspondingly.}
    \end{subfigure}\hfill
    \begin{subfigure}[t]{0.484\textwidth}
        \includegraphics[width=0.475\textwidth]{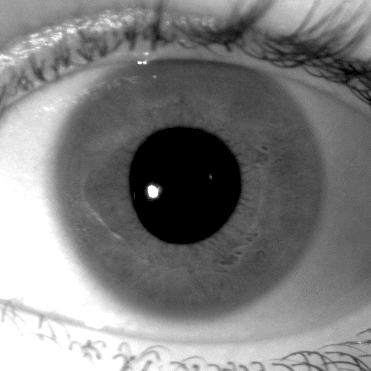}\hskip3mm
        \includegraphics[width=0.475\textwidth]{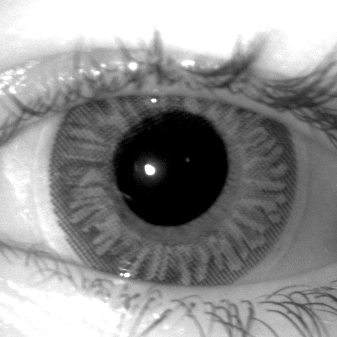}
        \caption{Illustration of a {\bf textured contact lens attack}. An authentic iris (left) and the same eye wearing textured contact lens (right). Notre Dame file IDs: 07013d5451 and 07013d5343, correspondingly.}
    \end{subfigure}
    \caption{Illustration of two presentation attacks that were reported as successful in spoofing commercial sensors in the past for with the purpose of impersonation (a) and recognition evasion (b).}
    \label{fig:samples:print_and_lens}    
\end{figure}  

\begin{figure*}[!htb]
    \centering
    \begin{subfigure}[t]{0.21\textwidth}
        \includegraphics[width=\textwidth]{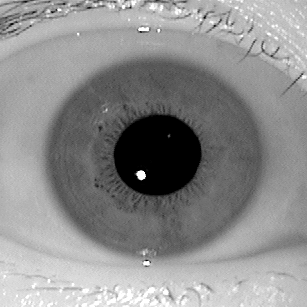}
        \caption{An authentic iris image; Notre Dame file ID: 04202d1496.}
    \end{subfigure}\hfill
    \begin{subfigure}[t]{0.255\textwidth}    
        \includegraphics[width=\textwidth]{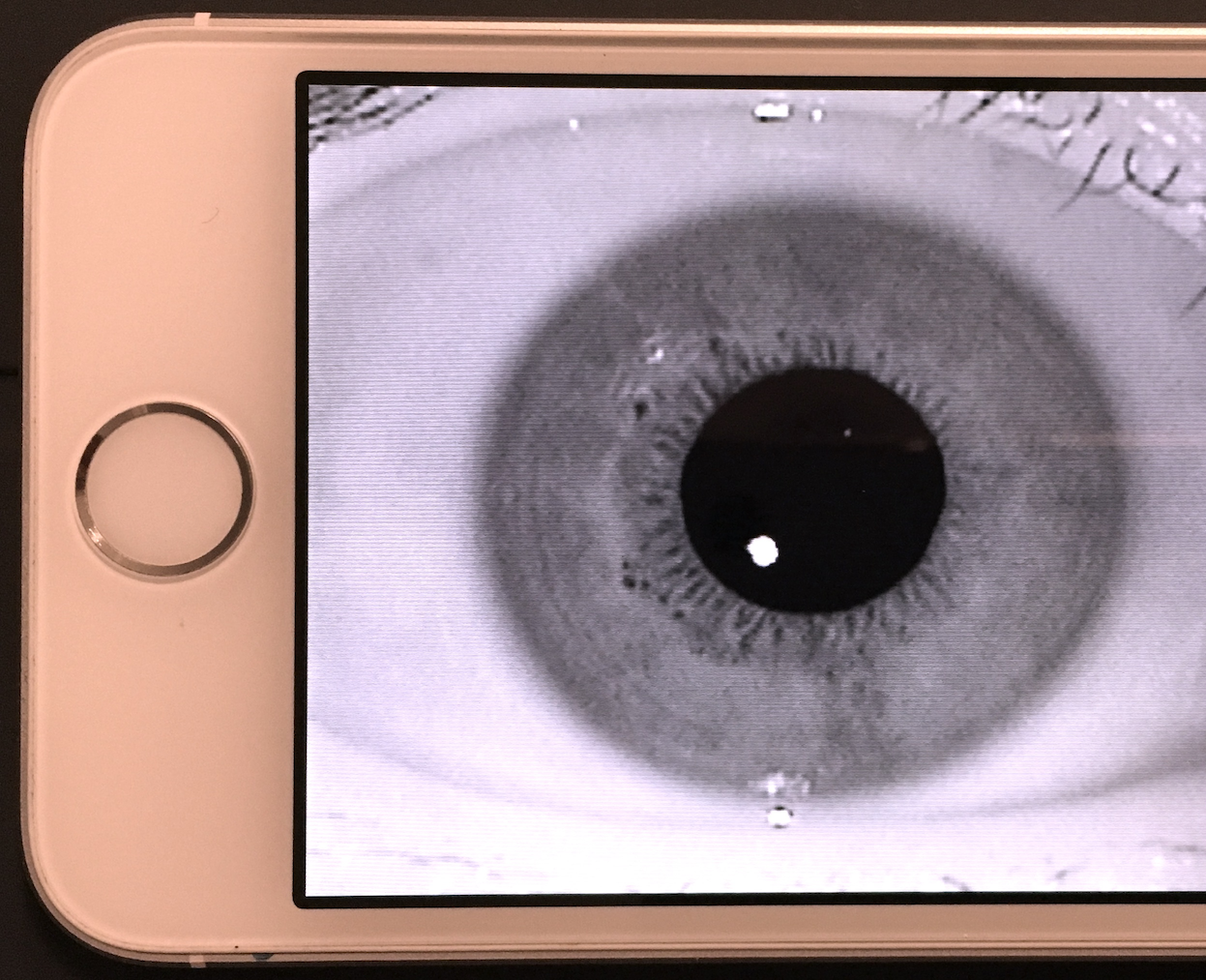}
        \caption{An iris image shown in (a) displayed on the iPhone and photographed by the iPad.}
    \end{subfigure}\hfill
    \begin{subfigure}[t]{0.26\textwidth}  
        \includegraphics[width=\textwidth]{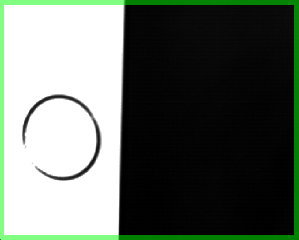}
        \caption{An iris image shown in (a) displayed on the iPhone as seen by the AD100 iris sensor.}
    \end{subfigure}\hfill
    \begin{subfigure}[t]{0.21\textwidth}
        \includegraphics[width=\textwidth]{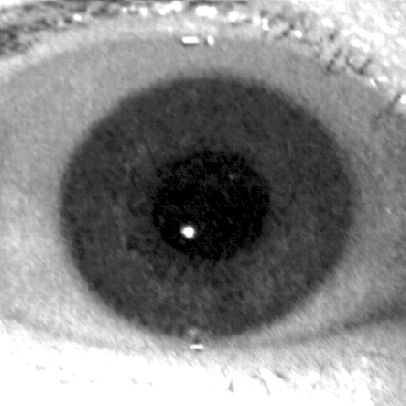}
        \caption{An iris image shown in (a) displayed on the Kindle E-reader and photographed by the AD100 iris sensor.}
    \end{subfigure}
    \caption{An illustration of a {\bf display attack}. Only passive displays, such as Kindle, have a potential to be used in successful presentation attacks directed to commercial equipment.}
    \label{fig:samples:display}
\end{figure*}

\begin{figure*}[!htb]
    \centering
    \begin{subfigure}[t]{0.485\textwidth}
        \includegraphics[width=0.475\textwidth]{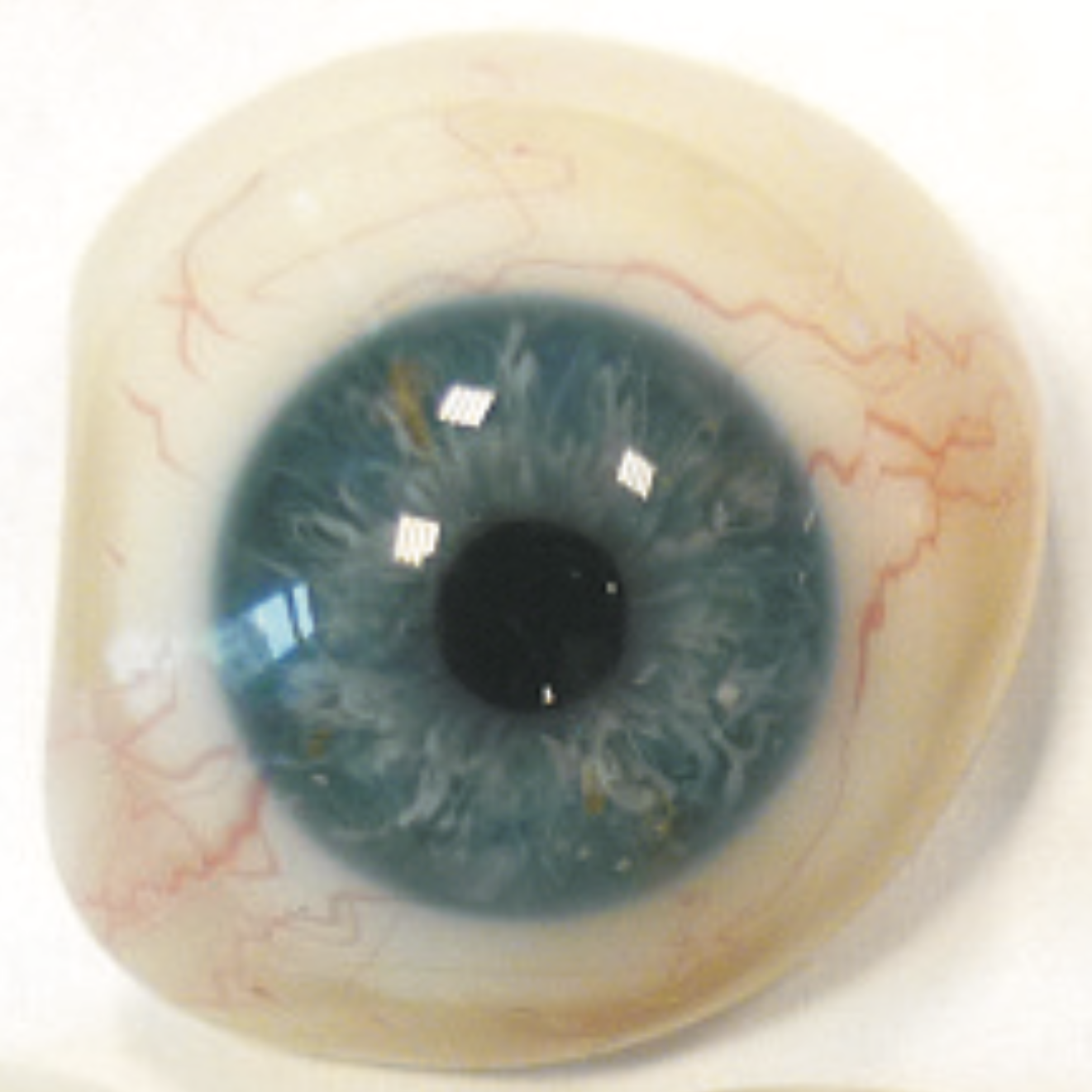}\hskip3mm
        \includegraphics[width=0.475\textwidth]{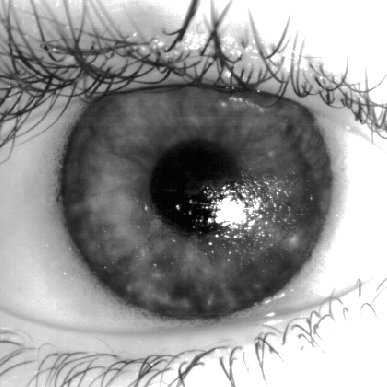}
        \caption{Left: {\bf glassy prosthesis}; source: A. Czajka, {\it Biometrics} course, Univ. of Notre Dame, Fall 2014. Right: {\bf glassy prosthesis} placed in the eye socket and photographed by the AD100 iris recognition sensor; Notre Dame file ID: 06117d493.}
    \end{subfigure}\hfill
    \begin{subfigure}[t]{0.235\textwidth}
        \includegraphics[width=\textwidth]{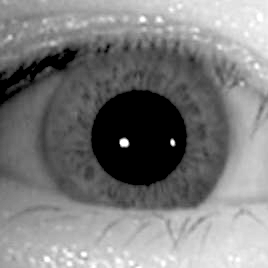}
        \caption{An iris image with an embedded {\bf synthetic iris} texture. Sample taken from  {\sf CASIA-Iris-Syn V4}; file ID: S6002S05.}
    \end{subfigure}\hfill
    \begin{subfigure}[t]{0.235\textwidth}
        \includegraphics[width=\textwidth]{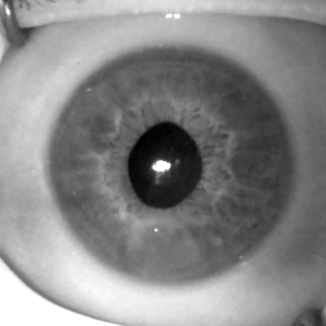}
        \caption{{\bf Cadaver iris} acquired 5 hours after death. Sample taken from  {\sf Post-Mortem-Iris v1.0}; file ID: 0004\_L\_1\_3.}
    \end{subfigure}\hfill
    \caption{An illustration of artifacts and non-living eye proposed in the literature in the context of attacks on iris recognition systems.}
    \label{fig:samples:other}
\end{figure*}

\subsubsection{Attack Technology: Paper Printouts}

Printouts can be produced in many ways. There is no consensus on whether  color or black\&white printing is significantly better, or whether matte or glossy paper is better, to make a successful {Presentation Attack} instrument. However, it is typically easier to get commercial sensors to generate a sample from such {an} artifact when a hole is cut in the place where the pupil is printed, in order to produce specular reflections from an authentic cornea hidden behind the printout when taking a picture, as shown in Fig. \ref{fig:samples:print_and_lens}a.

The earliest demonstration of a successful print attack that we are aware of is in 2002 by Thalheim \etal \cite{Thalheim_CT_2002}. They used an inkjet printer {and} matte paper{,} with {print} resolution of $2400\times1200$ dpi, and cut a hole in the place of the pupil. This artifact was presented to the Panasonic Authenticam BM-ET100 with PrivateID software by Iridian by an attacker hidden behind the printout. The attacker was able to (a) get a correct match between the printout and the reference calculated for an authentic eye, and (b) enroll a printed iris texture, thus ending up with a system that granted access to anyone possessing the enrolled printout. Similar experiments were repeated by Pacut and Czajka in 2006 \cite{Pacut_ICCST_2006} with a different sensor. In their experiments, irises printed in {color on a glossy paper were  matched to the corresponding reference generated for bona fide presentation in 15.6\% and 86.7\% cases by Panasonic BM-ET300 and BM-ET100, correspondingly.} This first statistical evaluation of presentation attacks with COTS equipment showed that the detection mechanisms implemented in commercial sensors of that time were insufficient.

Ruiz-Albacete \etal \cite{Ruiz-Albacete_LNCS_2008} considered two types of attack: (1) {fake (printed)} images used to attempt enrollment and verification, and (2) an original image is used for enrollment, and then a {fake} image used to attempt verification against the enrollment. Both types of attack may potentially be relevant, but the second type is more commonly studied. They use a modification of the Masek iris matching software \cite{MASEK_SOFTWARE_URL} for the experiments. Using a matching threshold that represents 0.1\% FMR and 17\% FNMR on the original images, they find a success rate of 34\% for attack type 1 and 37\% for attack type 2. This success rate is computed using the 72\% of {fake} images that were correctly segmented, and the matching threshold of 0.1\% FMR is higher than normal operation for iris matching.

The above studies show that {print attacks do not match targeted enrollments as well as bona fide presentations do.} But these print attacks were possible. Presenting an iris printed on paper is the simplest way today to attempt impersonation, and hence any non-zero success rate is alarming. However, more than a decade has passed since the above experiments, and many PAD methods have been proposed since then.

\subsubsection{Attack Technology: Textured Contact Lenses}

The term ``textured contact lenses'' refers to contact lenses that are manufactured to have a visual texture to them; see Fig.  \ref{fig:samples:print_and_lens}b. There are also lenses that are ``colored'' in the sense of being clear but tinted with a certain color, and having no visual texture. ``Clear'' contact lenses are neither tinted nor have the visible texture. The term ``cosmetic contact lenses'' is also sometimes used, as the coloring and texturing is for cosmetic (appearance) effect and not for vision correction. The basic problem is that the texture in the contact lens partially overlays the natural iris texture, and hence the image of an iris wearing a textured contact is a mix of contact lens texture and natural iris texture. Additionally, the contact lens moves on the surface of the eye, so that the exact mixture is different from image to image.

It was clear in Daugman's early work that an iris wearing a textured contact lens would generate an iris code that did not match the code from the iris without the lens \cite{Daugman_WMIP_2003}. More recently, Baker \etal \cite{Baker_BTAS_2009, Baker_CVIU_2010} showed that even clear contact lenses with no visible texture cause a small degradation in match score. Clear lenses for toric prescription, and lenses with logo or lettering embedded in them cause slightly larger, but still small, degradation in match score. However, Doyle \etal \cite{Doyle_ICB_2013} showed that wearing textured contacts nearly guarantees a FNM result. Images of the same iris from two different sessions wearing the same brand of contact lens do not match appreciably better than images of the iris wearing textured contacts at one time and no contacts at another time. The contact lens dataset used in this work was the first to be made generally available to the research community, and has since been enlarged and used in the LivDet competitions \cite{Yambay_IJCB_2014,Yambay_IJCB_2017}. Yadav \etal \cite{Yadav_TIFS_2014} presented results that largely confirm those of Doyle \etal \cite{Doyle_ICB_2013}. They added study of a dataset representing additional commercial iris sensors and additional manufacturers of lenses. For a moderate security setting (FMR of 1-in-10,000) they observed a drop in verification rate of 22\% to 38\%, depending on the sensor, when matching an image of an iris with no contact lens with the other image that had a textured contact lens, and from 50\% to 64\% when matching images of the iris wearing the same brand of textured contact lens. 

The general lesson is that an attacker whose goal is to evade detection by generating a FNM result can do so relatively easily by using textured contact lenses. It is also widely believed in the research community that an attacker could use textured contact lenses to impersonate a targeted enrollment. First among the conditions is that the attacker would have custom-designed textured contacts chosen to match the targeted enrollment. Also, the textured lenses should be opaque, in the sense of the texture in the textured portion blocking 100\% of the natural iris texture. And the textured region in the lens should be broad, to represent a minimal dilation condition, so as to decrease chances of natural texture showing from underneath the contact lens. While this attack seems plausible in principle, it may be difficult and expensive to achieve. We are not aware of this attack having ever (yet) been successfully demonstrated.

\subsubsection{Attack Technology: Displays}

Numerous papers suggest that an iris image or video displayed on an electronic screen can be used in a presentation attack \cite{HeXiaofu_ICB_2009,Singh_ICT_2011,Huang_WACV_2013,Wei_FRONTEX_2013,Connell_ASSP_2013,Sun_HoBAS_2014,Sequeira_IJCNN_2014,Marsico_PRL_2015,Kumar_NCVPRIPG_2015,He_BTAS_2016,Sequeira_TSP_2016,Fathy_WPC_2017}. This can only be successful when the electronic display and the sensor operate in the same range of wavelength, as illustrated in Figs. \ref{fig:samples:display}a-b. In particular, iris recognition methods proposed in academic papers for visible-light iris images, if implemented in practice, would have to use visible-light acquisition devices that would photograph iris images displayed on regular LCD screens, as demonstrated in various papers \cite{Das_PRL_2016,Raghavendra_IJCB_2014,Raghavendra_TIFS_2015,Raja_BTAS_2015,Raja_SIN_2016,Raja_TIFS_2015}.

This, however, cannot be generalized to commercial iris sensors, as they use near-infrared light to illuminate the iris as recommended by ISO/IEC 29794-6. The sensors may additionally cut the light outside of the 700-900 nm range by applying near-infrared filters. Fig. \ref{fig:samples:display}c presents what the IrisGuard AD100 sensor can ``see'' when the content is presented on the iPhone display, in the same way as presented in Fig. \ref{fig:samples:display}b. We do not know any off-the-shelf LCD displays emitting near-infrared light, and we do not know any commercial iris recognition systems operating in visible light. Hence, the probability of using regular, visible-light displays in spoofing of current commercial iris recognition systems is minimal. Some earlier tests confirm this: ``The tested system was shown to be resistant to (...) an image shown on an iPhone screen'' \cite{Dunstone_BTT_2011}. 

An exception is application of early e-readers {which} implement {\it e-ink} technology that does not require a backlit display, which can present good quality content also if illuminated and observed in near-infrared light by a commercial iris recognition sensor, Fig. \ref{fig:samples:display}d.

\subsubsection{Attack Technology: Prosthetic Eyes}

Use of prosthetic eyes is often mentioned as a potential presentation attack \cite{Zuo_TIFS_2007,Toth_EB_2009,Rigas_PRL_2015,Toth_EB_2015,Czajka_TIFS_2015, Galbally_IWBF_2016, Czajka_Handbook_2016}. Such prostheses are typically hand-crafted by ocularists with care to make the final product as similar as possible to the living eye, Fig. \ref{fig:samples:other}a, left.

In general, preparation of a prosthetic eye requires time and a lot of experience. The resulting product is typically so good that even {near-infrared} images, acquired by commercial {sensors}, resemble {near-infrared} samples of living, healthy eyes, Fig. \ref{fig:samples:other}a, right{.} {O}nly specular reflections observed in the central part of the image may suggest unusual structure of the cornea. This means that it is possible to generate an image of the prosthetic eye that is compliant with ISO/IEC 19794-6:2011 and use it in evading recognition. Indeed, Dunstone \etal \cite{Dunstone_BTT_2011} report that ``tests using a glass eye with a contact lens and blacked-out pupil demonstrated that the removal of visible artefacts in the pupil region, due to misalignment or other factors, did lead to successful spoofs.''

However, we are not aware of any successful impersonation attack that used a prosthetic eye with an iris texture matching a living eye texture. While it is theoretically possible, it would require a significant amount of labor by the ocularist who would have to copy a complicated iris pattern in fine detail.

\subsection{Actual Eye}
\label{sec:KnownVulnerabilities_ActualEye}

\subsubsection{Attack Technology: Non-Conformant Use}

Iris recognition requires user cooperation. Thus the easiest way to evade the recognition is a presentation that does not comply with the expected manner of presentation. Such intentional, non-conformant presentations may include excessive eyelid closure that results in a smaller number of iris features possible to be used in matching, thus increasing the probability of incorrect match; or looking away from a camera lens that causes the 2D projection of an actual iris to deviate from a circular shape. If an algorithm does not implement adequate methods for compensating off-axis gaze, such presentation may result in a false match. An attacker can also intentionally increase {the} mutual rotation between the sensor and the eye, by either rotating the camera (for instance upside-down), or rotating the head, or both \cite{Czajka_TIFS_2017}. Since iris recognition is sensitive to eye rotation, and not all sensors implement countermeasures preventing their excessive rotation during acquisition, the attacker may be able to generate a {false non-match result}.

A well-documented non-conformant use of one's eye for a presentation attack in an operational environment was based on administering eye drops that result in excessive mydriasis to bypass an iris recognition-based border check in the United Arab Emirates \cite{Al-Raisi_TI_2008}. An immediate, and pioneering in operational environment, countermeasure applied by UAE was to reject images with the pupil-to-iris radius ratio larger than 60\%.

\subsubsection{Attack Technology: Cadavers}

The idea of using non-living organs in presentation attacks has probably emerged from movies{.} We are not aware of any reported successful attack on a commercial iris recognition system that use{d} cadaver eyes. However, it is possible to acquire {a} post-mortem iris image using commercial iris sensors, in cold temperatures {(around 6\degree~Celsius / 42.8\degree~Fahrenheit)} even up to one month after death, and get a correct match between this sample and either ante-mortem counterpart \cite{Sansola_MastersThesis_2015}, or the other post-mortem image of the same eye \cite{Trokielewicz_ICB_2016,Trokielewicz_BTAS_2016}. The earliest experiments known to us with matching post-mortem iris images were by Sansola \cite{Sansola_MastersThesis_2015}. She used an IriTech IriShield MK 2120U system to show that post-mortem iris recognition is plausible up to 11 days after death. Sansola also presented the only case known to us so far of correct matching of ante-mortem iris image and the corresponding sample taken 9 hours 40 minutes post-mortem. Trokielewicz \etal \cite{Trokielewicz_BTAS_2016} were the first to present the {biometric recognition} accuracy of post-mortem iris recognition up to 34 days after death and for four different iris matching methods, and published the only database of post-mortem iris images available to date. Bolme \etal \cite{Bolme_BTAS_2016} presented a study of iris decomposition in outdoor conditions in cold, medium and warm temperatures and ``found a small number of irises that could be matched and only in the early stages of decomposition.'' Recently, Sauerwein \etal \cite{Sauerwein_JFO_2017} confirmed earlier conclusions of Trokielewicz \etal about viability of post-mortem iris recognition when the body is kept in cold temperatures, and conclusions delivered by Bolme \etal, reporting that in warm temperatures it's rather difficult to acquire a clear iris image. These papers demonstrate that it is possible to use cadavers to get a correct match if the iris is imaged in the first days after death.

\subsubsection{Coercion}

We are not aware of any reported cases of presenting irises under coercion in commercial system. Also, there are no published papers reporting any research in this area. This may be a consequence of a relatively difficult data collection that (a) should be done in authentic situations incorporating coercion, and thus (b) could be rejected by the Institutional Review Board. On the other hand, it is relatively easy to imagine a scenario in which such an attack happens in real-world{, not monitored} setups. Hence, for the sake of completeness, we list {this as a potential vulnerability}.

\section{Benchmark databases}
\label{sec:BenchmarkDatabases}

This section summarizes relevant datasets available to the research community. Most were created specifically to serve as iris PAD benchmarks. We also list a few that were first introduced in papers focused on biometric recognition, and then proved useful in development of iris PAD. We do not discuss {dataset} licensing details. 

The presented datasets vary significantly in many factors.  Table \ref{tab:BenchmarkDatabases} compares their most important technical details. The benchmarks are grouped by the institution that published the data, and oldest datasets are presented first.

Instead of providing a short summary of each dataset separately, in the following subsection we make comments related to the dimensions used in Tab. \ref{tab:BenchmarkDatabases} juxtaposing papers, or group of papers, to illustrate different approaches applied by the authors. Additionally, in the next subsection we provide comments on general aspects of the preparation and distribution of good quality PAD datasets.

\subsection{Summary of current benchmarks}

\paragraph{Type of samples.} We follow the most popular categorization of samples into authentic ones, paper printouts, textured contact lenses, prosthetic eyes, post-mortem irises, {and} synthetic irises, but also samples acquired in replay attacks, as well as time series representing pupil dynamics, eye movement and eye gaze. The {\bf first observation} is that not all datasets offer both {authentic} and {fake} samples. For instance, five datasets ({\sf IIITD Iris Spoofing}, {\sf Post-Mortem-Iris v1.0}, {\sf CASIA-Iris-Syn V4}, {\sf Synthetic Iris Textured Based} and {\sf Synthetic Iris Model Based}) offer only {fake} samples. In contrast, two datasets ({\sf Pupil-Dynamics} v1.0 and {\sf CAVE}) offer only {authentic} samples. These example datasets are certainly still useful, and can be used either in development of open-set PAD, or can serve as an additional source of samples when merged with other datasets. The {\bf second observation} is that categories of samples are populated non-uniformly across datasets. The most popular are {\bf static samples}: 17 different databases offer images of irises printed on paper and presented to the sensor. The following databases include printouts having the pupil area cut out: {\sf LivDet-Iris Warsaw 2013}, {\sf LivDet-Iris Warsaw 2015}, {\sf LivDet-Iris Warsaw 2017}, {\sf LivDet-Iris Clarkson 2015 LG} and {\sf ETPAD v1}. In preparation of all remaining datasets, the authors presented the original printouts to the sensors. The second most popular static artifacts are images of eyes wearing textured contact lens, offered currently by 11 databases. One important factor differentiating these benchmarks is whether they include contact lenses provided by different vendors. All datasets, except for {\sf CASIA-Iris-Fake}, include textured contact lenses from different manufacturers. 

{The} {\sf IIITD Iris Spoofing} dataset is the only benchmark that provides a combination of the two above attack means. Namely, it includes photographs of paper printouts of images acquired for eyes wearing textured contact lenses. However, the authors report worse genuine comparison scores when comparing authentic eyes with these hybrid attacks, compared to either using images of textured contact lenses or using images of paper printouts of living eyes. Hence, it seems that this hybrid way of preparing the artifacts does not improve the attacks.

Other types of static {fake} samples are less popular. Two datasets ({\sf IIITD Combined Spoofing} and {\sf CASIA-Iris-Fake}) include synthetic irises. Five datasets ({\sf PAVID}, {\sf GUC-LF-VIAr-DB}, {\sf VSIA}, and the one prepared by Das \etal \cite{Das_PRL_2016}) offer recordings of authentic eyes replayed on a screen and presented to another visible-light sensor, typically a smartphone camera. {{\sf VSIA} database is unique in the sense that it offers fake samples originating from 5 different attack types corresponding to the same authentic iris image.} {We found also two unique databases, one offering images of prosthetic eyes ({\sf CASIA-Iris-Fake}) and the other offering images of postmortem irises ({\sf Post-Mortem-Iris v1.0}).}

There are also databases offering {\bf dynamic measurements}. There is one benchmark ({\sf Pupil-Dynamics v1.0}) composed of times series representing pupil size before and after light stimuli. Three datasets ({\sf EMBD}, {\sf ETPAD v1} and {\sf ETPAD v2}) offer eye movement data, and one ({\sf CAVE}) offers the eye gaze positions.

\paragraph{Wavelength of the illuminating light and sensors used in acquisition.} In a majority of datasets including static samples{,} near-infrared illumination has been used to acquire images. {\sf Post-Mortem-Iris v1.0} is unique in this respect since it offers both near-infrared and visible-light images of the same specimens. Visible-light acquisition makes sense due to interest in moving iris recognition onto mobile devices that are rarely equipped with near-infrared, iris-recognition-specific illumination. However, since there is no standard for visible-light iris image format (such as ISO/IEC 19794-6:2011 for near-infrared samples), the resulting quality of visible-light samples depends on the subjective assessment of the dataset creator. Samples in all databases were acquired in laboratory environment{s}, except for {\sf UVCLI} which offers images of the same specimens acquired in both indoor and outdoor conditions.

\paragraph{Spatial or temporal resolution of samples.} The majority of authors used commercial iris recognition sensors and the prevalent resolution of images is thus $640 \times 480$ (\verb+IMAGE_TYPE_VGA+ format defined by ISO/IEC 19794-6:2011). Visible-light samples are acquired by general-purpose cameras and have greater resolution, except for {\sf MobBIOfake} benchmark. The Nyquist theorem provides a theoretical limit to what maximum spatial frequency we may observe for a given sampling rate. For lower scanning resolutions, one is less capable to use high-frequency properties of patterns in the PAD. In particular, it is relatively easy to print an iris at the resolution at least twice as the actual scanning resolution used in commercial equipment, and thus make the artificial pattern ``invisible'' to PAD methods, especially when this sensor is equipped with anti-aliasing filter.

\paragraph{Unique patterns in the authentic and fake subsets.} The information about identity (either authentic or fake) associated to each sample is important and allows to perform subject-disjoint {analyses}. Number of unique identities {represented} by authentic samples is provided by the authors except for {\sf LivDet-Iris Clarkson 2015 Dalsa} and {{\sf IIITD-WVU}.} However, the number of unique fake identities, either derived from actual identities (as for paper printouts) or artificially generated (as for contact lenses or synthetic irises) is less often provided. The authors of {\sf LivDet-Iris Warsaw} (all three editions), {\sf ATVS-FIr}, {\sf ETPAD v1}, {\sf GUC-LF-VIAr-DB}, {\sf VSIA} {and {\sf MobBIOfake}} datasets declare that the same identities are represented by both authentic and {fake} samples. This allows not only to test the PAD mechanisms but also to verify the robustness of the iris recognition software to print attacks. {One should note that providing a reliable number of unique fake patterns in the case of patterned contact lenses is not possible. Fundamentally, the texture seen by the sensor is a mix of the contact lens texture and some amount of the natural iris texture seen through the clear parts of the textured lens. Contact lenses do not stay in the same position on the eye, and so the mix of lens and iris texture can change from image to image.}

\paragraph{Number of samples representing authentic and fake specimens.} Databases differ significantly in this dimension, offering from zero to more than 100 thousand {fake} samples. It starts to be a reasonably large number of samples to prepare PAD solutions with good generalization capabilities.

\paragraph{Train/test split.} {O}fficial splits into train, validation and test subsets facilitate fair comparison of different algorithms using the same benchmark. Without official splits, cross-validation techniques applied in different ways on the entire dataset may {yield} results that are impossible to compare. For instance, non-subject disjoint splits may result in underestimation of the error rates, when compared to subject-disjoint evaluations. In the worst case some papers may report only the accuracy on the training set (\eg if all samples from a given benchmark were used in training). Thus, offering official, well-designed splits (\eg subject-disjoint, sensor-disjoint, artifact-brand-disjoint, etc.) leverages progress in development of PAD techniques that generalize well into unknown specimens.

\subsection{Preparation and distribution of good quality PAD datasets}

The previous subsection shows that the ways {in which} the datasets are prepared, described and distributed are heterogeneous. In this subsection we discuss {ways} to increase the uniformity {and usefulness} of future benchmarks.

\paragraph{Assessment of the quality of fake samples.} There are two contradictory dimensions that should be considered simultaneously: a) diversity of data that helps in development of solutions that generalize well to unknown samples, b) high quality of samples to make them close to artifacts used in real attacks. ``High quality'' does not necessarily refer to common definitions such as resolution, image clarity or contrast. {Rather,} {it} should be understood as a possibility to use the artifacts in {\it successful} presentation attacks conducted on commercial system{s}. These two goals, however, cannot be achieved at the same time. It is relatively easy to increas{e} the diversity, sometimes called ``difficulty'', and immediately fall into the trap of adding samples that would never be correctly processed or even acquired by a commercial system. On the other hand, by strict control of the quality one may produce samples that illustrate only a narrow spectrum of possible attacks.

Ideally, we should aim at both high quality and high diversity of samples. One possible approach is to start with high, yet reasonable diversity, and decrease it until all samples are successfully used in real presentation attack. This is a very rare practice, and the only databases that followed such quality control are {\sf LivDet-Iris Warsaw 2013}, {\sf LivDet-Iris Warsaw 2015} and partially {\sf LivDet-Iris Warsaw 2017} ({\it train} and {\it known test} subsets). The authors of {the} {\sf LivDet-Iris Warsaw} benchmarks first enrolled all subjects to the Panasonic ET100, created printed versions of their irises (using different printers, resolutions, papers, number of color channels, and applying various image enhancement methods prior printing), and they selected to the final benchmark only those samples that were matched to the genuine references by a commercial system.

\paragraph{Lack of standards related to data re-distribution.} The way that the PAD databases are re-distributed differs among the benchmarks. {L}egal aspects related to the license agreements {of course} must adhere to specific rules that are in force in the country of data owner. However, various aspects such as declaration of time from executing the license to getting a copy of the data, or type of metadata attached to the samples are barely standardized.

\paragraph{Standard format of presenting a database.} Papers offering PAD benchmarks use various formats of presenting the data and the baseline results. In particular, it is a rare practice to provide a number of fake identities represented in artifacts. Database creators also rarely discuss how the quality of {fake samples} was verified and if the artifacts correspond to real presentation attacks. Approximately half of current benchmarks offer official splits into train and test subsets{.} {H}owever informing if the proposed splits are subject-, sensor-, or {attack-instrument-disjoint}, is rare. Also, despite the ISO/IEC SC37 efforts to standardize the PAD evaluation, the performance of baseline methods does not always {conform to the ISO/IEC 30107-3 standard}.

\section{Presentation Attack Detection Methods}
\label{sec:PADmethods}

\subsection{Classification of methodologies}
\label{sec:PADmethods_Classification}

A useful framework for understanding the many different PAD research efforts is built from two simple distinctions: Is the iris (eye) considered as a static or a dynamic object? Is the stimulation of the iris by the sensor considered as passive (not designed to induce a change in the iris) or active? These two dimensions result in four classes of PAD methods, earlier proposed by Czajka \cite{Czajka_TIFS_2015}, {that} we use in this survey:

\begin{enumerate}[label=\textbf{\arabic*})]
    \item {\bf Static iris passively imaged.} Methods of this class employ a still image able to reveal only static eye features. No additional active measurement steps are performed. Usually the same picture as used later in iris recognition is employed for PAD. The use of various texture descriptors, such as LBP or BSIF, is a good example of methods belonging to this class.
    
    \item {\bf Static iris actively imaged:} As above, methods of this kind do not use eye dynamics. However, an iris image acquisition is performed with an additional stimulation of the eye that delivers an extra information about structural properties of the eye not observed without such stimulation. An example PAD method in this group can be based on multi-spectral imaging, in which additional information not related to iris dynamics is derived from multiple measurements.
    
    \item {\bf Dynamic iris passively imaged.} Methods of this group detect dynamic properties of the measured object, yet without its stimulation. For instance, an algorithm detecting spontaneous pupil size oscillations (a.k.a. {\it hippus}) belongs to this group.
    
    \item {\bf Dynamic iris, actively imaged.} Methods belonging to this category are the most comprehensive, and dynamic features of the eye are estimated with the specially designed stimulation. This increases the chances to find features of an authentic object that significantly differ from a noise. {Analysis} of the stimulated pupil reflex is an example method belonging to this class.
\end{enumerate}

It is interesting to see how many of the proposed PAD methods can be implemented in current iris acquisition systems with little or no effort. This is the third dimension we use in grouping of the PAD methods, in which we group the methods into two classes:

\begin{enumerate}[label=\textbf{\arabic*})]
    \item {\bf Commercially-ready PAD methods.} Methods in this group can be applied in a {\it basic iris sensor}, which a) has two illuminants that can do direct-eye illumination or cross-eye illumination (\eg IrisGuard AD100 or LG4000) using at least two different near-infrared wavelengths (\eg CrossMatch ISCAN2 or Vista EY2P), or implements a single visible-light illumination, b) is capable of acquiring still images and analyzing iris videos in a single presentation, and c) allows to upgrade its firmware to incorporate PAD-related processing. For instance, a PAD method using LBP texture features and an SVM classification can be implemented in such a basic iris sensor. 
    \item {\bf Hypothesized PAD methods.} Methods in this group require some hardware beyond that in the basic iris sensor. For instance, use of pupil dynamics in PAD would require adding visible light stimulus to the hardware.
\end{enumerate}

Current commercial iris sensors vary in illumination, optics, and the acquisition procedure, and the technical details are often not fully available. We are thus aware that some commercial systems implement more complicated capture processes than assumed in the basic model, and some methods identified as hypothesized in this survey might be implementable in current equipment from selected vendors.

\begin{figure}[!htb]
    \centering
    \includegraphics[width=0.55\textwidth]{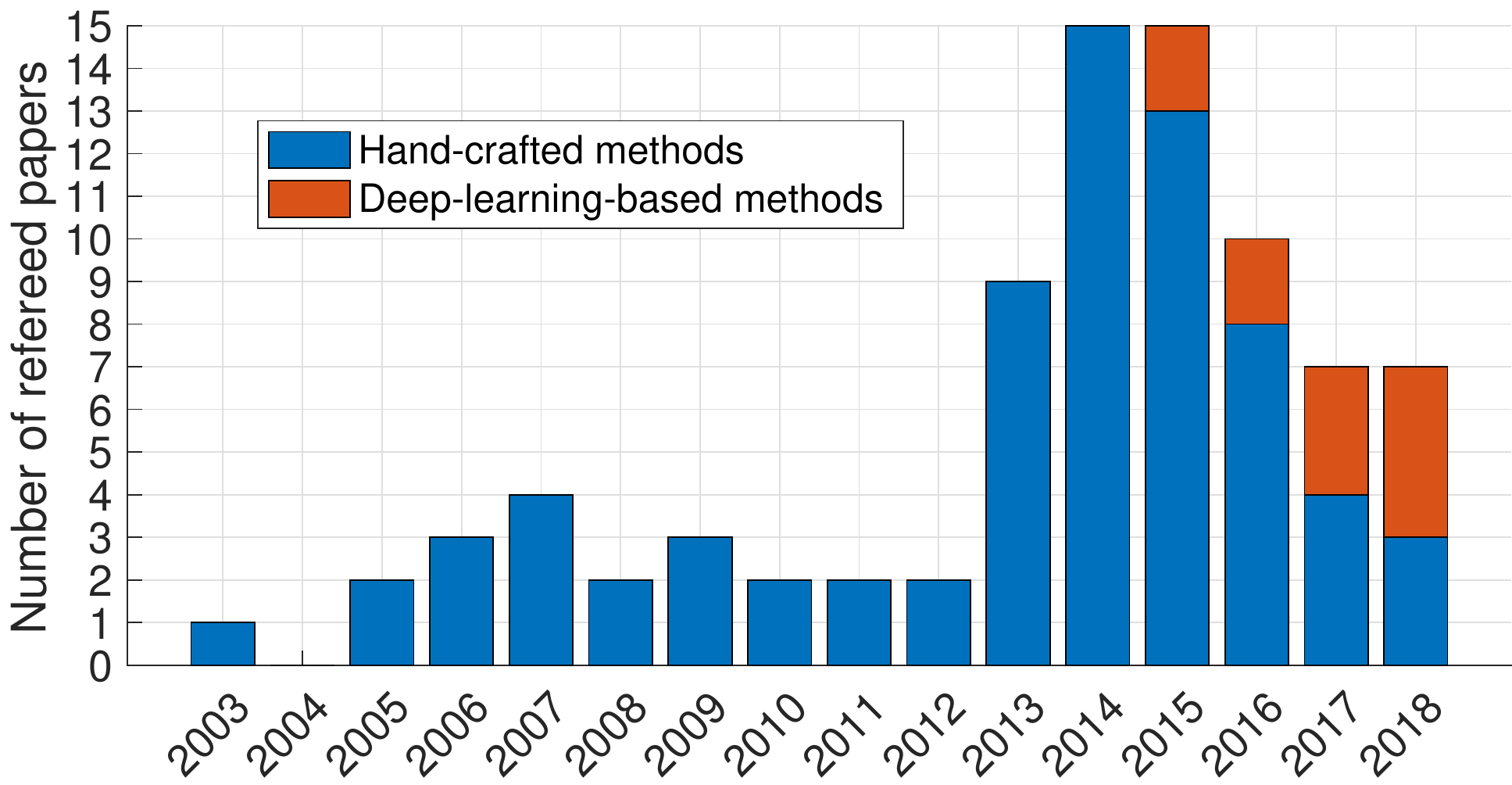}
    \caption{Number of papers proposing iris presentation attack detection methods. Note that a single paper may offer more than one algorithm{.} Thus, the actual number of the proposed PAD methods is larger than a simple paper count.}
    \label{fig:noOfPapers}
\end{figure}

Figure \ref{fig:noOfPapers} depicts the number of papers proposing various PAD methods discussed in this survey. There are three general conclusions from this Figure. First, it seems that the first iris PAD competitions organized in 2013 (LivDet-Iris, Tab. \ref{tab:Competitions}) and 2014 (MobILive, Tab. \ref{tab:Competitions}), stimulated research in iris PAD since in 2014 and 2015 a larger number of iris PAD-related papers appeared. Second, we can observe a gradual decrease in the number of accepted papers, probably due to more demanding state-of-the-art results observed each year that are more difficult to surpass. Third, the number of methods using feature extractors that learn an appropriate processing directly has come to be larger than the number of methods that use experts' knowledge in algorithm's design.

\begin{figure}[!htb]
    \centering
    \includegraphics[width=0.7\textwidth]{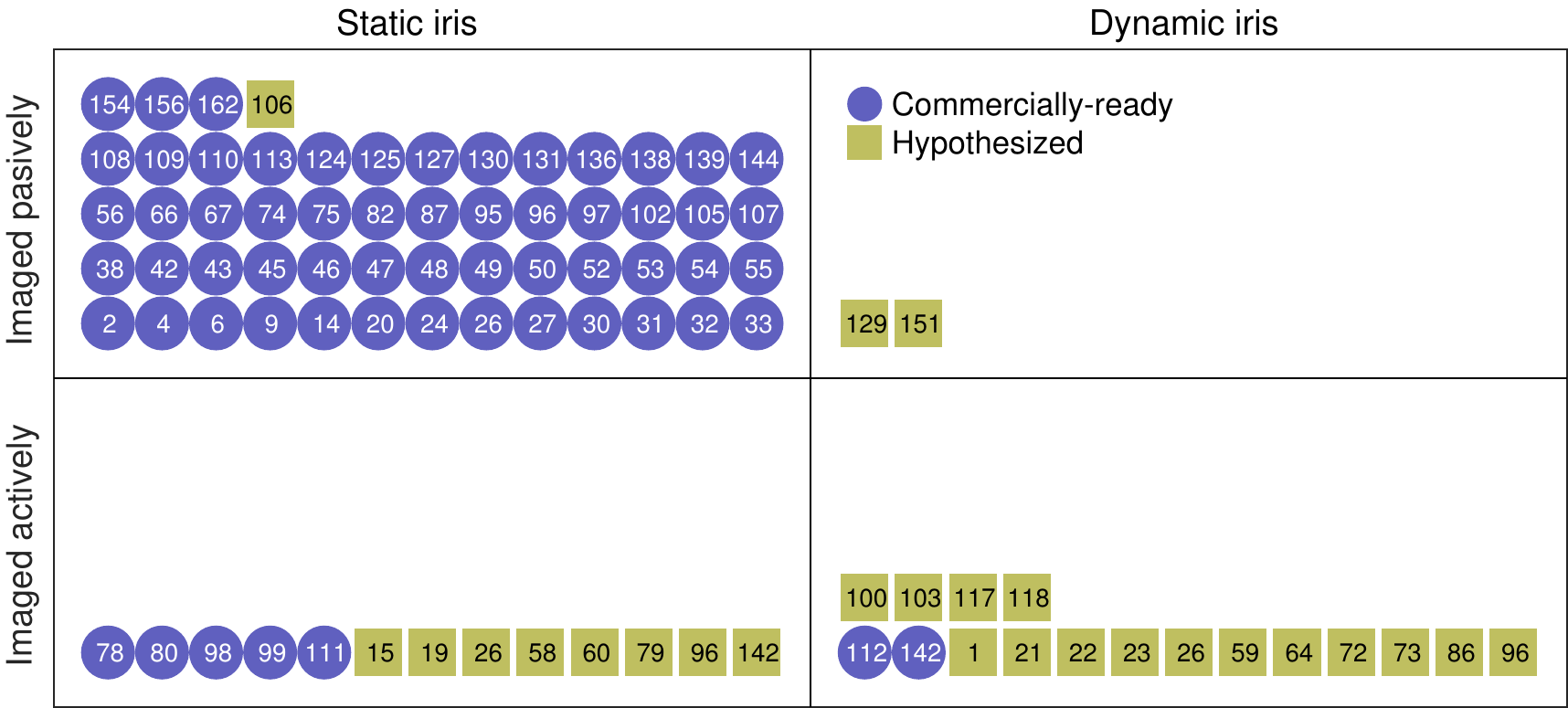}
    \caption{References to papers offering methods grouped into four main categories: a) static iris, passively imaged, b) dynamic iris, passively imaged, c) static iris, actively imaged, and d) dynamic iris, actively imaged. Additionally, in each category we show a split into methods that can be used in a baseline configuration of current commercial sensors (blue circles), and methods that hypothetically can be used in sensors going beyond a baseline configuration ({orange} squares).}
    \label{fig:splits}
\end{figure}

Figure \ref{fig:splits} depicts how the methods discussed in this survey are grouped. A few papers propose multiple methods, and so may appear in more than one quadrant in Fig. \ref{fig:splits}. One immediate observation is that methods using a single iris image, and not analyzing dynamic features of the eye, greatly outnumber all other {methods}. Also, almost all approaches in this group are commercially-ready algorithms. The second largest group of methods corresponds to stimulus-driven measurement of dynamic iris features. Two remaining groups, related to active measurement of static iris and passive measurement of dynamic iris, are less populated. This may suggest that it is more difficult to achieve good performance when dynamic features are not stimulus-driven, and the active, thus more complicated, measurement of a static iris does not significantly increase the PAD reliability when compared to a simpler, passive measurement.

\subsection{Past reviews and general considerations}
\label{sec:PADmethods_General}

Despite a rich literature on iris PAD, there are currently no surveys providing a comprehensive assessment of the state of the art. The first short survey by Galbally \etal \cite{Galbally_SWB_2007} lists ideas and implementations proposed by Daugman (detection of ``alien'' spatial frequencies present for printed irises, coaxial retinal back reflection, Purkinje reflections, detection of spontaneous pupil size changes), Pacut and Czajka (estimation of 3D properties of an eyeball, and pupil light reflex), along with challenge-response transactions, wherein subjects' blinking or eye movement is analyzed. The same PAD methods are mentioned by Nixon \etal \cite{Nixon_HoB_2008}, and Singh and Singh \cite{Singh_ICT_2011} in their short surveys on biometric liveness detection. {These iris PAD methods are grouped by Toth into hardware-based and software-based techniques \cite{Toth_EB_2015}, and hardware-based methods are divided into three subcategories a) intrinsic properties of a living body, b) involuntary signals of a living body, and c) bodily responses to external stimuli, and replaced with coarser categorization \cite{Toth_EB_2009}.}

Looking for potential iris PAD surveys published in last five years, Wei \etal \cite{Wei_FRONTEX_2013} give a brief overview of PAD in the context of the European FastPass project. Bowyer and Doyle \cite{Bowyer_COMPUTER_2014} give a brief overview specifically of the problem of detecting textured contact lenses. They make the point that a technique that appears successful when trained and tested with images representing one contact lens manufacturer may fail drastically to generalize to lenses from a different manufacturer. Akhtar \etal \cite{Akhtar_SECURITY_2015} categorize iris PAD into frequency spectrum analysis, reflectance analysis, dynamics analysis, and texture analysis. They suggest a number of directions for future research, including a comprehensive evaluation framework to rate PAD performance, integrating comparison scores with liveness values, and cross-sensor and cross-dataset liveness detection. Galbally and Gomez-Barrero \cite{Galbally_IWBF_2016} divide the area into sensor-level (hardware-based) and feature-level (software-based) approaches. Thavalengal and Corcoran \cite{Thavalengal_CEM_2016} discuss the challenges of implementing iris recognition on smartphones. They consider the literature on iris PAD as divided into two categories: ``techniques that require special hardware or user interaction'' and ``algorithms designed to work on static images / videos''. They suggest that a PAD technique appropriate to smartphones should not require additional hardware, not require additional user interaction, and should have computational requirements that can be met by a smartphone processor or possible dedicated digital signal processor. {Galbally and Gomez-Barrero \cite{Galbally_IETbook_Ch11_2017} divide presentation attack detection techniques into sensor-level, feature-level and score-level methods, and consider three types of presentation attack instruments: photos, contact-lenses and artificial eyes. The most recent short summary of iris presentation attack detection was proposed by Morales \etal \cite{Morales_PAD_Handbook_2018}. The authors consider zero-effort, photo and video, contact lens and synthetic eye attacks in their work. They also group various PAD approaches found in the literature into hardware-based, software-based and challenge-response, and suggest either serial or parallel integration of PAD with biometric recognition.}

Although we are aware that multi-modal recognition may decrease spoofing probability{, we do not consider multi-modal biometrics as PAD methodology.} For example, Johnson \etal \cite{Johnson_IFS_2010} suggest that fusing results in a multi-modal biometric system makes it more spoof-resistant, even if not all modes are spoofed simultaneously by an impostor. De Marsico \etal \cite{DeMarsico_IVC_2014} suggest that in a multi-biometric setup, which implementation forces several modes to be used indissolubly (for instance the same high-resolution sample is used for face and iris recognition), the PAD method applied to only one mode, for instance face, may strengthen the security of the other mode, for instance iris.

PAD is {an important standardization effort} of the ISO/IEC Joint Technical Committee 1, sub-committee 37 (SC37) on biometrics, since 2011, when the first working draft of the PAD-related standard was prepared. The term ``presentation attack detection'' was developed in 2012 by the SC37 experts, when the fourth working draft of the PAD standard was prepared, and provided a unified definition of previously inconsistent terms such as ``anti-spoofing'', ``liveness detection'', ``spoof detection'', or ``{artefact} detection''. Currently, ISO/IEC 30107 {has four parts}. ISO/IEC 30107-1:2016 harmonizes the PAD-related vocabulary and is freely available at the ISO/IEC Information Technology Task Force (ITTF) web site\footnote{\url{http://standards.iso.org/ittf/PubliclyAvailableStandards/index.html}}. ISO/IEC 30107-2:2017 defines data formats to communicate the PAD results. ISO/IEC 30107-3:2017 provides vocabulary terms related to PAD testing and reporting, and specifies methods and error metrics used to assess the PAD performance. ISO/IEC NP 30107-4 is a new proposal that aims at providing recommendations for assessing the performance of PAD {on} mobile devices.

\subsection{Static-Passive PAD methods}
\label{sec:PADmethods_StaticPassive}

The general approach in this category is that a classifier is trained to categorize images as {authentic} or {fake}, based on a set of features that describe image texture and / or quality. Particular instances of this approach may differ in: (1) the features used, (2) whether the features are computed for the whole image or only the detected iris region, (3) the classifier used, (4) the dataset(s) used in training and testing, and (5) the training and testing methodology. {APCER and BPCER} can change greatly between two different datasets, or two different train and test methodologies for the same dataset. For this reason, it is generally not meaningful to compare accuracy numbers between publications that use different datasets, or different train and test methodologies for the same dataset. 

Most works in this category focus on two main types of attack: (1) presenting a printed image or an image on a display instead of a live iris, and (2) wearing a textured contact lens. Research on detecting textured contacts may be more advanced than research for print attacks, as will be outlined below. Each of these two types of attack can be studied for visible-light images, and / or for near-{infrared} images. This leads to {several} natural sub-categories, below, for summarizing nearly all publications in this category. The same general approach is often viable for multiple sub-categories, and so some publications present results for multiple sub-categories.

\paragraph{Print/Display Attack In Visible-Light}
Publications that consider print or display attack in the context of visible-light images include {\cite{Akhtar_AVSS_2014,Akhtar_ICCST_2014,Alonso-Fernandez_MIPRO_2014,Das_PRL_2016,Gragnaniell_PRL_2015,Menotti_TIFS_2015,Raghavendra_TIFS_2015,Raja_SIN_2016,Sequeira_VISAPPa_2014,Sequeira_IJCNN_2014,Sequeira_TSP_2016}}.

One commonly used dataset in these works is the {\sf MobBIOfake} dataset. Multiple publications report achieving zero or near-zero classification error in PAD for {fake (printed and displayed) images} (\eg \cite{Gragnaniell_PRL_2015}), but research in this area has mostly used the same type of {fake} images in both the training and the testing, {with a few exceptions, for instance \cite{Sequeira_TSP_2016} or \cite{Yambay_IJCB_2017}, which showed that the classification error rates increase significantly when models are trained and tested with different presentation attack instrument species. However, the one-class classification solution proposed by Sequeira \etal \cite{Sequeira_TSP_2016} does not render better results for all tested  presentation attack instrument species, showing that application of open-set classification in the PAD context requires additional research efforts.}

Menotti \etal \cite{Menotti_TIFS_2015} describe one of the earliest iris PAD approaches to use deep learning, which they call {\it SpoofNet}. They experiment with datasets representing visible-light and near-{infrared} print attacks, and also consider face and fingerprint print attacks.

{Raghavendra and Busch
\cite{Raghavendra_IJCB_2014} consider print or display attack presentations, in which the fake image is on a flat surface. They approach PAD in visible-light images by analyzing the variation in focus in the depth images from a light field camera. A discrete wavelet transform analysis is used to estimate the difference in focus values. Results indicate that an APCER between 0.5\% - 2.5\% can be achieved using the proposed approach, depending on the combination of camera for the authentic image and method of presenting the fake image.}

Sun \etal \cite{Sun_PAMI_2014} use Hierarchical Visual Codebook approach to detect multiple types of fake iris images in the {\sf CASIA-Iris-Fake} dataset. In addition to printed iris images and textured contact lenses, the dataset also contains iris texture printed on plastic eyeballs and images ``artificially synthesized from iris images with cosmetic contact lenses.'' 

\paragraph{Print/Display Attack In Near-{Infrared}}
Publications that consider print or display attack in the context of near-{infrared} images include
\cite{Bhogal_BF_2017,Czajka_ICMMAR_2013,Galbally_ICB_2012,Galbally_TIP_2014,Gragnaniello_TIFS_2015,HeXiaofu_ICB_2009,Karunya_ACCS_2015,Menotti_TIFS_2015,Ortiz-Lopez_ICCST_2011,Pacut_ICCST_2006,Czajka_SPIE_2007,Raghavendra_EUSIPCO_2014,Sequeira_IJCNN_2014,Sun_PAMI_2014,Takano_SCE_2009,Pinto_DLB_2018,Sequeira_VISAPPa_2014,Chen_WACV_2018,Raja_BIOSIG_2016,Sollinger_IET_2017}. {These papers typically propose methods based on texture descriptors, spatial frequency analysis, image quality metrics, or deep-learning approaches.} 

Czajka \cite{Czajka_ICMMAR_2013} offered the first publicly available database for this type of attack ({\sf LivDet-Iris Warsaw 2013} in Tab. \ref{tab:BenchmarkDatabases}). The other commonly used dataset in this area is the {\sf ATVS-FIr} datasets proposed by Galbally \etal \cite{Galbally_ICB_2012}, who report that a combination of just two features, iris-to-image size ratio and pupil dilation, achieved zero classification error on the {\sf ATVS-FIr} dataset. Neither of these two features seem specific to presentation attacks, suggesting that authentic and {fake} samples in {\sf ATVS-FIr} may have been acquired in different ways that can be estimated by non-PAD-related features. In any case, given the zero or near-zero classification error rates reported in some papers using {\sf ATVS-FIr} and {\sf LivDet-Iris Warsaw 2013}, it seems time to retire them from use in iris PAD research. Datasets used in the recent LivDet-Iris competition \cite{Yambay_IJCB_2017} could be a better choice, or larger and more challenging new datasets, especially in light of experiments presented by Pinto \etal \cite{Pinto_DLB_2018}, which show that generalization capabilities of deep-learning based models to new presentation attack instruments {are} limited. 

\paragraph{Textured-Contacts Attack In Visible-Light}

We are aware of just one paper to date that looks at detection of textured contact lenses in visible-light images \cite{Yadav_IJCB_2017}. Reasons contributing to the lack of works in this sub-category are: (1) greater difficulty of creating experimental datasets for studying contact-lens attacks compared to the relative simplicity of creating datasets to study print / display attacks, and (2) lack of any commercially viable application of visible-light iris recognition. Yadav \etal \cite{Yadav_IJCB_2017} created the Unconstrained Visible Contact Lens Iris ({\sf UVCLI}) dataset, which contains visible-light images of 70 subjects with and without textured contact lenses, from two different acquisition sessions. They report a baseline identity verification experiment for indoor, same-session (!) iris images without contact lenses, which achieves an EER of just over 13\%. This illustrates the challenge in obtaining acceptable accuracy with visible-light iris recognition. The EER when matching between a live enrollment and a contact-lens probe is generally around 40\%. This illustrates the magnitude of the problem created by textured contacts. Experimenting with three algorithms originally developed for detecting textured contacts in near-{infrared} images, they report a maximum {CCR} of about 83\%. And, this accuracy is achieved with a train/test methodology that is not lens-type-disjoint{, which} would likely result in significantly lower estimated accuracy.

\paragraph{Textured-Contacts Attack In Near-{Infrared}}

Detection of textured contacts in near-{infrared} images has seen more work than the sub-categories summarized above, namely
\cite{Daugman_WMIP_2003,Doyle_ICB_2013,Doyle_BTAS_2013,Doyle_IEEEAccess_2015,Fathy_NRSC_2017,Gragnaniello_TIFS_2015,Gragnaniello_PRL_2016,Gragnaniello_SITIS_2014,Gragnaniello_SITIS_2016,He_ICB_2007,He_BTAS_2016,He_CCPR_2008,HeZhaofeng_ICB_2009,Kohli_ICB_2013,Komulainen_IJCB_2014,Lovish_CAIP_2015,Pala_CVPR_2017,Raghavendra_ESI_2014,Raghavendra_WACV_2017,Sequeira_IJCNN_2014,Silva_SIBGRAPI_2015,Sun_PAMI_2014,Tomeo-Reyes_ISIS_2013,Wei_CPR_2008,Yadav_TIFS_2014,Zhang_CPR_2010,Singh_ISBA_2018,Sequeira_VISAPPa_2014,Chen_WACV_2018}.

A number of early papers reported perfect or near-perfect accuracy in detecting textured contact lenses. However, these works generally had the same types of textured contacts in both the train and test data. Doyle and Bowyer \cite{Doyle_IEEEAccess_2015} emphasized the importance of lens-type-disjoint train and testing methodology in order to obtain a more real-world estimate of error rates. Their results show that {CCR} on lens types not in the training data is generally much lower than for lens types in the training data, and that training on a larger variety of textured lens types improves generalization to unseen lens types.

Komulainen \etal {\cite{Komulainen_IJCB_2014,Komulainen_IETbook_Ch12_2017}} present a detailed analysis of using BSIF texture features to detect textured contact lenses. They use the {\sf ND CLD 2013} dataset and present cross-sensor results and leave-one-lens-type-out results. One conclusion from their study is that BSIF texture features outperform LBP features for this problem.

Clear contact lenses are generally thought to not degrade iris matching performance enough to be considered a presentation attack. Nevertheless, several groups have considered the problem of detecting clear contact lenses as well as detecting textured contact lenses: \cite{Doyle_ICB_2013,Gragnaniello_PRL_2016,Silva_SIBGRAPI_2015,Yadav_TIFS_2014,Singh_ISBA_2018}. Silva \etal train a CNN for the three-class problem of classifying iris images as having: (1) a textured contact lens, (2) a clear contact lens, or (3) no contact lens \cite{Silva_SIBGRAPI_2015}. Interestingly, they find that different numbers of convolutional layers are better for images acquired with different iris sensors, and they report that using $256\times256$ versions of the images results in lower {CCR} than using $64\times64$ versions.
 
The hot current direction in this sub-category is to find methods that generalize well to unseen types of textured contact lenses and to images from different sensors. Experimental datasets that can support this line of research have been collected at multiple institutions and made available to other researchers, and been used in the LivDet-Iris competitions \cite{Yambay_IJCB_2017}.

\paragraph{PAD for Targeted Synthesized Impersonations}

Galbally \etal \cite{Galbally_Handbook_2016} describe an unusual approach to creating a presentation attack, and a method to detect such an attack. Using the stolen iris code, the attacker synthesizes a (somewhat) realistic-looking iris texture image that, if segmented and coded by standard iris recognition algorithms, will result in an very similar iris code.  Galbally \etal show that it is possible to synthesize iris images that can give iris codes that are close matches to the original stolen iris code. Note that the human visual perception of the original iris texture and the synthesized iris texture may be that they look very different. The authors also show that {fake} images created using this approach can be detected using an approach based on image quality features. {Also, Fathy \etal \cite{Fathy_NRSC_2017} train a classifier to distinguish between real and synthetic iris images from the {\sf CASIA-Iris-Syn} dataset.}

\paragraph{PAD for Image-Orientation Attacks}

In Daugman-style iris recognition, difference in mutual rotation between the enrolled and probe samples may result in a false non-match. Because of this, rotating an iris image by 180 degrees allows for generating a second, distinct biometric reference for the same eye.  Also, rotating a sensor by 180 degrees can be a way to conduct a concealment attack{.} (A few manufacturers implement hardware countermeasures against accidental rotation of a sensor\footnote{\eg BMT-20 offered by CMITech: \url{http://www.cmi-tech.com/pdf/cmitech-data_sheet-bmt-20-jan2015.pdf}, or IriShield series offered by IriTech Inc.: \url{http://www.iritech.com/products/hardware}}.) Czajka \etal \cite{Czajka_TIFS_2017} propose and compare two approaches to detect the orientation of an iris image. The first one employs ``hand-crafted'' geometrical and intensity features classified by an SVM, while the second employs a CNN that learns an appropriate feature extraction and classification directly from the data. The SVM was able to correctly classify from 98.4\% to 100\% of left/right orientations, and from 94.4\% to 98.5\% of upright/upside-down orientations, depending on the sensor used. The CNN was better than the SVM when the same sensor was used in training and testing, and slightly worse in cross-sensor evaluations. 

\paragraph{Hybrid}

{More recently, Chen and Ross \cite{Chen_WACV_2018} proposed an interesting multi-task convolutional neural network-based approach that simultaneously performs iris localization and presentation attack detection. The authors obtained state-of-the-art performance when testing their solution on two public benchmarks: {\sf LivDet-Iris Warsaw 2015} and {\sf CASIA-Iris-Fake}.}

\subsection{Static-Active PAD methods}
\label{sec:PADmethods_StaticActive}

A straightforward idea to extend {Static-Active} PAD methods is to use static features calculated for multiple images, and make a decision fusion. This approach is presented by Raja \etal \cite{Raja_BTAS_2015} who applied Laplacian pyramid decomposition for multiple visible-light images acquired by a mobile device, and used them with an SVM classifier. The novel element of this work is the creation and evaluation of a (visible-light) video-based presentation attack. The attack video is assumed to come from playing back the enrolled iris video on a display and recording it from another mobile device. 

Two other approaches are more common in this category: (1) multi-spectral image analysis, and (2) investigation of the selected three-dimensional properties of the eye.

\paragraph{Multi-spectral imaging}

This approach assumes that multiple measurements of light reflected from the eye, for a small number (typically two) {of} bands, will deliver features that can distinguish between bona fide and attack {presentations}. The proposed methods use either bands of near-infrared light \cite{Park_LNCS_2005,Park_OptEng_2007,Lee_BS_2006}, or a mixture of near-infrared and visible light \cite{Chen_PRL_2012,Thavalengal_TCE_2016,Hsieh_Sensors_2018}. Park and Kang \cite{Park_LNCS_2005,Park_OptEng_2007} propose to fuse images of the iris regions acquired for individual bands into a single intensity image used for matching. This provides an additional barrier against fake irises being able to produce a usable image, because the fused texture of the fake would likely not be able to match the fused texture of the real iris. Experimental results suggest that this approach appears to provide strong {recognition of fake} iris images. In turn, Lee \etal \cite{Lee_BS_2006} propose a scheme based on the fact that the iris and the sclera reflect near-infrared light of 750 nm and 850 nm differently, and for most types of {fake} irises, such as a printout, the reflectance will not change significantly between the bands. 

Thavalengal \etal \cite{Thavalengal_TCE_2016} explore the use of a hybrid visible-light and near-infrared sensor on a smartphone. The visible-light and the near-{infrared} images are used together to obtain multi-spectral features to classify a single frame as bona fide or attack {presentation}. Also, pupil characteristics are analyzed across a sequence of frames. Results of initial proof-of-concept experiments suggest that this two-stage {PAD} approach can be highly effective. Hsieh \etal \cite{Hsieh_Sensors_2018} propose a dual-band (near-infrared and visible-light) acquisition and apply ICA to separate the actual iris pattern and the texture printed on a contact lens. This allows for both presentation {attack} detection and also increase of the biometric recognition performance for eyes wearing textured contact lenses.

\paragraph{Estimation of three-dimensional features}

The simplest approaches in this group {compare} near-infrared reflections from the cornea and lens to check the basic three-dimensional properties of the eyeball. Lee \etal \cite{Lee_ICB_2005} {follow an early suggestion by Daugman} and detect Purkinje reflections, \ie specular reflections that occur at the outer and the inner boundaries of the cornea, and the outer and the inner boundaries of the lens. The PAD method proposed by Lee \etal acquires two images, in sequence, each with a different collimated near-infrared LED. One of these is used to acquire the distance from the sensor to the eye. This value is used to instantiate an eye model, and based on the eye model, windows in the image are defined to search for the Purkinje reflections. If the reflections are found in the {predicted} locations, then the iris image is accepted as coming from a bona fide presentation. 

Pacut and Czajka \cite{Pacut_ICCST_2006,Czajka_SPIE_2007} assume the cornea to have a spherical shape and, due to its moistness, to generate specular reflections of {near-infrared} light. In their experiments, the sensor was extended with two supplementary sources of {near-infrared} light, placed equidistant to the camera lens. They stimulated reflections from the cornea by switching on and off these additional diodes in a predefined manner. For a bona fide presentation, the detected sequence of reflections should match the original sequence stimulating the {near-infrared} diodes. This simple method proved to be effective for paper printouts, since the authors did {report APCER=}BPCER=0.

An estimation of more complicated 3D properties {was} proposed by Lee and Park \cite{Lee_IMA_2010}, who used photometric stereo approach and multiple illuminants from different directions, to exploit the fact that the surface of a real, live iris is not flat and so will cast different shadows with illumination from different directions. This provides a means of detecting printed iris images and contact lens images, where the surface that provides the apparent texture does not also cast shadows. Experiments are performed with a dataset of 600 live iris images and 600 fake iris images, and an EER of 0.33\% is reported. 

Connell \etal \cite{Connell_ASSP_2013} use a structured-light approach to classify the texture of the iris region as resulting from a{n authentic} iris or a textured contact. The basic idea is that a light stripe projected onto the iris region appears more like a straight {line} for a{n authentic} iris (or clear contact lens), and more like a curved line in the case of a textured contact. Hughes and Bowyer \cite{Hughes_HICSS_2013} present an approach to contact lens detection based on stereo imaging. They approach the problem as classifying the texture in the iris region as coming from a surface better approximated as planar (bona fide {presentation}) or spherical (contact lens attack {presentation}). These two approaches are able to cleanly separate the textured contact images from the others, however they requires custom imaging systems (light-stripe projection or stereo vision) to acquire images.

\subsection{Dynamic-Passive PAD methods}
\label{sec:PADmethods_DynamicPassive}

Shaydyuk and Cleland \cite{Shaydyuk_ICCST_2016} describe an experiment to use laser speckle contrast imaging as a means of liveness testing in retinal biometrics. Liveness testing is approached here as detecting blood flow in the retinal vasculature. This approach could in principle be used in iris imaging. However, typical retinal imaging is generally considered to be less user-friendly than typical iris imag{ing}, and the imaging device for laser speckle contrast would be more complicated.

Villalobos-Castaldi and {Suaste-G\'{o}mez} \cite{Villalbos-Castaldi_BF_2014} explore the use of pupillary oscillation as a biometric trait, which would naturally incorporate liveness detection. The key insight is that the spontaneous pupillary oscillations (``hippus'') that all irises undergo can be unique to a particular eye to a degree that allows it to be used as a biometric. A custom imaging apparatus was used to record video sequences of about 5 seconds duration for each of 50 subjects. {EER = 0.23\%} is reported. However, the observation of hippus may be subject dependent and earlier studies \cite{Pacut_ICCST_2006} suggest its limited usefulness when applied to PAD.

Raja \etal \cite{Raja_TIFS_2015} create a dataset of video clips {and} manually process the video to obtain 30 frames without eye blinks. Attack clips are created by acquiring video of clips playing on an iPad. The PAD detection scheme consists of using Eulerian video magnification to analyze whether the video is of a live iris or a video playback on an electronic screen. They report that 100\% correct classification as bona fide / attack {presentation} can be achieved with as few as 11 frames.

\subsection{Dynamic-Active PAD methods}
\label{sec:PADmethods_DynamicActive}

PAD ideas dominating in this subsection can be in general grouped into those employing conscious reactions to stimuli, and {those employing} unconditioned responses.

\paragraph{Conscious reactions to stimuli}

Adamiak \etal \cite{Adamiak_FedCSIS_2015} use a gaze direction estimation algorithm to develop a challenge-response approach. In their experiments, it is assumed that a marker to attract the subject's gaze is displayed randomly at one of three locations on the display. They find that the ``number of required presentations of a marker necessary for obtaining a T=95\% level of confidence that a subject is actually following the marker equals 23.'' This level of user interaction is probably too extensive for use in typical iris recognition scenarios. However, it is possible that different eye tracking technology could improve this approach, as proposed by Rigas and Komogortsev \cite{Rigas_IJCB_2014,Rigas_PRL_2015}. The basic insight for their approach is that the eye gaze direction as estimated for a{n authentic} eye is different from that estimated for a {fake (printed)} iris. They suggest that, ``due to the hardware similarities between eye tracking and iris capturing systems'', their approach could be used with iris recognition. They report EER in the range of 3-5\% for print attack detection with about 7 seconds of eye-movement recording, and moderate increase in EER for recording times as short as one second. Additionally, Matthew \cite{Matthew_PhD_thesis_2016} suggests to use {a} similar eye tracking device, in addition to an iris acquisition sensor, to detect if a person looks at the designated area in case of coercion. This is the only approach {that we are aware of} proposed to detect coerced use of someone's eye. 

\paragraph{Unconditioned reflex}

Komogortsev \etal 
\cite{Komogortsev_ICB_2013, Komogortsev_TIFS_2015} discuss PAD in the context of eye movement biometrics. They do not deal  specifically with iris texture or iris recognition, but their detailed model of the ``oculomotor plant'' may be relevant to some approaches to iris PAD.

The other ideas in this group use an obvious {behavior} of the pupil, which constricts and dilates depending on visible-light stimulus. Park \cite{Park_AMDO_2006} describes an approach based on pupil dilation and the iris texture near the pupillary boundary. The sensor uses visible light to change pupil dilation and near-{infrared} to acquire a low-dilation and a high-dilation image. The iris texture near the pupillary boundary is compared between the two images. Lack of a dilation change and lack of similarity in the texture comparison are indications of an attack.

Kanematsu \etal \cite{Kanematsu_SICE_2007} compare iris image brightness calculated in {two angular sections} of the iris stimulated by visible light, and the normalized difference between these quantities is used as a PAD score. The authors were able to recognize correctly all printed irises that were kept still, shaken, brought back and forward, or rotated during acquisition.

Puhan \etal \cite{Puhan_ISCE_2011} propose an approach to detecting textured contact lenses by comparing the iris region texture before and after dilation induced by lighting change. However, the authors present no experimental evaluation on real images of subjects wearing textured contact lenses. 

Huang \etal \cite{Huang_WACV_2013} propose to use two iris images acquired under varying illumination and thus with different pupil dilation ratio. The authors use two features calculated for a pair of iris images: a) Kullback-Leibler divergence measuring the difference between two sets of four image patches selected within the iris annulus and b) ratio of iris and pupil diameters. The trained classifier was able to recognize static (attack {presentation}) and dynamic (bona fide {presentation}) objects with {a CCR} of 82.0\%--99.7\%, depending on the strength of the stimulus. 

{P}resentation attack detection based on pupil dynamics has been first proposed by Pacut and Czajka \cite{Pacut_ICCST_2006,Czajka_SPIE_2007,Czajka_patent_2011}. The authors applied a nonlinear pupil reaction model proposed earlier by Clynes and Kohn. Pupil size was measured for 4 seconds after stimulating the eye with visible light and the resulting time series was represented in the feature space defined by model parameters, {classified} for each measurement by a two-layer nonlinear perceptron. {P}aper printouts were easily recognized by this method. Later Czajka extended this work to recognize time series representing odd or no pupil reactions, evaluated both negative and positive visible light stimuli, and used an SVM in classification \cite{Czajka_TIFS_2015,Czajka_Handbook_2016}. He suggests that changes in pupil reaction, when a subject is acting under stress, potentially could be detected automatically without a need of voluntary actions. However, there are no experiments presenting viability of the above approach, mainly due to unfathomable data collection{.} {Recently Czajka and Becker \cite{Czajka_PAD_Handbook_2018} applied also a few variants of recurrent neural networks to recognize correct pupil reaction, however neural models do not present a significant improvement over parametric models explored earlier by Czajka}.

Thavalengal \etal \cite{Thavalengal_TCE_2016} propose a PAD scheme that exploits a hybrid visible-light and near-{infrared} sensor that is available for smartphones. Thus they have four wavebands to work with: blue, green, red and near-{infrared}. A one-class classifier for multi-spectral features of live images is used in the first stage, and an analysis of pupil dynamics is used in the second stage. They find that analysis of the pupil dynamics is important in reducing the APCER in mannequin-based attacks from over 6\% to 0\%.

\section{Competitions}
\label{sec:Competitions}

Competitions use a benchmark dataset and a standardized evaluation protocol to estimate performance of algorithms developed by participants. The dataset is typically split into a training portion, used by competitors to develop their algorithm, and a sequestered testing portion, used by the organizers to make the accuracy estimate. All of this allows for a more rational assessment of the current state-of-the-art than comparing accuracy numbers across different published papers. Table \ref{tab:Competitions} summarizes four international iris liveness competitions organized to date.

The first iris PAD competition was {\it LivDet-Iris 2013}, organized by Clarkson University, University of Notre Dame, and Warsaw University of Technology \cite{Yambay_IJCB_2014}. The competition attracted algorithm submissions from three other universities: Universidad Autonoma de Madrid, {Universit\`{a}} degli Studi di Napoli Federico II, and Universidade do Porto. The training data included images of both iris printouts and textured contact lenses. All iris printouts used in this competition were first used to successfully spoof a commercial iris recognition system, and hence the{se} printouts represented real presentation attacks. The first observation from LiveDet 2013 was that recognition of the {fake} images in the testing data was very hard, despite the relative simplicity of the attacks. The best algorithm, in terms of {an average of APCER and BPCER}, detected 88.07\% of {fake (printed)} images while rejecting 5.23\% of authentic irises. The same method detected 92.73\% of textured contact lenses, while rejecting 29.67\% of authentic irises. The second observation was that detection of textured contact lenses is more difficult than detection of printed iris images. One factor in explaining this is that the printed image reveals its artificial nature in the entire image, while textured contacts change only (a part of) the iris region. 

The second iris PAD competition was organized in 2014 by INESC TEC and Universidade do Porto \cite{Sequeira_IJCB_2014}. This competition used images of irises printed on a paper and acquired by mobile device in visible light. There were six participants and the best algorithm was perfect in recognizing {fake} images and incorrectly rejected only 0.5\% of authentic irises. This achievement dramatically differs from the results observed for the LivDet 2013 winner. However, there are two possible factors why the {\sf MobBIOfake} dataset, used in this competition, resulted in such high accuracy. First, we read that the ``ranking was updated after each new submission by evaluating the algorithms in the same randomly obtained subset of the test set composed by 200 images'' \cite{Sequeira_IJCB_2014}. {This means that the participants were able to observe the performance on a subset of testing data to increase the generalization capabilities of their submissions.} Second, the authors did not report how the quality of the data was controlled, in terms of whether the {fake} images would be accepted for use by any commercial, visible-light iris recognition system operating on mobile devices. Consequently, the {\sf MobBIOfake} dataset might not be challenging for the submitted algorithms.

{\it LivDet-Iris 2015} \cite{Yambay_ISBA_2017}, a continuation of {\it LivDet-Iris 2013}, was organized by Clarkson University and Warsaw University of Technology. The organizers extended their databases used in 2013 edition{,} did not break out their analysis by attack type, paper printouts versus printed contacts, and presented averaged results that suggest a significant improvement in accuracy since 2013. The best ACPER was 5.48\%, versus 9.98\% in {\it LivDet-Iris 2013}, and the best BPCER was 1.68\%, versus 12.18\% in {\it LivDet-Iris 2013}. It is important to note that the winning algorithm presented perfect classification on the {\sf LivDet-Iris Warsaw 2015} test partition. This suggests that this dataset should be retired from use as a benchmark.

The most recent edition of LiveDet-Iris is {\it LivDet-Iris 2017} \cite{Yambay_IJCB_2017} was organized again by Clarkson University, Warsaw University of Technology, University of Notre Dame, West Virginia University, and Indraprastha Institute of Information Technology, Delhi. This competition again used paper iris printouts and textured contact lenses as the types of attack, and introduced two novel elements to the evaluation protocol. The first novel element was splitting the testing datasets, used by Clarkson, Warsaw and Notre Dame in evaluation of the submitted algorithms, into {\it known} and {\it unknown} partitions. The {\it known} partitions were composed of samples acquired by the same sensor and in a similar environment. In turn, the {\it unknown} partitions included samples having different properties from those in the {\it known} subsets. The second novel element was cross-dataset testing. The organizers obtained three submissions. One used SVM on top of the SID, the second was based on CNN, and the underlying concepts of the third method, submitted by an anonymous participant, have not been revealed. The winning solution achieved APCER=0.55\% and BPCER=2.23\% on {\it known} partition, and APCER=23.8\% and BPCER=3.36\% on {\it unknown} partition. In cross-dataset testing, the same winning solution presented APCER=14.71\% and BPCER=3.36\%. These results clearly suggest that generalization to unknown samples is far more difficult than recognizing attacks of known properties.

{Yambay and Schuckers have recently prepared a concise summary of all editions of LivDet-Iris competitions \cite{Yambay_PAD_Handbook_2018}.}

\section{Evaluation of Presentation Attack Detection methods}
\label{sec:EvaluationAndPerformance}

Evaluation of PAD effectiveness fundamentally differs from evaluation of biometric system performance. ISO/IEC 30107-3:2017 lists five dimensions that differentiate these two assessments. First, it is virtually impossible to get a representative number of samples of a given presentation attack instrument due to indeterminate ways the attacker can prepare them. This means that methods used to evaluate biometric recognition will not provide good statistical estimates that would generalize well to other databases. {For instance, Sequeira \etal \cite{Sequeira_TSP_2016} consider the traditional classification approach, in which the assumption is made that both authentic and fake samples accurately represent the bona fide and attack classes, as a ``One-attack'' methodology. They suggest alternative classification approaches such as the ``Unseen-Attack'', in which a binary model is evaluated with samples representing an unknown type of attack that is not present in the training step. They also propose a ``Single-Class'' approach, in which a one-class classifier is trained only with the authentic samples and evaluated with both authentic and fake samples.} Second, the evaluation results are application dependent and thus hard to compare. Third, the PAD evaluation always includes non-cooperative subjects, and the ways they interact with a system are impossible to generalize to other potential attackers. Consequently, the same evaluation protocol may end up with different results depending on subjects used in testing or data preparation. Fourth, the PAD data collected by one biometric system may be insufficient to predict the performance of another system, due to proprietary sensors acquiring PAD signals. And fifth, the same-quality samples presented by testers having different skills may result in large differences in the estimated performance. It may happen that bad-quality samples in hands of a skilled attacker may be more effective than high-quality artifacts presented to the sensor by a novice. Consequently, ISO/IEC 30107-3:2017 provides the following recommendations: a) presentation attack instrument types shall be tested separately, b) acknowledge that a given presentation attack instrument is successful if at least one successful attack was observed, c) when the error rates are calculated, such as APCER or BPCER, the details about a given PAD mechanism, the presentation attack instrument types, the application, the test approach, and the tester's skills should be provided.

There also is a need to address a common pitfall related to use of biased data in PAD training and evaluation. The data will be biased when samples have additional cues for being classified as {authentic} or {fake} that are correlated with the true class labels, {although} not related to the presentation attack type. One example is use one {set of} camera settings to acquire authentic samples, and different settings to acquire PA samples. Another example could be different ratio of males and females in {authentic} and {fake} classes. Since there is much higher probability of observing mascara for women than for men, some cosmetics-related properties of an image (\eg darker eyelashes) can be linked by the classifier with the state of being authentic {or} {fake}. Biased data will especially influence the {deep-learning-based} approaches, since we have limited control of what kind of features these structures derive from samples to perform a classification. Consequently, we propose to add the following dimension to be considered in PAD evaluation protocol: the countermeasures to avoid bias in the evaluation, and estimation of the potentially remaining bias in the data should be provided along with the PAD evaluation results.

\section{Steps Toward Improved Iris PAD}
\label{sec:Conclusions}

Despite a huge effort by both research and industry to develop effective countermeasures, this survey shows that there is still a significant gap between the actual reliability of current PAD methods and the hoped-for reliability. This section proposes a few ideas to push the effectiveness of iris PAD forward.

\paragraph{Generalization to unknown presentation attacks}

One important limitation of current iris PAD methods is limited generalization to unknown presentation attack types. This can be observed when analyzing the LivDet-Iris 2017 results \cite{Yambay_IJCB_2017} {and the work \cite{Sequeira_TSP_2016} that demonstrates the faults in the evaluation of the performance of the PAD methods by using models trained and tested with a single presentation attack instrument species.} The {LivDet-Iris 2017} winning algorithm accepted, on average, 14.71\% of unknown artifacts used in the competition, and for the most challenging of the LivDet 2017 datasets (IIITD/WVU), the false acceptance of {fake samples} by the same winning algorithm was almost 30\%! Open-set classification and anomaly detection are two research areas that may bring new PAD solutions to the table. This generalization requirement is explicitly articulated in one of the biggest PAD-related research efforts worldwide, \ie the IARPA's Odin program\footnote{https://www.iarpa.gov/index.php/research-programs/odin/odin-baa; last accessed March 21, 2018}. The goal of the Odin project is to use PAD to identify both known and {\it unknown} presentation attacks.

\paragraph{Open-source iris PAD methods}

There are various open-source initiatives, such as OpenCV in computer vision, that effectively collect contemporary solutions and global knowledge in the field as ready-to-use software tools. To our knowledge, there are no such initiatives for iris PAD; we do not know of even a single well-documented iris PAD algorithm that is available to the research community as open source. Therefore, one idea is to create a repository of open-source iris PAD methods that can be tested and continuously updated by volunteers working in the iris PAD area. These methods could then serve as baselines for various PAD evaluations.

\paragraph{Seamless exchange of iris PAD databases}

Attackers are typically one step ahead of our PAD proposals, due to informal, and hence probably effective, exchange of skills and conclusions among them. It thus seems that making the PAD databases more versatile and more accessible to the research institutions might get us closer to PAD solutions capable to counteract up-to-date attacks. One possible idea is to create a well-maintained, cross-national, free-to-access, repository of links to iris PAD databases. This repository would gather contact details for the data distributors, distribution rules, declared reaction times (from an execution of the license agreement to getting a copy of the data), and current performance achieved on a given dataset. This would promote the existing PAD-related data, and enable the community to ``retire'' some datasets as no longer challenging.

\paragraph{Trusted and accessible platform for PAD evaluation}

The only current iris PAD evaluation initiative known to us is the LivDet-Iris series \cite{Yambay_IJCB_2014,Yambay_ISBA_2017,Yambay_IJCB_2017}. We are not aware of any platforms specifically prepared to offer ongoing, asynchronous evaluation of iris PAD algorithms, as for instance {\it FVC-onGoing} -- a platform for an on-line evaluation of fingerprint recognition algorithms\footnote{https://biolab.csr.unibo.it/FVCOnGoing/UI/Form/Home.aspx; last accessed March 21, 2018}. The usage of open-science platforms such as BEAT, proposed by Anjos \etal \cite{Anjos_arXiv_2017} as a part of the BEAT European Project, should facilitate these efforts.

\section{Suggested reading}
\label{sec:SuggestedReading}

For those wanting to begin a deeper dive into PAD for iris, we suggest a list of six{, quite different} readings. The purpose is to help to establish a firm foundation and boundaries for better understanding of the big picture of iris PAD research.

\paragraph{Current state-of-the-art experimental competition}

Yambay \etal \cite{Yambay_IJCB_2017} discuss the results of the iris ``LivDet-Iris 2017'' competition. At the time that this survey is written, LivDet-Iris 2017 is the most recent rigorous evaluation of iris PAD techniques. The LivDet-Iris 2017 competition deals with the current two main types of attack -- images of printed iris images and images of persons wearing textured contact lenses. It also deals with the current important theme of real-world testing data being different in some respect from the training data. As you read this paper, keep in mind that LivDet-Iris competitions have so far happened every other year, and the standards for rigorous evaluation are rapidly evolving, so this paper may be quickly superseded by more recent work.

\paragraph{What if your biometric is stolen? Revocable biometrics}

Often in the popular press, and still on occasion in the research literature, there will be a comment about one danger of biometrics being that if you biometric is stolen, it is compromised forever. This is simply misinformation born out of a lack of knowledge about the field. First, the presentation of printed irises, passively displayed on e-reader, with a goal to impersonate us can be detected by most of the methods presented in this survey. Second, the study of ``template protection'', ``cancelable'' or ``revocable'' biometrics goes back about two decades. There is a rich literature on the subject, for instance \cite{Jain_EURASIP_2008,Rathgeb_EURASIP_2011}, and at least one major iris recognition company has been using a revocable biometric scheme for a number of years. The review by Patel \etal \cite{Patel_SPM_2015} is recommended reading for an overview of this important topic.

\paragraph{Standard terminology}

We mentioned earlier that terminology has historically been used inconsistently in this area. In this context, it is worth pointing out that there is a relevant ISO standard (ISO/IEC 30107-1:2016). Reading the standard, while possibly a chore, is good for giving a precise definition to many important fundamental concepts for iris PAD, and should move the field toward more consistent use of terminology in the future.

\paragraph{Early liveness ideas by John Daugman}

Some familiarity with the history of iris recognition helps to have a mature perspective. In this regard, it is recommended to return to the beginning and read Daugman's iris recognition patent \cite{Daugman_patent_1994}. There we see that the two main categories of presentation attack {(presenting a photograph to a sensor and wearing a contact lens)} were already envisioned{.} Later, Daugman proposed a few ideas that became a basis of some current effective PAD methods \cite{Daugman_IMAIP_2000}. In particular, looking for spontaneous and stimulated pupil size variations, or finding anomalies in Fourier spectrum were mentioned by Daugman 18 years ago.

\paragraph{Reverse engineering a matching iris pattern}

Galbally \etal \cite{Galbally_Handbook_2016} describe their approach to reverse-engineering an image that can generate a match to a targeted enrollment. This paper is recommended because it discusses a type of attack that is very different from the print attacks and contact lens attacks that dominate the iris PAD literature. It should give a better appreciation of the need for a systems-level approach to designing against presentation attacks.

\paragraph{Hollywood's favorite spoof: ``cold irises''}

You may recall a favorite movie in which a character used an eyeball extracted from a body to carry out an impersonation attack on a biometric system; Tom Cruise in {\it Minority Report}, Loki in {\it The Avengers}, ... There is a small amount of research published on the feasibility of iris recognition to verify the identity of a deceased person. Understandably, this is a difficult area in which to carry out experimental research. For those interested in this topic, the paper by Trokielewicz \etal \cite{Trokielewicz_BTAS_2016} is a good starting point.

\clearpage
\noindent
\begin{landscape}
\begin{table}[htb!]
    \begin{center}
        \caption{Technical properties of datasets used in development of iris PAD methods. Abbreviations are explained on p. \pageref{tabs:explanation}.}
        \label{tab:BenchmarkDatabases}
        \tiny
        \vskip-3mm
        \begin{threeparttable}
        \begin{tabular}{llllllrrrrrc}
            {\bf Research group(s)}
            & {\bf Benchmark name}
            & {\bf Type}        
            & {\bf Wavelength}
            & {\bf Sensor(s)}
            & {\bf Spatial or temporal}
            & \multicolumn{2}{c}{{\bf \# Distinct irises}}
            & \multicolumn{3}{c}{{\bf \# Samples}}
            & {\bf Train/test} \\
            & {\bf [paper] \color{blue}{[www link]}} & {\bf of samples} & {\bf range} & {\bf used} & {\bf resolution} & live & fake\tnote{1} & live & fake & total & {\bf split}\\
            \hline
            \hline
            Clarkson Univ., USA & {\sf LivDet-Iris Clarkson 2013} \cite{Yambay_IJCB_2014} & CL & NIR & DA &  N/R  & 64 & {N/A} & N/R & N/R & N/R & {yes} \\
            & {\sf LivDet-Iris Clarkson 2015 LG} \cite{Yambay_ISBA_2017} & PP, CL & NIR & L2 & $640 \times 480$ px& 90 & {N/A} & 828 & 2,898 & 3,726 & {yes} \\
             & {\sf LivDet-Iris Clarkson 2015 Dalsa} \cite{Yambay_ISBA_2017} & PP, CL & NIR & DA & N/R & N/R & {N/A} & 1,078 & 3,177 & 4,255 & {yes} \\ 
             & {\sf LivDet-Iris Clarkson 2017}\tnote{2}\hskip2mm\cite{Yambay_IJCB_2017} & PP, CL & NIR & L2, DA, IP & $640 \times 480$ px\tnote{3}& 50 & {N/A} & 3,954 & 4,141 & 8,095 & {yes} \\
            \hline
            Indraprastha Inst. & {\sf IIITD-WVU}\tnote{4}\hskip2mm\cite{Yambay_IJCB_2017}\color{blue}{\cite{IIITD_DBs_URL}} & CL, PP & NIR & C, V, IS, HP, KM & irregular\tnote{5} & N/R & {N/A} & 2,952 & 4,507 & 7,459 & {yes} \\
            of Information & {\sf IIITD Contact Lens Iris} \cite{Kohli_ICB_2013} \color{blue}{\cite{IIITD_DBs_URL}} & CL & NIR & C, V & $640 \times 480$ px& 202 & {N/A} & N/R & N/R & 6,570 & {yes} \\ 
            Technology Delhi, IN & {\sf IIITD Iris Spoofing}\tnote{6}\hskip2mm\cite{Gupta_ICPR_2014} & PP & NIR & C, V, HP & $640 \times 480$ px\tnote{7}& 202 & N/R & 0 & 4,848 & 4,848 & {no} \\ 
             & {\sf IIITD Combined Spoofing} & PP, CL, SY & NIR & C, V, HP & $640 \times 480$ px & 1,744 & 2000\tnote{8} & 9,325 & 11,368 & 20,693 & {no} \\ 
             & {\sf Database}\tnote{9}\hskip2mm\cite{Kohli_BTAS_2016} \color{blue}{\cite{IIITD_DBs_URL}} &  & & & & & & & & & \\ 
            & {\sf UVCLI}\tnote{10}\hskip2mm\cite{Yadav_IJCB_2017} & CL & VIS & CN6 & N/R & 70 & {N/A} & 1,877 & 1,925 & 3,802 & {no} \\ \hline 
            Univ. of Notre Dame, USA & {\sf ND CCL 2012} \cite{Doyle_ICB_2013} & CL & NIR & L4 & $640 \times 480$ px & {270} & {N/A} & {2,800} & {1,400} & 4,200 & {yes} \\ 
             & {\sf ND CLD 2013} \cite{Doyle_BTAS_2013} & CL & NIR & A, L4 & $640 \times 480$ px & {330} & {N/A} & {3,400} & {1,700} & 5,100 & {yes} \\ 
             & {\sf ND CLD 2015} \cite{Doyle_IEEEAccess_2015} & CL & NIR & A, L4 & $640 \times 480$ px & 556 & {N/A} & 4,800 & 2,500 & 7,300 & {yes} \\ 
            \hline
            Universidad Aut\'{o}noma  & {\sf ATVS-FIr} \cite{Galbally_ICB_2012} & PP & NIR & L3 & $640 \times 480$ px & 100 & 100 & 800 & 800 & 1,600 & {yes} \\ 
            de Madrid, ES & & & & & & & & & & & \\ 
            \hline
            Warsaw Univ. & {\sf LivDet-Iris Warsaw 2013} \cite{Czajka_ICMMAR_2013} \color{blue}{\cite{WARSAW_DBs_URL}} & PP & NIR & A & $640 \times 480$ px & 284 & 276 & 852 & 815 & 1,667 & {yes} \\ 
            of Technology, PL & {\sf LivDet-Iris Warsaw 2015}\tnote{11}\hskip2mm\cite{Yambay_ISBA_2017} \color{blue}{\cite{WARSAW_DBs_URL}}& PP & NIR  & A & $640\times 480$ px & 384 & 376 & 2,854 & 4,705 & 7,559 & {yes} \\
            & {\sf LivDet-Iris Warsaw 2017}\tnote{12}\hskip2mm\cite{Yambay_IJCB_2017} \color{blue}{\cite{WARSAW_DBs_URL}}& PP & NIR & A, P$^{\mbox{\tiny WUT-1}}$ & $640\times 480$ px & 457 & 446 & 5,168 & 6,845 & 12,013 & {yes} \\ 
            & {\sf Pupil-Dynamics v1.0}\tnote{13}\hskip2mm\cite{Czajka_TIFS_2015} \color{blue}{\cite{WARSAW_DBs_URL}}& PD & NIR & P$^{\mbox{\tiny WUT-2}}$ & 25 Hz & 52 & 0 & 204 & 0 & 204 & {no} \\
            & {\sf Post-Mortem-Iris v1.0} \cite{Trokielewicz_BTAS_2016} \color{blue}{\cite{WARSAW_DBs_URL}}& PM & NIR & IS & $640 \times 480$ px & 0 & 34 & 0 & 480 & 480 & {no} \\ 
            & & PM & VIS & TG3 & 4,608 $\times$ 3,456 px & 0 & 34 & 0 & 850 & 850 & {no} \\ \hline 
            Texas State Univ., USA & {\sf EMBD v2} \cite{Holland_TIFS_2013} & EM & NIR & TX, EL, PS & 75, 300 and 1,000 Hz & 227 & 0 & 1,808 & 0 & 1,808 & {no} \\ 
            & {\sf ETPAD v1} \cite{Rigas_IJCB_2014} \color{blue}{\cite{ETPAD_v1_URL}} & EM, PP & NIR & EL, BM & 1,000 Hz, $640 \times 480$ px & 100 & 100 & 400 & 800 & 1,200 & {no} \\ 
            & {\sf ETPAD v2} \color{blue}{\cite{ETPAD_v2_URL}} & EM, PP & NIR & EL, BM & 1,000 Hz, $640 \times 480$ px & 200 & 200 & 800 & 800 & 1,600 & {no} \\ 
            \hline
            Chinese Academy & {\sf CASIA-Iris-Syn V4} \cite{Wei_ICPR_2008} \color{blue}{\cite{CASIA_IrisSynv4_URL}} & SY & N/A & N/A & $640 \times 480$ px & 0 & 1,000 & 0 & 10,000 & 10,000 & {no} \\   
            of Sciences  Int. & {\sf CASIA-Iris-Fake} \cite{Sun_PAMI_2014} & PP, CL, & NIR & H & $640 \times 480$ px & 1,000 & 815 & 6,000 & 4,120 & 10,240 & {no} \\             
           of Automation, CN & & PE, SY & & & & & & & & \\ 
	   \hline
            West Virginia Univ., USA & {\sf Synthetic Iris Textured Based} \cite{Shah_ICIP_2006} \color{blue}{\cite{WVU_DB1_URL}} & SY & N/A & N/A & N/R & 0 & 1,000 & 0 & 7,000 & 7,000 & {no} \\ 
           & {\sf Synthetic Iris Model Based} \cite{Zuo_TIFS_2007} \color{blue}{\cite{WVU_DB2_URL}} & SY & N/A & N/A & N/R & 0 & 10,000 & 0 & 160,000 & 160,000 & {no} \\
            \hline
            Columbia Univ., USA & {\sf CAVE} \cite{Smith_UIST_2013} \color{blue}{\cite{CAVE_DB_URL}} & EG & VIS & CN3 & 5,184 $\times$ 3,456 px & 56 & 0 & 5,880 & 0 & 5,880 & {no} \\ 
	   \hline   
            Gj\o vik University & {\sf PAVID} \cite{Raja_BTAS_2015} \color{blue}{\cite{GUC_PAVID_DB_URL}} & RA & VIS & IP, NL & N/R & 152 & 152 & 608 & 608 & 1,216 & {no} \\
            College, NO & {\sf GUC-LF-VIAr-DB} \cite{Raghavendra_IJCB_2014} \color{blue}{\cite{GUC_LF_VIAr_DB_URL}} & PP, RA & VIS & LY, CN5 & N/R & 104 & 104 & 4,847 & 7,607 & 12,454 & {no} \\
            & {\sf VSIA} \cite{Raghavendra_TIFS_2015} \color{blue}{\cite{GUC_VISIA_DB_URL}} & PP, RA & VIS & CN5 & N/R & 110 & 110 & 550 & 2,750 & 3,300 & {no} \\ 
            & {\sf VISSIV} \cite{Raja_TIFS_2015} \color{blue}{\cite{GUC_VISSIV_DB_URL}} & RA & VIS & NL, IP & N/R & 62 & 62 & 248 & 248 & 496 & {yes} \\
            \hline
	   Griffith University, AU & (no name) \cite{Das_PRL_2016} & RA & VIS & NK & 3,264 $\times$ 2,448 px & 50 & 50 & 500 & 500 & 1,000 & {yes} \\ 
            Indian Statistical Institute, IN & & & & & & & & & & & \\
            University of Las Palmas & & & & & & & & & & & \\
            de Gran Canaria, ES & & & & & & & & & & & \\
            \hline
            INESC TEC, PT and & {\sf MobBIOfake} \cite{Sequeira_VISAPPa_2014} & PP & VIS & AT & {$200\times250$ px} & {100} & {100} & 800 & 800 & 1,600 &  {no} \\ 
            Universidade Federal & & & & & & & & & & & \\
            de S\~{a}o Paulo, BR & & & & & & & & & & \\
            \hline  
            Univ. of Rome, IT & {\sf MICHE-I} \cite{Marsico_PRL_2015} \color{blue}{\cite{MICHE_I_DB_URL}} & PP & VIS & GS, IP, GT & 2,322 $\times$ 4,128 px, & 184 & 40 & 3,652 & 80 & 3,732 & {no} \\
            Univ. of Salerno, IT & & & & & 1,536 $\times$ 2,048 px and & & & & & &\\
            Univ. of Naples Federico II, IT & & & & & $640\times480$ px & & & & & &\\
            George Mason Univ., USA & & & & & & & & & & & \\
            \hline            
        \end{tabular}
		\begin{tablenotes}
		\begin{minipage}{0.1\textwidth}
		\end{minipage}
		\begin{minipage}{0.5\textwidth}
		\tiny
		    \item
        	\item {\bf Notes:}
        	\item N/A = not applicable, N/R = not reported
        	\item[1] number of distinct patterns representing different subjects
			\item[2] superset of the {\sf LivDet-Iris Clarkson 2015}
			\item[3] unknown for Dalsa sensor
		\end{minipage}
		\begin{minipage}{0.5\textwidth}
		\tiny
		    \item
		    \item[4] includes {\sf IIIT-Delhi CLI} and {\sf IIITD IS} samples
		    \item[5] prevailing resolution is $640\times480$px
			\item[6] subset of live samples from the {\sf IIIT-Delhi CLI} has been used		    
			\item[7] unknown for ``print+scan'' samples (prepared with HP scanner)
			\item[8] known only for synthetic irises
		\end{minipage}\hfill
		\begin{minipage}{0.5\textwidth}
		\tiny
		    \item
		    \item
			\item[9] includes {\sf IIIT-Delhi CLI} and {\sf IIITD IS} databases
			\item[10] database not available at the time of writing this paper
			\item[11] superset of the {\sf LivDet-Iris Warsaw 2013}
			\item[12] superset of the {\sf LivDet-Iris Warsaw 2015}
			\item[13] iris segmentation results are made publicly available
		\end{minipage}\hfill		
		\end{tablenotes}
	\end{threeparttable}	
    \end{center}
\end{table}
\end{landscape}

\begin{landscape}
\begin{table}[htb!]
    \begin{center}
        \caption{Summary of iris presentation attack detection competitions open to the public.}
        \label{tab:Competitions}
        \tiny
        \vspace{0.5cm}
        \begin{threeparttable}
        \begin{tabular}{llllllllll}
            {\bf Competition name}
            & {\bf Organizers}
            & {\bf Type}        
            &{\bf  Wavelength}
            & {\bf Type}
            & {\bf Number}
            & \multicolumn{2}{l}{{\bf Best performance (\%)\tnote{1}}}
            & {\bf Algorithm} 
            \\
            & & {\bf of fakes} & & {\bf of evaluation} & {\bf of submissions} & BPCER & APCER & {\bf name}\\
            \hline
            \hline
	    LivDet-Iris 2013 \cite{Yambay_IJCB_2014} & Clarkson Univ., USA & PP, CL & NIR & known {fake} / {authentic} type & 3 & 28.56 & 5.72 & Federico \\
	    & Warsaw Univ. of Technology, PL & & & & & & &\\   
	    & Univ. of Notre Dame, USA & & & & & & & \\\hline
	    LivDet-Iris 2015 \cite{Yambay_ISBA_2017} & Clarkson Univ., USA & PP, CL & NIR & known {fake} / {authentic} type & 4 & 1.68 & 5.48 & Federico \\
	    & Warsaw Univ. of Technology, PL & & & & & & & \\\hline
	    LivDet-Iris 2017 \cite{Yambay_IJCB_2017}  & Clarkson Univ., USA & PP, CL & NIR & known {fake} / {authentic} type & 3 & 0.59 & 0.94 & UNINA \\
	    & Warsaw Univ. of Technology, PL & & & unknown {fake} / {authentic} type & 3 & 23.80 & 3.64 & anonymous \\
	    &Univ. of Notre Dame, USA  & & & cross-sensor & 3 & 3.36 & 14.71 & anonymous \\
	    & West Virginia Univ., USA & & & & & & & \\
	    & IIITD Delhi, India & & & & & & & \\\hline
	    MobILive 2014 \cite{Sequeira_IJCB_2014}  & INESC TEC, PT & PP & VIS & known {fake} / {authentic} type & 6 & 0.5 & 0.0 & IIT Indore \\
	    & Univ. of Porto, PT & & & & & & & \\
	    \hline
        \end{tabular}
        \begin{tablenotes}
        \tiny
        \item
        \item {\bf Notes:}
        \item[1] based on {average of APCER and BPCER} calculated on all datasets in a given evaluation category
		\item
		\item
		\item
		\item
		\hrulefill
		\label{tabs:explanation}
		\item {\normalsize Explanation of abbreviations used in the tables \ref{tab:BenchmarkDatabases} and \ref{tab:Competitions}}
		\item
		\item {\bf Type of samples:} 
		\item ~~~PP -- live + paper printouts; CL -- live + textured contact lenses; PE -- live + prosthetic eyes; SY -- live + synthetic irises; 
		\item ~~~RA -- live + replay attack; PD -- pupil dynamics; EM -- eye movement tracking; EG -- eye gaze video; PM -- post-mortem (cadaver) iris
		\item
      		\item {\bf Wavelength:} 
      		\item ~~~NIR -- Near-Infrared light; VIS -- visible light
		\item
		\item {\bf Sensor(s) used:}
		\item\noindent%
		\begin{minipage}{0.65\textwidth}
		\item Commercial iris recognition sensors:
		\item A -- IrisGuard AD100
		\item H -- IrisGuard H100
		\item C -- Cogent CIS 202
		\item L2 -- LG 2200
		\item L3 -- LG Iris Access EOU3000		
		\item L4 -- LG 4000
		\item V -- Vista Imaging VistaFA2E
		\item BM -- CMTech BMT-20
		\item IS -- IriTech IriShield M2120U
		\item 
		\item Commercial eye trackers:
		\item TX -- Tobi TX300 binocular eye tracker (300 Hz)
		\item EL -- EyeLink 1000 monocular eye tracker (1000 Hz)	
		\item 
	    \item Prototype iris recognition sensors:
		\item P$^{\mbox{\tiny WUT-1}}$ -- Aritech ARX-3M3C (SONY EX-View CCD), Fujinon DV10X7.5A-SA2, B+W 092 NIR filter
		\item P$^{\mbox{\tiny WUT-2}}$ -- DMK 4002-IR (SONY ICX249AL CCD), B+W 092 NIR filter
		\end{minipage}
		\begin{minipage}{0.6\textwidth}
		\item Non-biometric (general-purpose) equipment:
		\item LY -- Lytro Light Field Camera		
		\item IP -- iPhone 5S
		\item NL -- Nokia Lumia 1020
		\item GS -- Galaxy Samsung IV
		\item GT -- Galaxy Tablet II
		\item DA -- Dalsa (unknown model)
		\item CN3 -- Canon EOS Rebel T3i with EF-S 18-135 mm IS f/3.5-5.6 zoom lens
		\item CN5 -- Canon 550 D
		\item CN6 -- Canon 60 D		
		\item NK -- Nikon D 800 with 20-300 mm lens
		\item PS -- PlayStation eye camera (75 Hz)
		\item AT -- back 8MP camera in Asus Transformer Pad TF 300T 
		\item HP -- HP flatbed optical scanner
		\item KM -- Konica Minolta Bizhub C454E
		\item TG3 -- Olympus TG-3
		\end{minipage}
		\end{tablenotes}
	\end{threeparttable}
    \end{center}
\end{table}

\end{landscape}

\bibliographystyle{ACM-Reference-Format}
\bibliography{iris-pad-survey}


\begin{thebibliography}{00}


\ifx \showCODEN    \undefined \def \showCODEN     #1{\unskip}     \fi
\ifx \showDOI      \undefined \def \showDOI       #1{#1}\fi
\ifx \showISBNx    \undefined \def \showISBNx     #1{\unskip}     \fi
\ifx \showISBNxiii \undefined \def \showISBNxiii  #1{\unskip}     \fi
\ifx \showISSN     \undefined \def \showISSN      #1{\unskip}     \fi
\ifx \showLCCN     \undefined \def \showLCCN      #1{\unskip}     \fi
\ifx \shownote     \undefined \def \shownote      #1{#1}          \fi
\ifx \showarticletitle \undefined \def \showarticletitle #1{#1}   \fi
\ifx \showURL      \undefined \def \showURL       {\relax}        \fi
\providecommand\bibfield[2]{#2}
\providecommand\bibinfo[2]{#2}
\providecommand\natexlab[1]{#1}
\providecommand\showeprint[2][]{arXiv:#2}

\bibitem[\protect\citeauthoryear{Adamiak, \.{Z}urek, and \'{S}lot}{Adamiak
  et~al\mbox{.}}{2015}]%
        {Adamiak_FedCSIS_2015}
\bibfield{author}{\bibinfo{person}{Krzysztof Adamiak}, \bibinfo{person}{Dominik
  \.{Z}urek}, {and} \bibinfo{person}{Krzysztof \'{S}lot}.}
  \bibinfo{year}{2015}\natexlab{}.
\newblock \showarticletitle{Liveness detection in remote biometrics based on
  gaze direction estimation}. In \bibinfo{booktitle}{{\em Federated Conf. on
  Computer Science and Information Systems (FedCSIS)}}.
  \bibinfo{publisher}{IEEE}, \bibinfo{address}{\L\'{o}d\'{z}, Poland},
  \bibinfo{pages}{225--230}.
\newblock
\showDOI{%
\url{https://doi.org/10.15439/2015F307}}


\bibitem[\protect\citeauthoryear{Akhtar, Michelon, and Foresti}{Akhtar
  et~al\mbox{.}}{2014}]%
        {Akhtar_ICCST_2014}
\bibfield{author}{\bibinfo{person}{Zahid Akhtar}, \bibinfo{person}{Christian
  Michelon}, {and} \bibinfo{person}{Gian~Luca Foresti}.}
  \bibinfo{year}{2014}\natexlab{}.
\newblock \showarticletitle{Liveness detection for biometric authentication in
  mobile applications}. In \bibinfo{booktitle}{{\em {IEEE} Int. Carnahan Conf.
  on Security Technology (ICCST)}}. \bibinfo{publisher}{IEEE},
  \bibinfo{address}{Rome, Italy}, \bibinfo{pages}{1--6}.
\newblock
\showISSN{1071-6572}
\showDOI{%
\url{https://doi.org/10.1109/CCST.2014.6986982}}


\bibitem[\protect\citeauthoryear{Akhtar, Micheloni, and Foresti}{Akhtar
  et~al\mbox{.}}{2015}]%
        {Akhtar_SECURITY_2015}
\bibfield{author}{\bibinfo{person}{Zahid Akhtar}, \bibinfo{person}{Christian
  Micheloni}, {and} \bibinfo{person}{Gian~Luca Foresti}.}
  \bibinfo{year}{2015}\natexlab{}.
\newblock \showarticletitle{Biometric Liveness Detection: Challenges and
  Research Opportunities}.
\newblock \bibinfo{journal}{{\em IEEE Security Privacy\/}}
  \bibinfo{volume}{13}, \bibinfo{number}{5} (\bibinfo{date}{Sept}
  \bibinfo{year}{2015}), \bibinfo{pages}{63--72}.
\newblock
\showISSN{1540-7993}
\showDOI{%
\url{https://doi.org/10.1109/MSP.2015.116}}


\bibitem[\protect\citeauthoryear{Akhtar, Micheloni, Piciarelli, and
  Foresti}{Akhtar et~al\mbox{.}}{2014}]%
        {Akhtar_AVSS_2014}
\bibfield{author}{\bibinfo{person}{Zahid Akhtar}, \bibinfo{person}{Christian
  Micheloni}, \bibinfo{person}{Claudio Piciarelli}, {and}
  \bibinfo{person}{Gian~Luca Foresti}.} \bibinfo{year}{2014}\natexlab{}.
\newblock \showarticletitle{MoBio\_LivDet: Mobile biometric liveness
  detection}. In \bibinfo{booktitle}{{\em {IEEE} Int. Conf. on Advanced Video
  and Signal Based Surveillance (AVSS)}}. \bibinfo{publisher}{IEEE},
  \bibinfo{address}{Seoul, South Korea}, \bibinfo{pages}{187--192}.
\newblock
\showDOI{%
\url{https://doi.org/10.1109/AVSS.2014.6918666}}


\bibitem[\protect\citeauthoryear{Al-Raisi and Al-Khouri}{Al-Raisi and
  Al-Khouri}{2008}]%
        {Al-Raisi_TI_2008}
\bibfield{author}{\bibinfo{person}{Ahmad~N. Al-Raisi} {and}
  \bibinfo{person}{Ali~M. Al-Khouri}.} \bibinfo{year}{2008}\natexlab{}.
\newblock \showarticletitle{Iris recognition and the challenge of homeland and
  border control security in UAE}.
\newblock \bibinfo{journal}{{\em Telematics and Informatics\/}}
  \bibinfo{volume}{25}, \bibinfo{number}{2} (\bibinfo{year}{2008}),
  \bibinfo{pages}{117 -- 132}.
\newblock
\showISSN{0736-5853}
\showDOI{%
\url{https://doi.org/10.1016/j.tele.2006.06.005}}


\bibitem[\protect\citeauthoryear{Alonso-Fernandez and
  Big\"{u}n}{Alonso-Fernandez and Big\"{u}n}{2014}]%
        {Alonso-Fernandez_MIPRO_2014}
\bibfield{author}{\bibinfo{person}{Fernando Alonso-Fernandez} {and}
  \bibinfo{person}{Josef Big\"{u}n}.} \bibinfo{year}{2014}\natexlab{}.
\newblock \showarticletitle{Exploiting periocular and RGB information in fake
  iris detection}. In \bibinfo{booktitle}{{\em Int. Convention on Information
  and Communication Technology, Electronics and Microelectronics (MIPRO)}}.
  \bibinfo{publisher}{IEEE}, \bibinfo{address}{Opatija, Croatia},
  \bibinfo{pages}{1354--1359}.
\newblock
\showDOI{%
\url{https://doi.org/10.1109/MIPRO.2014.6859778}}


\bibitem[\protect\citeauthoryear{Anjos, El-Shafey, and Marcel}{Anjos
  et~al\mbox{.}}{2017}]%
        {Anjos_arXiv_2017}
\bibfield{author}{\bibinfo{person}{Andr\'{e} Anjos}, \bibinfo{person}{Laurent
  El-Shafey}, {and} \bibinfo{person}{S\'{e}bastien Marcel}.}
  \bibinfo{year}{2017}\natexlab{}.
\newblock \showarticletitle{{BEAT: An Open-Source Web-Based Open-Science
  Platform}}.
\newblock  (\bibinfo{year}{2017}).
\newblock
\showeprint{cs.SE/1704.02319. Retrieved from https://arxiv.org/pdf/1704.02319}


\bibitem[\protect\citeauthoryear{Baker, Hentz, Bowyer, and Flynn}{Baker
  et~al\mbox{.}}{2009}]%
        {Baker_BTAS_2009}
\bibfield{author}{\bibinfo{person}{Sarah~E. Baker}, \bibinfo{person}{Amanda
  Hentz}, \bibinfo{person}{Kevin~W. Bowyer}, {and} \bibinfo{person}{Patrick~J.
  Flynn}.} \bibinfo{year}{2009}\natexlab{}.
\newblock \showarticletitle{Contact lenses: Handle with care for iris
  recognition}. In \bibinfo{booktitle}{{\em {IEEE} Int. Conf. on Biometrics:
  Theory Applications and Systems (BTAS)}}. \bibinfo{publisher}{IEEE},
  \bibinfo{address}{Washington, DC, USA}, \bibinfo{pages}{1--8}.
\newblock
\showDOI{%
\url{https://doi.org/10.1109/BTAS.2009.5339050}}


\bibitem[\protect\citeauthoryear{Bhogal, S\"{o}llinger, Trung, and Uhl}{Bhogal
  et~al\mbox{.}}{2017}]%
        {Bhogal_BF_2017}
\bibfield{author}{\bibinfo{person}{Amrit Pal~Singh Bhogal},
  \bibinfo{person}{Dominik S\"{o}llinger}, \bibinfo{person}{Pauline Trung},
  {and} \bibinfo{person}{Andreas Uhl}.} \bibinfo{year}{2017}\natexlab{}.
\newblock \showarticletitle{Non-reference image quality assessment for
  biometric presentation attack detection}. In \bibinfo{booktitle}{{\em Int.
  Workshop on Biometrics and Forensics}}. \bibinfo{publisher}{IEEE},
  \bibinfo{address}{Coventry, UK}, \bibinfo{pages}{1--6}.
\newblock
\showDOI{%
\url{https://doi.org/10.1109/IWBF.2017.7935080}}


\bibitem[\protect\citeauthoryear{{BIPLab -- Biometric and Image Processing Lab,
  University of Salerno}}{{BIPLab -- Biometric and Image Processing Lab,
  University of Salerno}}{2014}]%
        {MICHE_I_DB_URL}
\bibfield{author}{\bibinfo{person}{{BIPLab -- Biometric and Image Processing
  Lab, University of Salerno}}.} \bibinfo{year}{2014}\natexlab{}.
\newblock \bibinfo{title}{{Mobile Iris CHallenge Evaluation I (MICHE-I)
  Database}}.
\newblock   (\bibinfo{date}{March} \bibinfo{year}{2014}).
\newblock
\showURL{%
Retrieved Retrieved August 8, 2017 from
  \url{http://biplab.unisa.it/MICHE/database/}}


\bibitem[\protect\citeauthoryear{Bolme, Tokola, Boehnen, Saul, Sauerwein, and
  Steadman}{Bolme et~al\mbox{.}}{2016}]%
        {Bolme_BTAS_2016}
\bibfield{author}{\bibinfo{person}{David~S. Bolme}, \bibinfo{person}{Ryan~A.
  Tokola}, \bibinfo{person}{Chris~B. Boehnen}, \bibinfo{person}{Tiffany~B.
  Saul}, \bibinfo{person}{Kelly~A. Sauerwein}, {and}
  \bibinfo{person}{Dawnie~Wolfe Steadman}.} \bibinfo{year}{2016}\natexlab{}.
\newblock \showarticletitle{Impact of environmental factors on biometric
  matching during human decomposition}. In \bibinfo{booktitle}{{\em {IEEE} Int.
  Conf. on Biometrics: Theory Applications and Systems (BTAS)}}.
  \bibinfo{publisher}{IEEE}, \bibinfo{address}{Niagara Falls, NY, USA},
  \bibinfo{pages}{1--8}.
\newblock
\showDOI{%
\url{https://doi.org/10.1109/BTAS.2016.7791177}}


\bibitem[\protect\citeauthoryear{Boser, Guyon, and Vapnik}{Boser
  et~al\mbox{.}}{1992}]%
        {Boser_CLT_1992}
\bibfield{author}{\bibinfo{person}{Bernhard~E. Boser},
  \bibinfo{person}{Isabelle~M. Guyon}, {and} \bibinfo{person}{Vladimir~N.
  Vapnik}.} \bibinfo{year}{1992}\natexlab{}.
\newblock \showarticletitle{A Training Algorithm for Optimal Margin
  Classifiers}. In \bibinfo{booktitle}{{\em Proceedings of the Fifth Annual
  Workshop on Computational Learning Theory}} {\em
  (\bibinfo{series}{COLT'92})}. \bibinfo{publisher}{ACM}, \bibinfo{address}{New
  York, NY, USA}, \bibinfo{pages}{144--152}.
\newblock
\showISBNx{0-89791-497-X}
\showDOI{%
\url{https://doi.org/10.1145/130385.130401}}


\bibitem[\protect\citeauthoryear{Bowyer and Doyle}{Bowyer and Doyle}{2014}]%
        {Bowyer_COMPUTER_2014}
\bibfield{author}{\bibinfo{person}{Kevin~W. Bowyer} {and}
  \bibinfo{person}{James~S. Doyle}.} \bibinfo{year}{2014}\natexlab{}.
\newblock \showarticletitle{Cosmetic Contact Lenses and Iris Recognition
  Spoofing}.
\newblock \bibinfo{journal}{{\em {IEEE} Computer\/}} \bibinfo{volume}{47},
  \bibinfo{number}{5} (\bibinfo{date}{May} \bibinfo{year}{2014}),
  \bibinfo{pages}{96--98}.
\newblock
\showISSN{0018-9162}
\showDOI{%
\url{https://doi.org/10.1109/MC.2014.118}}


\bibitem[\protect\citeauthoryear{Chen and Ross}{Chen and Ross}{2018}]%
        {Chen_WACV_2018}
\bibfield{author}{\bibinfo{person}{Cunjian Chen} {and} \bibinfo{person}{Arun
  Ross}.} \bibinfo{year}{2018}\natexlab{}.
\newblock \showarticletitle{A Multi-task Convolutional Neural Network for Joint
  Iris Detection and Presentation Attack Detection}. In
  \bibinfo{booktitle}{{\em {IEEE} Winter Conf. on Applications of Computer
  Vision (WACV)}}. \bibinfo{pages}{44--51}.
\newblock
\showDOI{%
\url{https://doi.org/10.1109/WACVW.2018.00011}}


\bibitem[\protect\citeauthoryear{Chen, Lin, and Ding}{Chen
  et~al\mbox{.}}{2012}]%
        {Chen_PRL_2012}
\bibfield{author}{\bibinfo{person}{Rui Chen}, \bibinfo{person}{Xirong Lin},
  {and} \bibinfo{person}{Tianhuai Ding}.} \bibinfo{year}{2012}\natexlab{}.
\newblock \showarticletitle{Liveness detection for iris recognition using
  multispectral images}.
\newblock \bibinfo{journal}{{\em Pattern Recognition Letters\/}}
  \bibinfo{volume}{33}, \bibinfo{number}{12} (\bibinfo{year}{2012}),
  \bibinfo{pages}{1513 -- 1519}.
\newblock
\showISSN{0167-8655}
\showDOI{%
\url{https://doi.org/10.1016/j.patrec.2012.04.002}}


\bibitem[\protect\citeauthoryear{{Chinese Academy of Sciences}}{{Chinese
  Academy of Sciences}}{2004}]%
        {CASIA_IrisSynv4_URL}
\bibfield{author}{\bibinfo{person}{{Chinese Academy of Sciences}}.}
  \bibinfo{year}{2004}\natexlab{}.
\newblock \bibinfo{title}{{CASIA-Iris-Syn v4}}.
\newblock   (\bibinfo{year}{2004}).
\newblock
\showURL{%
Retrieved March 21, 2018 from
  \url{http://biometrics.idealtest.org/dbDetailForUser.do?id=4}}


\bibitem[\protect\citeauthoryear{{CITeR}}{{CITeR}}{2006}]%
        {WVU_DB1_URL}
\bibfield{author}{\bibinfo{person}{{CITeR}}.} \bibinfo{year}{2006}\natexlab{}.
\newblock \bibinfo{title}{{Synthetic Iris Textured Based}}.
\newblock   (\bibinfo{year}{2006}).
\newblock
\showURL{%
Retrieved March 21, 2018 from
  \url{https://citer.clarkson.edu/biometric-dataset-collections/synthetic-iris-textured-based/}}


\bibitem[\protect\citeauthoryear{{CITeR}}{{CITeR}}{2007}]%
        {WVU_DB2_URL}
\bibfield{author}{\bibinfo{person}{{CITeR}}.} \bibinfo{year}{2007}\natexlab{}.
\newblock \bibinfo{title}{{Synthetic Iris Model Based}}.
\newblock   (\bibinfo{year}{2007}).
\newblock
\showURL{%
Retrieved March 21, 2018 from
  \url{https://citer.clarkson.edu/biometric-dataset-collections/synthetic-iris-model-based/}}


\bibitem[\protect\citeauthoryear{Connell, Ratha, Gentile, and Bolle}{Connell
  et~al\mbox{.}}{2013}]%
        {Connell_ASSP_2013}
\bibfield{author}{\bibinfo{person}{Jonathan Connell}, \bibinfo{person}{Nalini
  Ratha}, \bibinfo{person}{James Gentile}, {and} \bibinfo{person}{Ruud Bolle}.}
  \bibinfo{year}{2013}\natexlab{}.
\newblock \showarticletitle{Fake iris detection using structured light}. In
  \bibinfo{booktitle}{{\em {IEEE} Int. Conf. on Acoustics, Speech and Signal
  Processing}}. \bibinfo{publisher}{IEEE}, \bibinfo{address}{Vancouver, BC,
  Canada}, \bibinfo{pages}{8692--8696}.
\newblock
\showISSN{1520-6149}
\showDOI{%
\url{https://doi.org/10.1109/ICASSP.2013.6639363}}


\bibitem[\protect\citeauthoryear{Czajka}{Czajka}{2013}]%
        {Czajka_ICMMAR_2013}
\bibfield{author}{\bibinfo{person}{Adam Czajka}.}
  \bibinfo{year}{2013}\natexlab{}.
\newblock \showarticletitle{Database of iris printouts and its application:
  Development of liveness detection method for iris recognition}. In
  \bibinfo{booktitle}{{\em {IEEE} Int. Conf. on Methods and Models in
  Automation and Robotics (MMAR)}}. \bibinfo{publisher}{IEEE},
  \bibinfo{address}{Mi\c{e}dzyzdroje, Poland}, \bibinfo{pages}{28--33}.
\newblock
\showDOI{%
\url{https://doi.org/10.1109/MMAR.2013.6669876}}


\bibitem[\protect\citeauthoryear{Czajka}{Czajka}{2015}]%
        {Czajka_TIFS_2015}
\bibfield{author}{\bibinfo{person}{Adam Czajka}.}
  \bibinfo{year}{2015}\natexlab{}.
\newblock \showarticletitle{Pupil Dynamics for Iris Liveness Detection}.
\newblock \bibinfo{journal}{{\em {IEEE} Trans. Inf. Forens. Security\/}}
  \bibinfo{volume}{10}, \bibinfo{number}{4} (\bibinfo{date}{April}
  \bibinfo{year}{2015}), \bibinfo{pages}{726--735}.
\newblock
\showISSN{1556-6013}
\showDOI{%
\url{https://doi.org/10.1109/TIFS.2015.2398815}}


\bibitem[\protect\citeauthoryear{Czajka}{Czajka}{2016}]%
        {Czajka_Handbook_2016}
\bibfield{author}{\bibinfo{person}{Adam Czajka}.}
  \bibinfo{year}{2016}\natexlab{}.
\newblock \showarticletitle{Iris Liveness Detection by Modeling Dynamic Pupil
  Features}.
\newblock In \bibinfo{booktitle}{{\em Handbook of Iris Recognition}},
  \bibfield{editor}{\bibinfo{person}{Kevin~W. Bowyer} {and}
  \bibinfo{person}{Mark~J. Burge}} (Eds.). \bibinfo{publisher}{Springer
  London}, \bibinfo{address}{London}, \bibinfo{pages}{439--467}.
\newblock
\showISBNx{978-1-4471-6784-6}
\showDOI{%
\url{https://doi.org/10.1007/978-1-4471-6784-6_19}}


\bibitem[\protect\citeauthoryear{Czajka and Becker}{Czajka and Becker}{2018}]%
        {Czajka_PAD_Handbook_2018}
\bibfield{author}{\bibinfo{person}{Adam Czajka} {and} \bibinfo{person}{Benedict
  Becker}.} \bibinfo{year}{2018}\natexlab{}.
\newblock \showarticletitle{{Application of Dynamic Features of the Pupil for
  Iris Presentation Attack Detection}}.
\newblock In \bibinfo{booktitle}{{\em Handbook of Biometric Anti-Spoofing (2nd
  Edition, to appear)}}, \bibfield{editor}{\bibinfo{person}{S\'{e}bastien
  Marcel}, \bibinfo{person}{Mark Nixon}, \bibinfo{person}{Julian Fierrez},
  {and} \bibinfo{person}{Nicholas Evans}} (Eds.). \bibinfo{publisher}{Springer
  Int. Publishing AG}, \bibinfo{pages}{1--17}.
\newblock


\bibitem[\protect\citeauthoryear{Czajka, Bowyer, Krumdick, and
  VidalMata}{Czajka et~al\mbox{.}}{2017}]%
        {Czajka_TIFS_2017}
\bibfield{author}{\bibinfo{person}{Adam Czajka}, \bibinfo{person}{Kevin~W.
  Bowyer}, \bibinfo{person}{Michael Krumdick}, {and}
  \bibinfo{person}{Rosaura~G. VidalMata}.} \bibinfo{year}{2017}\natexlab{}.
\newblock \showarticletitle{Recognition of image-orientation-based iris
  spoofing}.
\newblock \bibinfo{journal}{{\em {IEEE} Trans. Inf. Forens. Security\/}}
  \bibinfo{volume}{PP}, \bibinfo{number}{99} (\bibinfo{year}{2017}),
  \bibinfo{pages}{1--1}.
\newblock
\showISSN{1556-6013}
\showDOI{%
\url{https://doi.org/10.1109/TIFS.2017.2701332}}


\bibitem[\protect\citeauthoryear{Czajka, Pacut, and Chochowski}{Czajka
  et~al\mbox{.}}{2011}]%
        {Czajka_patent_2011}
\bibfield{author}{\bibinfo{person}{Adam Czajka}, \bibinfo{person}{Andrzej
  Pacut}, {and} \bibinfo{person}{Marcin Chochowski}.}
  \bibinfo{year}{2011}\natexlab{}.
\newblock \bibinfo{title}{Method of eye aliveness testing and device for eye
  aliveness testing, {U}nited {S}tates {P}atent, {US} 8,061,842}.
\newblock   (\bibinfo{year}{2011}).
\newblock


\bibitem[\protect\citeauthoryear{Czajka, Strzelczyk, and Pacut}{Czajka
  et~al\mbox{.}}{2007}]%
        {Czajka_SPIE_2007}
\bibfield{author}{\bibinfo{person}{Adam Czajka}, \bibinfo{person}{Przemek
  Strzelczyk}, {and} \bibinfo{person}{Andrzej Pacut}.}
  \bibinfo{year}{2007}\natexlab{}.
\newblock \showarticletitle{Making iris recognition more reliable and spoof
  resistant}.
\newblock \bibinfo{journal}{{\em SPIE Newsroom\/}} (\bibinfo{date}{June}
  \bibinfo{year}{2007}).
\newblock
\showDOI{%
\url{https://doi.org/10.1117/2.1200706.0614}}


\bibitem[\protect\citeauthoryear{Das, Pal, Ferrer, and Blumenstein}{Das
  et~al\mbox{.}}{2016}]%
        {Das_PRL_2016}
\bibfield{author}{\bibinfo{person}{Abhijit Das}, \bibinfo{person}{Umapada Pal},
  \bibinfo{person}{Miguel~Angel Ferrer}, {and} \bibinfo{person}{Michael
  Blumenstein}.} \bibinfo{year}{2016}\natexlab{}.
\newblock \showarticletitle{A framework for liveness detection for direct
  attacks in the visible spectrum for multimodal ocular biometrics}.
\newblock \bibinfo{journal}{{\em Pattern Recognition Letters\/}}
  \bibinfo{volume}{82} (\bibinfo{year}{2016}), \bibinfo{pages}{232 -- 241}.
\newblock
\showISSN{0167-8655}
\showDOI{%
\url{https://doi.org/10.1016/j.patrec.2015.11.016}}
\newblock
\shownote{An insight on eye biometrics.}


\bibitem[\protect\citeauthoryear{Daugman}{Daugman}{1994}]%
        {Daugman_patent_1994}
\bibfield{author}{\bibinfo{person}{John Daugman}.}
  \bibinfo{year}{1994}\natexlab{}.
\newblock \bibinfo{title}{Biometric personal identification system based on
  iris analysis, {U}nited {S}tates {P}atent, {US} 5,291,560}.
\newblock   (\bibinfo{date}{11 July} \bibinfo{year}{1994}).
\newblock


\bibitem[\protect\citeauthoryear{Daugman}{Daugman}{2000}]%
        {Daugman_IMAIP_2000}
\bibfield{author}{\bibinfo{person}{John Daugman}.}
  \bibinfo{year}{2000}\natexlab{}.
\newblock \showarticletitle{Wavelet Demodulation Codes, Statistical
  Independence, and Pattern Recognition}. In \bibinfo{booktitle}{{\em Institute
  of Mathematics and its Applications, Proc. 2nd IMA-IP}}.
  \bibinfo{publisher}{IMA}, \bibinfo{address}{Horwood, London, UK},
  \bibinfo{pages}{244--260}.
\newblock


\bibitem[\protect\citeauthoryear{Daugman}{Daugman}{2003}]%
        {Daugman_WMIP_2003}
\bibfield{author}{\bibinfo{person}{John Daugman}.}
  \bibinfo{year}{2003}\natexlab{}.
\newblock \showarticletitle{Demodulation By Complex-Valued Wavelets For
  Stochastic Pattern Recognition}.
\newblock \bibinfo{journal}{{\em Int. Journal of Wavelets, Multi-resolution and
  Information Processing\/}}  \bibinfo{volume}{1} (\bibinfo{year}{2003}),
  \bibinfo{pages}{1--17}.
\newblock


\bibitem[\protect\citeauthoryear{Doyle and Bowyer}{Doyle and Bowyer}{2015}]%
        {Doyle_IEEEAccess_2015}
\bibfield{author}{\bibinfo{person}{James~S. Doyle} {and}
  \bibinfo{person}{Kevin~W. Bowyer}.} \bibinfo{year}{2015}\natexlab{}.
\newblock \showarticletitle{Robust Detection of Textured Contact Lenses in Iris
  Recognition Using BSIF}.
\newblock \bibinfo{journal}{{\em {IEEE} Access\/}}  \bibinfo{volume}{3}
  (\bibinfo{year}{2015}), \bibinfo{pages}{1672--1683}.
\newblock
\showISSN{2169-3536}
\showDOI{%
\url{https://doi.org/10.1109/ACCESS.2015.2477470}}


\bibitem[\protect\citeauthoryear{Doyle, Bowyer, and Flynn}{Doyle
  et~al\mbox{.}}{2013a}]%
        {Doyle_BTAS_2013}
\bibfield{author}{\bibinfo{person}{James~S. Doyle}, \bibinfo{person}{Kevin~W.
  Bowyer}, {and} \bibinfo{person}{Patrick~J. Flynn}.}
  \bibinfo{year}{2013}\natexlab{a}.
\newblock \showarticletitle{Variation in accuracy of textured contact lens
  detection based on sensor and lens pattern}. In \bibinfo{booktitle}{{\em
  {IEEE} Int. Conf. on Biometrics: Theory Applications and Systems (BTAS)}}.
  \bibinfo{publisher}{IEEE}, \bibinfo{address}{Arlington, VA, USA},
  \bibinfo{pages}{1--7}.
\newblock
\showDOI{%
\url{https://doi.org/10.1109/BTAS.2013.6712745}}


\bibitem[\protect\citeauthoryear{Doyle, Flynn, and Bowyer}{Doyle
  et~al\mbox{.}}{2013b}]%
        {Doyle_ICB_2013}
\bibfield{author}{\bibinfo{person}{James~S. Doyle}, \bibinfo{person}{Patrick~J.
  Flynn}, {and} \bibinfo{person}{Kevin~W. Bowyer}.}
  \bibinfo{year}{2013}\natexlab{b}.
\newblock \showarticletitle{Automated classification of contact lens type in
  iris images}. In \bibinfo{booktitle}{{\em 2013 Int. Conf. on Biometrics
  (ICB)}}. \bibinfo{publisher}{IEEE}, \bibinfo{address}{Madrid, Spain},
  \bibinfo{pages}{1--6}.
\newblock
\showISSN{2376-4201}
\showDOI{%
\url{https://doi.org/10.1109/ICB.2013.6612954}}


\bibitem[\protect\citeauthoryear{Drozdowski, Rathgeb, and Busch}{Drozdowski
  et~al\mbox{.}}{2017}]%
        {Drozdowski_BIOSIG_2017}
\bibfield{author}{\bibinfo{person}{Pawel Drozdowski},
  \bibinfo{person}{Christian Rathgeb}, {and} \bibinfo{person}{Christoph
  Busch}.} \bibinfo{year}{2017}\natexlab{}.
\newblock \showarticletitle{{Sic-Gen: A Synthetic Iris-Code Generator}}. In
  \bibinfo{booktitle}{{\em Int. Conf. of the Biometrics Special Interest Group
  (BIOSIG)}}. \bibinfo{publisher}{IEEE}, \bibinfo{address}{Darmstadt, Germany},
  \bibinfo{pages}{1--6}.
\newblock
\showDOI{%
\url{https://doi.org/10.23919/BIOSIG.2017.8053520}}


\bibitem[\protect\citeauthoryear{Dunstone and Poulton}{Dunstone and
  Poulton}{2011}]%
        {Dunstone_BTT_2011}
\bibfield{author}{\bibinfo{person}{Ted Dunstone} {and} \bibinfo{person}{Geoff
  Poulton}.} \bibinfo{year}{2011}\natexlab{}.
\newblock \showarticletitle{Vulnerability assessment}.
\newblock \bibinfo{journal}{{\em Biometric Technology Today\/}}
  \bibinfo{volume}{2011} (\bibinfo{date}{May} \bibinfo{year}{2011}),
  \bibinfo{pages}{5--7}.
\newblock
Issue 5.


\bibitem[\protect\citeauthoryear{E.Baker, Hentz, W.Bowyer, and J.Flynn}{E.Baker
  et~al\mbox{.}}{2010}]%
        {Baker_CVIU_2010}
\bibfield{author}{\bibinfo{person}{Sarah E.Baker}, \bibinfo{person}{Amanda
  Hentz}, \bibinfo{person}{Kevin W.Bowyer}, {and} \bibinfo{person}{Patrick
  J.Flynn}.} \bibinfo{year}{2010}\natexlab{}.
\newblock \showarticletitle{Degradation of Iris Recognition Performance Due to
  Non-Cosmetic Prescription Contact Lenses}.
\newblock \bibinfo{journal}{{\em Computer Vision and Image Understanding\/}}
  \bibinfo{volume}{114}, \bibinfo{number}{9} (\bibinfo{date}{September}
  \bibinfo{year}{2010}), \bibinfo{pages}{1030--1044}.
\newblock
\showDOI{%
\url{https://doi.org/10.1016/j.cviu.2010.06.002}}


\bibitem[\protect\citeauthoryear{Fathy and Ali}{Fathy and Ali}{2017}]%
        {Fathy_WPC_2017}
\bibfield{author}{\bibinfo{person}{Waleed S.-A. Fathy} {and}
  \bibinfo{person}{Hanaa~S. Ali}.} \bibinfo{year}{2017}\natexlab{}.
\newblock \showarticletitle{Entropy with Local Binary Patterns for Efficient
  Iris Liveness Detection}.
\newblock \bibinfo{journal}{{\em Wireless Personal Communications\/}}
  (\bibinfo{date}{04 Dec} \bibinfo{year}{2017}), \bibinfo{pages}{1--14}.
\newblock
\showISSN{1572-834X}
\showDOI{%
\url{https://doi.org/10.1007/s11277-017-5089-z}}


\bibitem[\protect\citeauthoryear{Fathy, Ali, and Mahmoud}{Fathy
  et~al\mbox{.}}{2017}]%
        {Fathy_NRSC_2017}
\bibfield{author}{\bibinfo{person}{Waleed S-A. Fathy},
  \bibinfo{person}{Hanaa~S. Ali}, {and} \bibinfo{person}{Imbaby~I. Mahmoud}.}
  \bibinfo{year}{2017}\natexlab{}.
\newblock \showarticletitle{Statistical representation for iris anti-spoofing
  using wavelet-based feature extraction and selection algorithms}. In
  \bibinfo{booktitle}{{\em National Radio Science Conf. (NRSC)}}.
  \bibinfo{publisher}{IEEE}, \bibinfo{address}{Alexandria, Egypt},
  \bibinfo{pages}{221--229}.
\newblock
\showDOI{%
\url{https://doi.org/10.1109/NRSC.2017.7893480}}


\bibitem[\protect\citeauthoryear{Galbally, Fierrez, and Ortega-Garcia}{Galbally
  et~al\mbox{.}}{2007}]%
        {Galbally_SWB_2007}
\bibfield{author}{\bibinfo{person}{Javier Galbally}, \bibinfo{person}{Julian
  Fierrez}, {and} \bibinfo{person}{Javier Ortega-Garcia}.}
  \bibinfo{year}{2007}\natexlab{}.
\newblock \showarticletitle{Vulnerabilities in Biometric Systems: Attacks and
  Recent Advances in Liveness Detection}. In \bibinfo{booktitle}{{\em Spanish
  Workshop on Biometrics (SWB)}}. \bibinfo{publisher}{Springer},
  \bibinfo{address}{Girona, Spain}, \bibinfo{pages}{1--8}.
\newblock


\bibitem[\protect\citeauthoryear{Galbally and Gomez-Barrero}{Galbally and
  Gomez-Barrero}{2016}]%
        {Galbally_IWBF_2016}
\bibfield{author}{\bibinfo{person}{Javier Galbally} {and}
  \bibinfo{person}{Marta Gomez-Barrero}.} \bibinfo{year}{2016}\natexlab{}.
\newblock \showarticletitle{A review of iris anti-spoofing}. In
  \bibinfo{booktitle}{{\em Int. Conf. on Biometrics and Forensics (IWBF)}}.
  \bibinfo{publisher}{IEEE}, \bibinfo{address}{Limassol, Cyprus},
  \bibinfo{pages}{1--6}.
\newblock
\showDOI{%
\url{https://doi.org/10.1109/IWBF.2016.7449676}}


\bibitem[\protect\citeauthoryear{Galbally and Gomez-Barrero}{Galbally and
  Gomez-Barrero}{2017}]%
        {Galbally_IETbook_Ch11_2017}
\bibfield{author}{\bibinfo{person}{Javier Galbally} {and}
  \bibinfo{person}{Marta Gomez-Barrero}.} \bibinfo{year}{2017}\natexlab{}.
\newblock \showarticletitle{Presentation Attack Detection in Iris Recognition}.
\newblock In \bibinfo{booktitle}{{\em Iris and Periocular Biometric
  Recognition}}, \bibfield{editor}{\bibinfo{person}{Christian Rathgeb} {and}
  \bibinfo{person}{Christoph Busch}} (Eds.). \bibinfo{publisher}{IET},
  \bibinfo{address}{London, UK}, Chapter~11, \bibinfo{pages}{235--263}.
\newblock
\showISBNx{978-1-78561-168-1}
\showURL{%
\url{https://www.theiet.org/resources/books/security/irisper.cfm}}


\bibitem[\protect\citeauthoryear{Galbally, Marcel, and Fierrez}{Galbally
  et~al\mbox{.}}{2014}]%
        {Galbally_TIP_2014}
\bibfield{author}{\bibinfo{person}{Javier Galbally}, \bibinfo{person}{Sebastien
  Marcel}, {and} \bibinfo{person}{Julian Fierrez}.}
  \bibinfo{year}{2014}\natexlab{}.
\newblock \showarticletitle{Image Quality Assessment for Fake Biometric
  Detection: Application to Iris, Fingerprint, and Face Recognition}.
\newblock \bibinfo{journal}{{\em {IEEE} Trans. Image Process.\/}}
  \bibinfo{volume}{23}, \bibinfo{number}{2} (\bibinfo{date}{February}
  \bibinfo{year}{2014}), \bibinfo{pages}{710--724}.
\newblock
\showISSN{1057-7149}
\showDOI{%
\url{https://doi.org/10.1109/TIP.2013.2292332}}


\bibitem[\protect\citeauthoryear{Galbally, Ortiz-Lopez, Fierrez, and
  Ortega-Garcia}{Galbally et~al\mbox{.}}{2012}]%
        {Galbally_ICB_2012}
\bibfield{author}{\bibinfo{person}{Javier Galbally}, \bibinfo{person}{Jaime
  Ortiz-Lopez}, \bibinfo{person}{Julian Fierrez}, {and} \bibinfo{person}{Javier
  Ortega-Garcia}.} \bibinfo{year}{2012}\natexlab{}.
\newblock \showarticletitle{Iris liveness detection based on quality related
  features}. In \bibinfo{booktitle}{{\em 2012 5th IAPR Int. Conf. on Biometrics
  (ICB)}}. \bibinfo{publisher}{IEEE}, \bibinfo{address}{New Delhi, India},
  \bibinfo{pages}{271--276}.
\newblock
\showISSN{2376-4201}
\showDOI{%
\url{https://doi.org/10.1109/ICB.2012.6199819}}


\bibitem[\protect\citeauthoryear{Galbally, Ross, Gomez-Barrero, Fierrez, and
  Ortega-Garcia}{Galbally et~al\mbox{.}}{2013}]%
        {Galbally_CVIU_2013}
\bibfield{author}{\bibinfo{person}{Javier Galbally}, \bibinfo{person}{Arun
  Ross}, \bibinfo{person}{Marta Gomez-Barrero}, \bibinfo{person}{Julian
  Fierrez}, {and} \bibinfo{person}{Javier Ortega-Garcia}.}
  \bibinfo{year}{2013}\natexlab{}.
\newblock \showarticletitle{Iris image reconstruction from binary templates: An
  efficient probabilistic approach based on genetic algorithms}.
\newblock \bibinfo{journal}{{\em Computer Vision and Image Understanding\/}}
  \bibinfo{volume}{117}, \bibinfo{number}{10} (\bibinfo{year}{2013}),
  \bibinfo{pages}{1512 -- 1525}.
\newblock
\showISSN{1077-3142}
\showDOI{%
\url{https://doi.org/10.1016/j.cviu.2013.06.003}}


\bibitem[\protect\citeauthoryear{Galbally, Savvides, Venugopalan, and
  Ross}{Galbally et~al\mbox{.}}{2016}]%
        {Galbally_Handbook_2016}
\bibfield{author}{\bibinfo{person}{Javier Galbally}, \bibinfo{person}{Marios
  Savvides}, \bibinfo{person}{Shreyas Venugopalan}, {and}
  \bibinfo{person}{Arun~A. Ross}.} \bibinfo{year}{2016}\natexlab{}.
\newblock \showarticletitle{Iris Image Reconstruction from Binary Templates}.
\newblock In \bibinfo{booktitle}{{\em Handbook of Iris Recognition}},
  \bibfield{editor}{\bibinfo{person}{Kevin~W. Bowyer} {and}
  \bibinfo{person}{Mark~J. Burge}} (Eds.). \bibinfo{publisher}{Springer
  London}, \bibinfo{address}{London}, \bibinfo{pages}{469--496}.
\newblock
\showISBNx{978-1-4471-6784-6}
\showDOI{%
\url{https://doi.org/10.1007/978-1-4471-6784-6_20}}


\bibitem[\protect\citeauthoryear{Gragnaniello, Poggi, Sansone, and
  Verdoliva}{Gragnaniello et~al\mbox{.}}{2014}]%
        {Gragnaniello_SITIS_2014}
\bibfield{author}{\bibinfo{person}{Diego Gragnaniello},
  \bibinfo{person}{Giovanni Poggi}, \bibinfo{person}{Carlo Sansone}, {and}
  \bibinfo{person}{Luisa Verdoliva}.} \bibinfo{year}{2014}\natexlab{}.
\newblock \showarticletitle{Contact Lens Detection and Classification in Iris
  Images through Scale Invariant Descriptor}. In \bibinfo{booktitle}{{\em Int.
  Conf. on Signal-Image Technology Internet-Based Systems (SITIS)}}.
  \bibinfo{publisher}{IEEE}, \bibinfo{address}{Marrakech, Morocco},
  \bibinfo{pages}{560--565}.
\newblock
\showDOI{%
\url{https://doi.org/10.1109/SITIS.2014.35}}


\bibitem[\protect\citeauthoryear{Gragnaniello, Poggi, Sansone, and
  Verdoliva}{Gragnaniello et~al\mbox{.}}{2015}]%
        {Gragnaniello_TIFS_2015}
\bibfield{author}{\bibinfo{person}{Diego Gragnaniello},
  \bibinfo{person}{Giovanni Poggi}, \bibinfo{person}{Carlo Sansone}, {and}
  \bibinfo{person}{Luisa Verdoliva}.} \bibinfo{year}{2015}\natexlab{}.
\newblock \showarticletitle{An Investigation of Local Descriptors for Biometric
  Spoofing Detection}.
\newblock \bibinfo{journal}{{\em {IEEE} Trans. Inf. Forens. Security\/}}
  \bibinfo{volume}{10}, \bibinfo{number}{4} (\bibinfo{date}{April}
  \bibinfo{year}{2015}), \bibinfo{pages}{849--863}.
\newblock
\showISSN{1556-6013}
\showDOI{%
\url{https://doi.org/10.1109/TIFS.2015.2404294}}


\bibitem[\protect\citeauthoryear{Gragnaniello, Poggi, Sansone, and
  Verdoliva}{Gragnaniello et~al\mbox{.}}{2016}]%
        {Gragnaniello_PRL_2016}
\bibfield{author}{\bibinfo{person}{Diego Gragnaniello},
  \bibinfo{person}{Giovanni Poggi}, \bibinfo{person}{Carlo Sansone}, {and}
  \bibinfo{person}{Luisa Verdoliva}.} \bibinfo{year}{2016}\natexlab{}.
\newblock \showarticletitle{Using iris and sclera for detection and
  classification of contact lenses}.
\newblock \bibinfo{journal}{{\em Pattern Recognition Letters\/}}
  \bibinfo{volume}{82} (\bibinfo{year}{2016}), \bibinfo{pages}{251 -- 257}.
\newblock
\showISSN{0167-8655}
\showDOI{%
\url{https://doi.org/10.1016/j.patrec.2015.10.009}}
\newblock
\shownote{An insight on eye biometrics.}


\bibitem[\protect\citeauthoryear{Gragnaniello, Sansone, Poggi, and
  Verdoliva}{Gragnaniello et~al\mbox{.}}{2016}]%
        {Gragnaniello_SITIS_2016}
\bibfield{author}{\bibinfo{person}{Diego Gragnaniello}, \bibinfo{person}{Carlo
  Sansone}, \bibinfo{person}{Giovanni Poggi}, {and} \bibinfo{person}{Luisa
  Verdoliva}.} \bibinfo{year}{2016}\natexlab{}.
\newblock \showarticletitle{Biometric Spoofing Detection by a Domain-Aware
  Convolutional Neural Network}. In \bibinfo{booktitle}{{\em Int. Conf. on
  Signal-Image Technology Internet-Based Systems (SITIS)}}.
  \bibinfo{publisher}{IEEE}, \bibinfo{address}{Naples, Italy},
  \bibinfo{pages}{193--198}.
\newblock
\showDOI{%
\url{https://doi.org/10.1109/SITIS.2016.38}}


\bibitem[\protect\citeauthoryear{Gragnaniello, Sansone, and
  Verdoliva}{Gragnaniello et~al\mbox{.}}{2015}]%
        {Gragnaniell_PRL_2015}
\bibfield{author}{\bibinfo{person}{Diego Gragnaniello}, \bibinfo{person}{Carlo
  Sansone}, {and} \bibinfo{person}{Luisa Verdoliva}.}
  \bibinfo{year}{2015}\natexlab{}.
\newblock \showarticletitle{Iris liveness detection for mobile devices based on
  local descriptors}.
\newblock \bibinfo{journal}{{\em Pattern Recognition Letters\/}}
  \bibinfo{volume}{57} (\bibinfo{year}{2015}), \bibinfo{pages}{81 -- 87}.
\newblock
\showISSN{0167-8655}
\showDOI{%
\url{https://doi.org/10.1016/j.patrec.2014.10.018}}
\newblock
\shownote{Mobile Iris \{CHallenge\} Evaluation part I (MICHE I).}


\bibitem[\protect\citeauthoryear{Gupta, Behera, Vatsa, and Singh}{Gupta
  et~al\mbox{.}}{2014}]%
        {Gupta_ICPR_2014}
\bibfield{author}{\bibinfo{person}{Priyanshu Gupta}, \bibinfo{person}{Shipra
  Behera}, \bibinfo{person}{Mayank Vatsa}, {and} \bibinfo{person}{Richa
  Singh}.} \bibinfo{year}{2014}\natexlab{}.
\newblock \showarticletitle{On Iris Spoofing Using Print Attack}. In
  \bibinfo{booktitle}{{\em Int. Conf. on Pattern Recognition (ICPR)}}.
  \bibinfo{publisher}{IEEE}, \bibinfo{address}{Stockholm, Sweden},
  \bibinfo{pages}{1681--1686}.
\newblock
\showISSN{1051-4651}
\showDOI{%
\url{https://doi.org/10.1109/ICPR.2014.296}}


\bibitem[\protect\citeauthoryear{He, Li, Liu, Liu, Sun, and He}{He
  et~al\mbox{.}}{2016}]%
        {He_BTAS_2016}
\bibfield{author}{\bibinfo{person}{Lingxiao He}, \bibinfo{person}{Haiqing Li},
  \bibinfo{person}{Fei Liu}, \bibinfo{person}{Nianfeng Liu},
  \bibinfo{person}{Zhenan Sun}, {and} \bibinfo{person}{Zhaofeng He}.}
  \bibinfo{year}{2016}\natexlab{}.
\newblock \showarticletitle{Multi-patch convolution neural network for iris
  liveness detection}. In \bibinfo{booktitle}{{\em 2016 IEEE 8th Int. Conf. on
  Biometrics Theory, Applications and Systems (BTAS)}}.
  \bibinfo{publisher}{IEEE}, \bibinfo{address}{Niagara Falls, NY, USA},
  \bibinfo{pages}{1--7}.
\newblock
\showDOI{%
\url{https://doi.org/10.1109/BTAS.2016.7791186}}


\bibitem[\protect\citeauthoryear{He, An, and Shi}{He et~al\mbox{.}}{2007}]%
        {He_ICB_2007}
\bibfield{author}{\bibinfo{person}{Xiaofu He}, \bibinfo{person}{Shujuan An},
  {and} \bibinfo{person}{Pengfei Shi}.} \bibinfo{year}{2007}\natexlab{}.
\newblock \showarticletitle{Statistical Texture Analysis-Based Approach for
  Fake Iris Detection Using Support Vector Machines}.
\newblock In \bibinfo{booktitle}{{\em Advances in Biometrics: Int. Conference,
  ICB 2007, Seoul, Korea, August 27-29, 2007. Proceedings}},
  \bibfield{editor}{\bibinfo{person}{Seong-Whan Lee} {and}
  \bibinfo{person}{Stan~Z. Li}} (Eds.). \bibinfo{publisher}{Springer Berlin
  Heidelberg}, \bibinfo{address}{Berlin, Heidelberg},
  \bibinfo{pages}{540--546}.
\newblock
\showISBNx{978-3-540-74549-5}
\showDOI{%
\url{https://doi.org/10.1007/978-3-540-74549-5_57}}


\bibitem[\protect\citeauthoryear{He, Lu, and Shi}{He et~al\mbox{.}}{2008}]%
        {He_CCPR_2008}
\bibfield{author}{\bibinfo{person}{Xiaofu He}, \bibinfo{person}{Yue Lu}, {and}
  \bibinfo{person}{Pengfei Shi}.} \bibinfo{year}{2008}\natexlab{}.
\newblock \showarticletitle{A Fake Iris Detection Method Based on FFT and
  Quality Assessment}. In \bibinfo{booktitle}{{\em Chinese Conf. on Pattern
  Recognition}}. \bibinfo{publisher}{IEEE}, \bibinfo{address}{Beijing, China},
  \bibinfo{pages}{1--4}.
\newblock
\showDOI{%
\url{https://doi.org/10.1109/CCPR.2008.68}}


\bibitem[\protect\citeauthoryear{He, Lu, and Shi}{He et~al\mbox{.}}{2009a}]%
        {HeXiaofu_ICB_2009}
\bibfield{author}{\bibinfo{person}{Xiaofu He}, \bibinfo{person}{Yue Lu}, {and}
  \bibinfo{person}{Pengfei Shi}.} \bibinfo{year}{2009}\natexlab{a}.
\newblock \showarticletitle{A New Fake Iris Detection Method}.
\newblock In \bibinfo{booktitle}{{\em {IEEE} Int. Conf. on Biometrics (ICB)}},
  \bibfield{editor}{\bibinfo{person}{Massimo Tistarelli} {and}
  \bibinfo{person}{Mark~S. Nixon}} (Eds.). \bibinfo{publisher}{Springer Berlin
  Heidelberg}, \bibinfo{address}{Berlin, Heidelberg},
  \bibinfo{pages}{1132--1139}.
\newblock
\showISBNx{978-3-642-01793-3}
\showDOI{%
\url{https://doi.org/10.1007/978-3-642-01793-3_114}}


\bibitem[\protect\citeauthoryear{He, Sun, Tan, and Wei}{He
  et~al\mbox{.}}{2009b}]%
        {HeZhaofeng_ICB_2009}
\bibfield{author}{\bibinfo{person}{Zhaofeng He}, \bibinfo{person}{Zhenan Sun},
  \bibinfo{person}{Tieniu Tan}, {and} \bibinfo{person}{Zhuoshi Wei}.}
  \bibinfo{year}{2009}\natexlab{b}.
\newblock \showarticletitle{Efficient Iris Spoof Detection via Boosted Local
  Binary Patterns}.
\newblock In \bibinfo{booktitle}{{\em {IEEE} Int. Conf. on Biometrics (ICB)}},
  \bibfield{editor}{\bibinfo{person}{Massimo Tistarelli} {and}
  \bibinfo{person}{Mark~S. Nixon}} (Eds.). \bibinfo{publisher}{Springer Berlin
  Heidelberg}, \bibinfo{address}{Berlin, Heidelberg},
  \bibinfo{pages}{1080--1090}.
\newblock
\showISBNx{978-3-642-01793-3}
\showDOI{%
\url{https://doi.org/10.1007/978-3-642-01793-3_109}}


\bibitem[\protect\citeauthoryear{Holland and Komogortsev}{Holland and
  Komogortsev}{2013}]%
        {Holland_TIFS_2013}
\bibfield{author}{\bibinfo{person}{Corey~D. Holland} {and}
  \bibinfo{person}{Oleg~V. Komogortsev}.} \bibinfo{year}{2013}\natexlab{}.
\newblock \showarticletitle{Complex Eye Movement Pattern Biometrics: The
  Effects of Environment and Stimulus}.
\newblock \bibinfo{journal}{{\em IEEE Transactions on Information Forensics and
  Security\/}} \bibinfo{volume}{8}, \bibinfo{number}{12} (\bibinfo{date}{Dec}
  \bibinfo{year}{2013}), \bibinfo{pages}{2115--2126}.
\newblock
\showISSN{1556-6013}
\showDOI{%
\url{https://doi.org/10.1109/TIFS.2013.2285884}}


\bibitem[\protect\citeauthoryear{Hsieh, Li, Wang, and Tien}{Hsieh
  et~al\mbox{.}}{2018}]%
        {Hsieh_Sensors_2018}
\bibfield{author}{\bibinfo{person}{Sheng-Hsun Hsieh}, \bibinfo{person}{Yung-Hui
  Li}, \bibinfo{person}{Wei Wang}, {and} \bibinfo{person}{Chung-Hao Tien}.}
  \bibinfo{year}{2018}\natexlab{}.
\newblock \showarticletitle{A Novel Anti-Spoofing Solution for Iris Recognition
  Toward Cosmetic Contact Lens Attack Using Spectral ICA Analysis}.
\newblock \bibinfo{journal}{{\em Sensors\/}} \bibinfo{volume}{18},
  \bibinfo{number}{3} (\bibinfo{year}{2018}), \bibinfo{pages}{1--15}.
\newblock
\showISSN{1424-8220}
\showDOI{%
\url{https://doi.org/10.3390/s18030795}}


\bibitem[\protect\citeauthoryear{Huang, Ti, zhen Hou, Tokuta, and Yang}{Huang
  et~al\mbox{.}}{2013}]%
        {Huang_WACV_2013}
\bibfield{author}{\bibinfo{person}{Xinyu Huang}, \bibinfo{person}{Changpeng
  Ti}, \bibinfo{person}{Qi zhen Hou}, \bibinfo{person}{Alade Tokuta}, {and}
  \bibinfo{person}{Ruigang Yang}.} \bibinfo{year}{2013}\natexlab{}.
\newblock \showarticletitle{An experimental study of pupil constriction for
  liveness detection}. In \bibinfo{booktitle}{{\em {IEEE} Workshop on
  Applications of Computer Vision (WACV)}}. \bibinfo{publisher}{IEEE},
  \bibinfo{address}{Tampa, FL, USA}, \bibinfo{pages}{252--258}.
\newblock
\showISSN{1550-5790}
\showDOI{%
\url{https://doi.org/10.1109/WACV.2013.6475026}}


\bibitem[\protect\citeauthoryear{Hughes and Bowyer}{Hughes and Bowyer}{2013}]%
        {Hughes_HICSS_2013}
\bibfield{author}{\bibinfo{person}{Ken Hughes} {and} \bibinfo{person}{Kevin~W.
  Bowyer}.} \bibinfo{year}{2013}\natexlab{}.
\newblock \showarticletitle{Detection of Contact-Lens-Based Iris Biometric
  Spoofs Using Stereo Imaging}. In \bibinfo{booktitle}{{\em 2013 46th Hawaii
  Int. Conf. on System Sciences}}. \bibinfo{publisher}{IEEE},
  \bibinfo{address}{Wailea, Maui, HI, USA}, \bibinfo{pages}{1763--1772}.
\newblock
\showISSN{1530-1605}
\showDOI{%
\url{https://doi.org/10.1109/HICSS.2013.172}}


\bibitem[\protect\citeauthoryear{{Image Analysis and Biometrics Lab}}{{Image
  Analysis and Biometrics Lab}}{2016}]%
        {IIITD_DBs_URL}
\bibfield{author}{\bibinfo{person}{{Image Analysis and Biometrics Lab}}.}
  \bibinfo{year}{2016}\natexlab{}.
\newblock \bibinfo{title}{{IIITD Contact Lens Iris Database, Iris Combined
  Spoofing Database}}.
\newblock   (\bibinfo{year}{2016}).
\newblock
\showURL{%
Retrieved August 8, 2017 from \url{http://iab-rubric.org/resources.html}}


\bibitem[\protect\citeauthoryear{Jain, Nandakumar, and Nagar}{Jain
  et~al\mbox{.}}{2008}]%
        {Jain_EURASIP_2008}
\bibfield{author}{\bibinfo{person}{Anil~K. Jain}, \bibinfo{person}{Karthik
  Nandakumar}, {and} \bibinfo{person}{Abhishek Nagar}.}
  \bibinfo{year}{2008}\natexlab{}.
\newblock \showarticletitle{Biometric Template Security}.
\newblock \bibinfo{journal}{{\em EURASIP Journal on Advances in Signal
  Processing\/}}  \bibinfo{volume}{2008}, Article \bibinfo{articleno}{113}
  (\bibinfo{date}{Jan.} \bibinfo{year}{2008}), \bibinfo{numpages}{17}~pages.
\newblock
\showISSN{1110-8657}
\showDOI{%
\url{https://doi.org/10.1155/2008/579416}}


\bibitem[\protect\citeauthoryear{Johnson, Tan, and Schuckers}{Johnson
  et~al\mbox{.}}{2010}]%
        {Johnson_IFS_2010}
\bibfield{author}{\bibinfo{person}{Peter Johnson}, \bibinfo{person}{Bin Tan},
  {and} \bibinfo{person}{Stephanie Schuckers}.}
  \bibinfo{year}{2010}\natexlab{}.
\newblock \showarticletitle{Multimodal fusion vulnerability to non-zero effort
  (spoof) imposters}. In \bibinfo{booktitle}{{\em {IEEE} Int. Workshop on
  Information Forensics and Security}}. \bibinfo{publisher}{IEEE},
  \bibinfo{address}{Seattle, WA, USA}, \bibinfo{pages}{1--5}.
\newblock
\showISSN{2157-4766}
\showDOI{%
\url{https://doi.org/10.1109/WIFS.2010.5711469}}


\bibitem[\protect\citeauthoryear{Kanematsu, Takano, and Nakamura}{Kanematsu
  et~al\mbox{.}}{2007}]%
        {Kanematsu_SICE_2007}
\bibfield{author}{\bibinfo{person}{Masashi Kanematsu},
  \bibinfo{person}{Hironobu Takano}, {and} \bibinfo{person}{Kiyomi Nakamura}.}
  \bibinfo{year}{2007}\natexlab{}.
\newblock \showarticletitle{Highly reliable liveness detection method for iris
  recognition}. In \bibinfo{booktitle}{{\em SICE Annual Conference}}.
  \bibinfo{publisher}{IEEE}, \bibinfo{address}{Takamatsu, Japan},
  \bibinfo{pages}{361--364}.
\newblock
\showDOI{%
\url{https://doi.org/10.1109/SICE.2007.4421008}}


\bibitem[\protect\citeauthoryear{Kannala and Rahtu}{Kannala and Rahtu}{2012}]%
        {Kannala_ICPR_2012}
\bibfield{author}{\bibinfo{person}{Juho Kannala} {and} \bibinfo{person}{Esa
  Rahtu}.} \bibinfo{year}{2012}\natexlab{}.
\newblock \showarticletitle{BSIF: Binarized statistical image features}. In
  \bibinfo{booktitle}{{\em Proceedings of the 21st Int. Conf. on Pattern
  Recognition (ICPR2012)}}. \bibinfo{publisher}{IEEE},
  \bibinfo{address}{Tsukuba, Japan}, \bibinfo{pages}{1363--1366}.
\newblock
\showISSN{1051-4651}


\bibitem[\protect\citeauthoryear{Karunya and Kumaresan}{Karunya and
  Kumaresan}{2015}]%
        {Karunya_ACCS_2015}
\bibfield{author}{\bibinfo{person}{R. Karunya} {and} \bibinfo{person}{S.
  Kumaresan}.} \bibinfo{year}{2015}\natexlab{}.
\newblock \showarticletitle{A study of liveness detection in fingerprint and
  iris recognition systems using image quality assessment}. In
  \bibinfo{booktitle}{{\em Int. Conf. on Advanced Computing and Communication
  Systems}}. \bibinfo{publisher}{IEEE}, \bibinfo{address}{Coimbatore, India},
  \bibinfo{pages}{1--5}.
\newblock
\showDOI{%
\url{https://doi.org/10.1109/ICACCS.2015.7324134}}


\bibitem[\protect\citeauthoryear{Kohli, Yadav, Vatsa, and Singh}{Kohli
  et~al\mbox{.}}{2013}]%
        {Kohli_ICB_2013}
\bibfield{author}{\bibinfo{person}{Naman Kohli}, \bibinfo{person}{Daksha
  Yadav}, \bibinfo{person}{Mayank Vatsa}, {and} \bibinfo{person}{Richa Singh}.}
  \bibinfo{year}{2013}\natexlab{}.
\newblock \showarticletitle{Revisiting iris recognition with color cosmetic
  contact lenses}. In \bibinfo{booktitle}{{\em {IEEE} Int. Conf. on Biometrics
  (ICB)}}. \bibinfo{publisher}{IEEE}, \bibinfo{address}{Madrid, Spain},
  \bibinfo{pages}{1--7}.
\newblock
\showISSN{2376-4201}
\showDOI{%
\url{https://doi.org/10.1109/ICB.2013.6613021}}


\bibitem[\protect\citeauthoryear{Kohli, Yadav, Vatsa, Singh, and Noore}{Kohli
  et~al\mbox{.}}{2016}]%
        {Kohli_BTAS_2016}
\bibfield{author}{\bibinfo{person}{Naman Kohli}, \bibinfo{person}{Daksha
  Yadav}, \bibinfo{person}{Mayank Vatsa}, \bibinfo{person}{Richa Singh}, {and}
  \bibinfo{person}{Afzel Noore}.} \bibinfo{year}{2016}\natexlab{}.
\newblock \showarticletitle{Detecting medley of iris spoofing attacks using
  DESIST}. In \bibinfo{booktitle}{{\em {IEEE} Int. Conf. on Biometrics: Theory
  Applications and Systems (BTAS)}}. \bibinfo{publisher}{IEEE},
  \bibinfo{address}{Niagara Falls, NY, USA}, \bibinfo{pages}{1--6}.
\newblock
\showDOI{%
\url{https://doi.org/10.1109/BTAS.2016.7791168}}


\bibitem[\protect\citeauthoryear{Kokkinos and Yuille}{Kokkinos and
  Yuille}{2008}]%
        {Kokkinos_CVPR_2008}
\bibfield{author}{\bibinfo{person}{Iasonas Kokkinos} {and}
  \bibinfo{person}{Alan Yuille}.} \bibinfo{year}{2008}\natexlab{}.
\newblock \showarticletitle{Scale invariance without scale selection}. In
  \bibinfo{booktitle}{{\em {IEEE} Int. Conf. on Computer Vision and Pattern
  Recognition (CVPR)}}. \bibinfo{publisher}{IEEE}, \bibinfo{address}{Anchorage,
  AK, USA}, \bibinfo{pages}{1--8}.
\newblock
\showISSN{1063-6919}
\showDOI{%
\url{https://doi.org/10.1109/CVPR.2008.4587798}}


\bibitem[\protect\citeauthoryear{Komogortsev}{Komogortsev}{2014a}]%
        {ETPAD_v1_URL}
\bibfield{author}{\bibinfo{person}{Oleg Komogortsev}.}
  \bibinfo{year}{2014}\natexlab{a}.
\newblock \bibinfo{title}{{Eye Tracker Print-Attack Database (ETPAD) v1}}.
\newblock   (\bibinfo{year}{2014}).
\newblock
\showURL{%
Retrieved August 8, 2017 from \url{http://cs.txstate.edu/~ok11/etpad_v1.html}}


\bibitem[\protect\citeauthoryear{Komogortsev}{Komogortsev}{2014b}]%
        {ETPAD_v2_URL}
\bibfield{author}{\bibinfo{person}{Oleg Komogortsev}.}
  \bibinfo{year}{2014}\natexlab{b}.
\newblock \bibinfo{title}{{Eye Tracker Print-Attack Database (ETPAD) v2}}.
\newblock   (\bibinfo{year}{2014}).
\newblock
\showURL{%
Retrieved August 8, 2017 from \url{http://cs.txstate.edu/~ok11/etpad_v2.html}}


\bibitem[\protect\citeauthoryear{Komogortsev and Karpov}{Komogortsev and
  Karpov}{2013}]%
        {Komogortsev_ICB_2013}
\bibfield{author}{\bibinfo{person}{Oleg~V. Komogortsev} {and}
  \bibinfo{person}{Alex Karpov}.} \bibinfo{year}{2013}\natexlab{}.
\newblock \showarticletitle{Liveness detection via oculomotor plant
  characteristics: Attack of mechanical replicas}. In \bibinfo{booktitle}{{\em
  {IEEE} Int. Conf. on Biometrics (ICB)}}. \bibinfo{publisher}{IEEE},
  \bibinfo{address}{Madrid, Spain}, \bibinfo{pages}{1--8}.
\newblock
\showISSN{2376-4201}
\showDOI{%
\url{https://doi.org/10.1109/ICB.2013.6612984}}


\bibitem[\protect\citeauthoryear{Komogortsev, Karpov, and Holland}{Komogortsev
  et~al\mbox{.}}{2015}]%
        {Komogortsev_TIFS_2015}
\bibfield{author}{\bibinfo{person}{Oleg~V. Komogortsev},
  \bibinfo{person}{Alexey Karpov}, {and} \bibinfo{person}{Corey~D. Holland}.}
  \bibinfo{year}{2015}\natexlab{}.
\newblock \showarticletitle{Attack of Mechanical Replicas: Liveness Detection
  With Eye Movements}.
\newblock \bibinfo{journal}{{\em {IEEE} Trans. Inf. Forens. Security\/}}
  \bibinfo{volume}{10}, \bibinfo{number}{4} (\bibinfo{date}{April}
  \bibinfo{year}{2015}), \bibinfo{pages}{716--725}.
\newblock
\showISSN{1556-6013}
\showDOI{%
\url{https://doi.org/10.1109/TIFS.2015.2405345}}


\bibitem[\protect\citeauthoryear{Komulainen, Hadid, and
  Pietik\"{a}inen}{Komulainen et~al\mbox{.}}{2014}]%
        {Komulainen_IJCB_2014}
\bibfield{author}{\bibinfo{person}{Jukka Komulainen}, \bibinfo{person}{Abdenour
  Hadid}, {and} \bibinfo{person}{Matti Pietik\"{a}inen}.}
  \bibinfo{year}{2014}\natexlab{}.
\newblock \showarticletitle{Generalized textured contact lens detection by
  extracting BSIF description from Cartesian iris images}. In
  \bibinfo{booktitle}{{\em {IEEE} Int. Joint Conf. on Biometrics (IJCB)}}.
  \bibinfo{publisher}{IEEE}, \bibinfo{address}{Clearwater, FL, USA},
  \bibinfo{pages}{1--7}.
\newblock
\showDOI{%
\url{https://doi.org/10.1109/BTAS.2014.6996237}}


\bibitem[\protect\citeauthoryear{Komulainen, Hadid, and
  Pietik\"{a}inen}{Komulainen et~al\mbox{.}}{2017}]%
        {Komulainen_IETbook_Ch12_2017}
\bibfield{author}{\bibinfo{person}{Jukka Komulainen}, \bibinfo{person}{Abdenour
  Hadid}, {and} \bibinfo{person}{Matti Pietik\"{a}inen}.}
  \bibinfo{year}{2017}\natexlab{}.
\newblock \showarticletitle{Contact lens detection in iris images}.
\newblock In \bibinfo{booktitle}{{\em Iris and Periocular Biometric
  Recognition}}, \bibfield{editor}{\bibinfo{person}{Christian Rathgeb} {and}
  \bibinfo{person}{Christoph Busch}} (Eds.). \bibinfo{publisher}{IET},
  \bibinfo{address}{London, UK}, Chapter~12, \bibinfo{pages}{265--290}.
\newblock
\showISBNx{978-1-78561-168-1}
\showURL{%
\url{https://www.theiet.org/resources/books/security/irisper.cfm}}


\bibitem[\protect\citeauthoryear{Kumar and Puhan}{Kumar and Puhan}{2015}]%
        {Kumar_NCVPRIPG_2015}
\bibfield{author}{\bibinfo{person}{Mohit Kumar} {and}
  \bibinfo{person}{Niladri~Bihari Puhan}.} \bibinfo{year}{2015}\natexlab{}.
\newblock \showarticletitle{Iris liveness detection using texture
  segmentation}. In \bibinfo{booktitle}{{\em National Conf. on Computer Vision,
  Pattern Recognition, Image Processing and Graphics (NCVPRIPG)}}.
  \bibinfo{publisher}{IEEE}, \bibinfo{address}{Patna, India},
  \bibinfo{pages}{1--4}.
\newblock
\showDOI{%
\url{https://doi.org/10.1109/NCVPRIPG.2015.7490042}}


\bibitem[\protect\citeauthoryear{LeCun, Bottou, Bengio, and Haffner}{LeCun
  et~al\mbox{.}}{1998}]%
        {Lecun_ProcIEEE_1998}
\bibfield{author}{\bibinfo{person}{Yann LeCun}, \bibinfo{person}{Leon Bottou},
  \bibinfo{person}{Yoshua Bengio}, {and} \bibinfo{person}{Patrick Haffner}.}
  \bibinfo{year}{1998}\natexlab{}.
\newblock \showarticletitle{Gradient-based learning applied to document
  recognition}.
\newblock \bibinfo{journal}{{\it Proc. {IEEE}}} \bibinfo{volume}{86},
  \bibinfo{number}{11} (\bibinfo{date}{Nov} \bibinfo{year}{1998}),
  \bibinfo{pages}{2278--2324}.
\newblock
\showISSN{0018-9219}
\showDOI{%
\url{https://doi.org/10.1109/5.726791}}


\bibitem[\protect\citeauthoryear{Lee and Park}{Lee and Park}{2010}]%
        {Lee_IMA_2010}
\bibfield{author}{\bibinfo{person}{Eui~Chul Lee} {and}
  \bibinfo{person}{Kang~Ryoung Park}.} \bibinfo{year}{2010}\natexlab{}.
\newblock \showarticletitle{Fake iris detection based on 3D structure of iris
  pattern}.
\newblock \bibinfo{journal}{{\em Int. Journal of Imaging Systems and
  Technology\/}} \bibinfo{volume}{20}, \bibinfo{number}{2}
  (\bibinfo{year}{2010}), \bibinfo{pages}{162--166}.
\newblock
\showISSN{1098-1098}
\showDOI{%
\url{https://doi.org/10.1002/ima.20227}}


\bibitem[\protect\citeauthoryear{Lee, Park, and Kim}{Lee et~al\mbox{.}}{2005}]%
        {Lee_ICB_2005}
\bibfield{author}{\bibinfo{person}{Eui~Chul Lee}, \bibinfo{person}{Kang~Ryoung
  Park}, {and} \bibinfo{person}{Jaihie Kim}.} \bibinfo{year}{2005}\natexlab{}.
\newblock \showarticletitle{Fake Iris Detection by Using Purkinje Image}.
\newblock In \bibinfo{booktitle}{{\em Lecture Notes in Computer Science}},
  \bibfield{editor}{\bibinfo{person}{David Zhang} {and}
  \bibinfo{person}{Anil~K. Jain}} (Eds.). \bibinfo{publisher}{Springer Berlin
  Heidelberg}, \bibinfo{address}{Berlin, Heidelberg},
  \bibinfo{pages}{397--403}.
\newblock
\showISBNx{978-3-540-31621-3}
\showDOI{%
\url{https://doi.org/10.1007/11608288_53}}


\bibitem[\protect\citeauthoryear{Lee, Park, and Kim}{Lee et~al\mbox{.}}{2006}]%
        {Lee_BS_2006}
\bibfield{author}{\bibinfo{person}{Sung~Joo Lee}, \bibinfo{person}{Kang~Ryoung
  Park}, {and} \bibinfo{person}{Jaihie Kim}.} \bibinfo{year}{2006}\natexlab{}.
\newblock \showarticletitle{Robust Fake Iris Detection Based on Variation of
  the Reflectance Ratio Between the Iris and the Sclera}. In
  \bibinfo{booktitle}{{\em Biometrics Symposium: Special Session on Research at
  the Biometric Consortium Conference}}. \bibinfo{publisher}{IEEE},
  \bibinfo{address}{Baltimore, MD, USA}, \bibinfo{pages}{1--6}.
\newblock
\showDOI{%
\url{https://doi.org/10.1109/BCC.2006.4341624}}


\bibitem[\protect\citeauthoryear{Lefohn, Budge, Shirley, Caruso, and
  Reinhard}{Lefohn et~al\mbox{.}}{2003}]%
        {Lefohn_CGA_2003}
\bibfield{author}{\bibinfo{person}{Aaron Lefohn}, \bibinfo{person}{Brian
  Budge}, \bibinfo{person}{Peter Shirley}, \bibinfo{person}{Richard Caruso},
  {and} \bibinfo{person}{Erik Reinhard}.} \bibinfo{year}{2003}\natexlab{}.
\newblock \showarticletitle{An ocularist's approach to human iris synthesis}.
\newblock \bibinfo{journal}{{\em {IEEE} Computer Graphics and Applications\/}}
  \bibinfo{volume}{23}, \bibinfo{number}{6} (\bibinfo{date}{Nov}
  \bibinfo{year}{2003}), \bibinfo{pages}{70--75}.
\newblock
\showISSN{0272-1716}
\showDOI{%
\url{https://doi.org/10.1109/MCG.2003.1242384}}


\bibitem[\protect\citeauthoryear{Lovish, Nigam, Kumar, and Gupta}{Lovish
  et~al\mbox{.}}{2015}]%
        {Lovish_CAIP_2015}
\bibfield{author}{\bibinfo{person}{Lovish}, \bibinfo{person}{Aditya Nigam},
  \bibinfo{person}{Balender Kumar}, {and} \bibinfo{person}{Phalguni Gupta}.}
  \bibinfo{year}{2015}\natexlab{}.
\newblock \showarticletitle{Robust Contact Lens Detection Using Local Phase
  Quantization and Binary Gabor Pattern}.
\newblock In \bibinfo{booktitle}{{\em Int. Conf. on Computer Analysis of Images
  and Patterns (CAIP)}}, \bibfield{editor}{\bibinfo{person}{G.~Azzopardi} {and}
  \bibinfo{person}{N.~Petkov}} (Eds.). \bibinfo{publisher}{Springer},
  \bibinfo{address}{Valletta, Malta}, \bibinfo{pages}{702--714}.
\newblock
\showISBNx{978-3-319-23192-1}


\bibitem[\protect\citeauthoryear{Marsico, Galdi, Nappi, and Riccio}{Marsico
  et~al\mbox{.}}{2014}]%
        {DeMarsico_IVC_2014}
\bibfield{author}{\bibinfo{person}{Maria~De Marsico}, \bibinfo{person}{Chiara
  Galdi}, \bibinfo{person}{Michele Nappi}, {and} \bibinfo{person}{Daniel
  Riccio}.} \bibinfo{year}{2014}\natexlab{}.
\newblock \showarticletitle{FIRME: Face and Iris Recognition for Mobile
  Engagement}.
\newblock \bibinfo{journal}{{\em Image and Vision Computing\/}}
  \bibinfo{volume}{32}, \bibinfo{number}{12} (\bibinfo{year}{2014}),
  \bibinfo{pages}{1161 -- 1172}.
\newblock
\showISSN{0262-8856}
\showDOI{%
\url{https://doi.org/10.1016/j.imavis.2013.12.014}}


\bibitem[\protect\citeauthoryear{Marsico, Nappi, Riccio, and Wechsler}{Marsico
  et~al\mbox{.}}{2015}]%
        {Marsico_PRL_2015}
\bibfield{author}{\bibinfo{person}{Maria~De Marsico}, \bibinfo{person}{Michele
  Nappi}, \bibinfo{person}{Daniel Riccio}, {and} \bibinfo{person}{Harry
  Wechsler}.} \bibinfo{year}{2015}\natexlab{}.
\newblock \showarticletitle{{Mobile Iris Challenge Evaluation (MICHE)-I,
  biometric iris dataset and protocols}}.
\newblock \bibinfo{journal}{{\em Pattern Recognition Letters\/}}
  \bibinfo{volume}{57} (\bibinfo{year}{2015}), \bibinfo{pages}{17 -- 23}.
\newblock
\showISSN{0167-8655}
\showDOI{%
\url{https://doi.org/10.1016/j.patrec.2015.02.009}}
\newblock
\shownote{{Mobile Iris CHallenge Evaluation part I (MICHE I)}.}


\bibitem[\protect\citeauthoryear{Masek and Kovesi}{Masek and Kovesi}{2003}]%
        {MASEK_SOFTWARE_URL}
\bibfield{author}{\bibinfo{person}{Libor Masek} {and} \bibinfo{person}{Peter
  Kovesi}.} \bibinfo{year}{2003}\natexlab{}.
\newblock \bibinfo{title}{{MATLAB Source Code for a Biometric Identification
  System Based on Iris Patterns}}.
\newblock   (\bibinfo{year}{2003}).
\newblock
\showURL{%
Retrieved January 3, 2018 from
  \url{http://www.peterkovesi.com/studentprojects/libor/sourcecode.html}}


\bibitem[\protect\citeauthoryear{Matthew}{Matthew}{2016}]%
        {Matthew_PhD_thesis_2016}
\bibfield{author}{\bibinfo{person}{Peter~William Matthew}.}
  \bibinfo{year}{January 2016}\natexlab{}.
\newblock \bibinfo{title}{Novel approaches to biometric security with an
  emphasis on liveness and coercion detection}.
\newblock   (\bibinfo{year}{January 2016}).
\newblock


\bibitem[\protect\citeauthoryear{Menotti, Chiachia, Pinto, Schwartz, Pedrini,
  Falcao, and Rocha}{Menotti et~al\mbox{.}}{2015}]%
        {Menotti_TIFS_2015}
\bibfield{author}{\bibinfo{person}{David Menotti}, \bibinfo{person}{Giovani
  Chiachia}, \bibinfo{person}{Allan Pinto}, \bibinfo{person}{William~Robson
  Schwartz}, \bibinfo{person}{Helio Pedrini}, \bibinfo{person}{Alexandre~Xavier
  Falcao}, {and} \bibinfo{person}{Anderson Rocha}.}
  \bibinfo{year}{2015}\natexlab{}.
\newblock \showarticletitle{Deep Representations for Iris, Face, and
  Fingerprint Spoofing Detection}.
\newblock \bibinfo{journal}{{\em {IEEE} Trans. Inf. Forens. Security\/}}
  \bibinfo{volume}{10}, \bibinfo{number}{4} (\bibinfo{date}{April}
  \bibinfo{year}{2015}), \bibinfo{pages}{864--879}.
\newblock
\showISSN{1556-6013}
\showDOI{%
\url{https://doi.org/10.1109/TIFS.2015.2398817}}


\bibitem[\protect\citeauthoryear{Morales, Fierrez, Galbally, and
  Gomez-Barrero}{Morales et~al\mbox{.}}{2018}]%
        {Morales_PAD_Handbook_2018}
\bibfield{author}{\bibinfo{person}{Aythami Morales}, \bibinfo{person}{Julian
  Fierrez}, \bibinfo{person}{Javier Galbally}, {and} \bibinfo{person}{Marta
  Gomez-Barrero}.} \bibinfo{year}{2018}\natexlab{}.
\newblock \showarticletitle{{Introduction to Iris Presentation Attack
  Detection}}.
\newblock In \bibinfo{booktitle}{{\em Handbook of Biometric Anti-Spoofing (2nd
  Edition, to appear)}}, \bibfield{editor}{\bibinfo{person}{S\'{e}bastien
  Marcel}, \bibinfo{person}{Mark Nixon}, \bibinfo{person}{Julian Fierrez},
  {and} \bibinfo{person}{Nicholas Evans}} (Eds.). \bibinfo{publisher}{Springer
  Int. Publishing AG}, \bibinfo{pages}{1--15}.
\newblock


\bibitem[\protect\citeauthoryear{{NISLab, NTNU}}{{NISLab, NTNU}}{2016a}]%
        {GUC_VISSIV_DB_URL}
\bibfield{author}{\bibinfo{person}{{NISLab, NTNU}}.}
  \bibinfo{year}{2016}\natexlab{a}.
\newblock \bibinfo{title}{{GUC - VIsible Spectrum Smartphone Iris Video
  (VISSIV) Database}}.
\newblock   (\bibinfo{year}{2016}).
\newblock
\showURL{%
Retrieved August 10, 2017 from
  \url{http://www.nislab.no/biometrics_lab/vissiv_db}}


\bibitem[\protect\citeauthoryear{{NISLab, NTNU}}{{NISLab, NTNU}}{2016b}]%
        {GUC_LF_VIAr_DB_URL}
\bibfield{author}{\bibinfo{person}{{NISLab, NTNU}}.}
  \bibinfo{year}{2016}\natexlab{b}.
\newblock \bibinfo{title}{{GUC Light Field Visible Spectrum Iris Artefact
  Database (GUC-LF-VIAr-DB)}}.
\newblock   (\bibinfo{year}{2016}).
\newblock
\showURL{%
Retrieved August 8, 2017 from
  \url{http://www.nislab.no/biometrics_lab/guc_lf_viar_db}}


\bibitem[\protect\citeauthoryear{{NISLab, NTNU}}{{NISLab, NTNU}}{2016c}]%
        {GUC_VISIA_DB_URL}
\bibfield{author}{\bibinfo{person}{{NISLab, NTNU}}.}
  \bibinfo{year}{2016}\natexlab{c}.
\newblock \bibinfo{title}{{GUC Visible Spectrum Iris Artefact (VSIA)
  Database}}.
\newblock   (\bibinfo{year}{2016}).
\newblock
\showURL{%
Retrieved August 8, 2017 from
  \url{http://www.nislab.no/biometrics_lab/vsia_db}}


\bibitem[\protect\citeauthoryear{{NISLab, NTNU}}{{NISLab, NTNU}}{2016d}]%
        {GUC_PAVID_DB_URL}
\bibfield{author}{\bibinfo{person}{{NISLab, NTNU}}.}
  \bibinfo{year}{2016}\natexlab{d}.
\newblock \bibinfo{title}{{PAVID - Presentation Attack Video Iris Database}}.
\newblock   (\bibinfo{year}{2016}).
\newblock
\showURL{%
Retrieved August 8, 2017 from
  \url{http://www.nislab.no/biometrics_lab/pavid_db}}


\bibitem[\protect\citeauthoryear{Nixon, Aimale, and Rowe}{Nixon
  et~al\mbox{.}}{2008}]%
        {Nixon_HoB_2008}
\bibfield{author}{\bibinfo{person}{Kristin~Adair Nixon},
  \bibinfo{person}{Valerio Aimale}, {and} \bibinfo{person}{Robert~K. Rowe}.}
  \bibinfo{year}{2008}\natexlab{}.
\newblock \showarticletitle{Spoof Detection Schemes}.
\newblock In \bibinfo{booktitle}{{\em Handbook of Biometrics}},
  \bibfield{editor}{\bibinfo{person}{Anil~K. Jain}, \bibinfo{person}{Patrick
  Flynn}, {and} \bibinfo{person}{Arun~A. Ross}} (Eds.).
  \bibinfo{publisher}{Springer US}, \bibinfo{address}{Boston, MA},
  \bibinfo{pages}{403--423}.
\newblock
\showISBNx{978-0-387-71041-9}
\showDOI{%
\url{https://doi.org/10.1007/978-0-387-71041-9_20}}


\bibitem[\protect\citeauthoryear{Ojala, Pietik\"{a}inen, and Harwood}{Ojala
  et~al\mbox{.}}{1994}]%
        {Ojala_ICPR_1994}
\bibfield{author}{\bibinfo{person}{Timo Ojala}, \bibinfo{person}{Matti
  Pietik\"{a}inen}, {and} \bibinfo{person}{David Harwood}.}
  \bibinfo{year}{1994}\natexlab{}.
\newblock \showarticletitle{Performance evaluation of texture measures with
  classification based on Kullback discrimination of distributions}. In
  \bibinfo{booktitle}{{\em Proceedings of 12th Int. Conf. on Pattern
  Recognition}}, Vol.~\bibinfo{volume}{1}. \bibinfo{publisher}{IEEE},
  \bibinfo{address}{Jerusalem, Israel}, \bibinfo{pages}{582--585 vol.1}.
\newblock
\showDOI{%
\url{https://doi.org/10.1109/ICPR.1994.576366}}


\bibitem[\protect\citeauthoryear{Ortiz-Lopez, Galbally, Fierrez, and
  Ortega-Garcia}{Ortiz-Lopez et~al\mbox{.}}{2011}]%
        {Ortiz-Lopez_ICCST_2011}
\bibfield{author}{\bibinfo{person}{Jaime Ortiz-Lopez}, \bibinfo{person}{Javier
  Galbally}, \bibinfo{person}{Julian Fierrez}, {and} \bibinfo{person}{Javier
  Ortega-Garcia}.} \bibinfo{year}{2011}\natexlab{}.
\newblock \showarticletitle{Predicting iris vulnerability to direct attacks
  based on quality related features}. In \bibinfo{booktitle}{{\em {IEEE} Int.
  Carnahan Conf. on Security Technology (ICCST)}}. \bibinfo{publisher}{IEEE},
  \bibinfo{address}{Barcelona, Spain}, \bibinfo{pages}{1--6}.
\newblock
\showISSN{1071-6572}
\showDOI{%
\url{https://doi.org/10.1109/CCST.2011.6095949}}


\bibitem[\protect\citeauthoryear{Pacut and Czajka}{Pacut and Czajka}{2006}]%
        {Pacut_ICCST_2006}
\bibfield{author}{\bibinfo{person}{Andrzej Pacut} {and} \bibinfo{person}{Adam
  Czajka}.} \bibinfo{year}{2006}\natexlab{}.
\newblock \showarticletitle{Aliveness Detection for Iris Biometrics}. In
  \bibinfo{booktitle}{{\em {IEEE} Int. Carnahan Conf. on Security Technology
  (ICCST)}}. \bibinfo{publisher}{IEEE}, \bibinfo{address}{Lexington, KY, USA},
  \bibinfo{pages}{122--129}.
\newblock
\showISSN{1071-6572}
\showDOI{%
\url{https://doi.org/10.1109/CCST.2006.313440}}


\bibitem[\protect\citeauthoryear{Pala and Bhanu}{Pala and Bhanu}{2017}]%
        {Pala_CVPR_2017}
\bibfield{author}{\bibinfo{person}{Federico Pala} {and} \bibinfo{person}{Bir
  Bhanu}.} \bibinfo{year}{2017}\natexlab{}.
\newblock \showarticletitle{Iris Liveness Detection by Relative Distance
  Comparisons}. In \bibinfo{booktitle}{{\em The IEEE Conf. on Computer Vision
  and Pattern Recognition (CVPR) Workshops}}. \bibinfo{publisher}{IEEE},
  \bibinfo{address}{Honolulu, HI, USA}, \bibinfo{pages}{664--671}.
\newblock


\bibitem[\protect\citeauthoryear{Park and Kang}{Park and Kang}{2005}]%
        {Park_LNCS_2005}
\bibfield{author}{\bibinfo{person}{Jong~Hyun Park} {and}
  \bibinfo{person}{Moon~Gi Kang}.} \bibinfo{year}{2005}\natexlab{}.
\newblock \showarticletitle{Iris Recognition Against Counterfeit Attack Using
  Gradient Based Fusion of Multi-spectral Images}.
\newblock In \bibinfo{booktitle}{{\em Advances in Biometric Person
  Authentication: Int. Wokshop on Biometric Recognition Systems, IWBRS 2005,
  Beijing, China, October 22-23, 2005. Proceedings}},
  \bibfield{editor}{\bibinfo{person}{Stan~Z. Li}, \bibinfo{person}{Zhenan Sun},
  \bibinfo{person}{Tieniu Tan}, \bibinfo{person}{Sharath Pankanti},
  \bibinfo{person}{G{\'e}rard Chollet}, {and} \bibinfo{person}{David Zhang}}
  (Eds.). \bibinfo{publisher}{Springer Berlin Heidelberg},
  \bibinfo{address}{Berlin, Heidelberg}, \bibinfo{pages}{150--156}.
\newblock
\showISBNx{978-3-540-32248-1}
\showDOI{%
\url{https://doi.org/10.1007/11569947_19}}


\bibitem[\protect\citeauthoryear{Park and Kang}{Park and Kang}{2007}]%
        {Park_OptEng_2007}
\bibfield{author}{\bibinfo{person}{Jong~Hyun Park} {and}
  \bibinfo{person}{Moon~Gi Kang}.} \bibinfo{year}{2007}\natexlab{}.
\newblock \showarticletitle{Multispectral iris authentication system against
  counterfeit attack using gradient-based image fusion}.
\newblock \bibinfo{journal}{{\em Optical Engineering\/}} \bibinfo{volume}{46},
  \bibinfo{number}{11} (\bibinfo{year}{2007}),
  \bibinfo{pages}{117003--117003--14}.
\newblock
\showISBNx{0091-3286}
\showDOI{%
\url{https://doi.org/10.1117/1.2802367}}


\bibitem[\protect\citeauthoryear{Park}{Park}{2006}]%
        {Park_AMDO_2006}
\bibfield{author}{\bibinfo{person}{Kang~Ryoung Park}.}
  \bibinfo{year}{2006}\natexlab{}.
\newblock \showarticletitle{Robust Fake Iris Detection}.
\newblock In \bibinfo{booktitle}{{\em Articulated Motion and Deformable
  Objects: 4th Int. Conference, AMDO 2006, Port d'Andratx, Mallorca, Spain,
  July 11-14, 2006. Proceedings}},
  \bibfield{editor}{\bibinfo{person}{Francisco~J. Perales} {and}
  \bibinfo{person}{Robert~B. Fisher}} (Eds.). \bibinfo{publisher}{Springer
  Berlin Heidelberg}, \bibinfo{address}{Berlin, Heidelberg},
  \bibinfo{pages}{10--18}.
\newblock
\showISBNx{978-3-540-36032-2}
\showDOI{%
\url{https://doi.org/10.1007/11789239_2}}


\bibitem[\protect\citeauthoryear{Patel, Ratha, and Chellappa}{Patel
  et~al\mbox{.}}{2015}]%
        {Patel_SPM_2015}
\bibfield{author}{\bibinfo{person}{Vishal~M. Patel}, \bibinfo{person}{Nalini~K.
  Ratha}, {and} \bibinfo{person}{Rama Chellappa}.}
  \bibinfo{year}{2015}\natexlab{}.
\newblock \showarticletitle{Cancelable Biometrics: A review}.
\newblock \bibinfo{journal}{{\em {IEEE} Signal Processing Magazine\/}}
  \bibinfo{volume}{32}, \bibinfo{number}{5} (\bibinfo{date}{Sept}
  \bibinfo{year}{2015}), \bibinfo{pages}{54--65}.
\newblock
\showISSN{1053-5888}
\showDOI{%
\url{https://doi.org/10.1109/MSP.2015.2434151}}


\bibitem[\protect\citeauthoryear{Pinto, Pedrini, Krumdick, Becker, Czajka,
  Bowyer, and Rocha}{Pinto et~al\mbox{.}}{2018}]%
        {Pinto_DLB_2018}
\bibfield{author}{\bibinfo{person}{Allan Pinto}, \bibinfo{person}{Helio
  Pedrini}, \bibinfo{person}{Michael Krumdick}, \bibinfo{person}{Benedict
  Becker}, \bibinfo{person}{Adam Czajka}, \bibinfo{person}{Kevin~W. Bowyer},
  {and} \bibinfo{person}{Anderson Rocha}.} \bibinfo{year}{2018}\natexlab{}.
\newblock \showarticletitle{Counteracting Presentation Attacks in Face,
  Fingerprint, and Iris Recognition}.
\newblock In \bibinfo{booktitle}{{\em Deep Learning in Biometrics}},
  \bibfield{editor}{\bibinfo{person}{Angshul~Majumdar Mayank~Vatsa,
  Richa~Singh}} (Ed.). \bibinfo{publisher}{CRC Press}, \bibinfo{address}{Boca
  Raton, London, New York}, \bibinfo{pages}{245--293}.
\newblock
\showISBNx{9781138578234}
\showURL{%
\url{https://www.crcpress.com/Deep-Learning-in-Biometrics/Vatsa-Singh-Majumdar/p/book/9781138578234}}


\bibitem[\protect\citeauthoryear{Puhan, Natarajan, and Hegde}{Puhan
  et~al\mbox{.}}{2011}]%
        {Puhan_ISCE_2011}
\bibfield{author}{\bibinfo{person}{Niladri~Bihari Puhan},
  \bibinfo{person}{Sudha Natarajan}, {and} \bibinfo{person}{A.~Suhas Hegde}.}
  \bibinfo{year}{2011}\natexlab{}.
\newblock \showarticletitle{A new iris liveness detection method against
  contact lens spoofing}. In \bibinfo{booktitle}{{\em {IEEE} Int. Symp. on
  Consumer Electronics (ISCE)}}. \bibinfo{publisher}{IEEE},
  \bibinfo{address}{Singapore}, \bibinfo{pages}{71--74}.
\newblock
\showISSN{0747-668X}
\showDOI{%
\url{https://doi.org/10.1109/ISCE.2011.5973786}}


\bibitem[\protect\citeauthoryear{Quinn, Grother, and Matey}{Quinn
  et~al\mbox{.}}{2018}]%
        {IREX_IX}
\bibfield{author}{\bibinfo{person}{George~W. Quinn}, \bibinfo{person}{Patrick
  Grother}, {and} \bibinfo{person}{James Matey}.}
  \bibinfo{year}{2018}\natexlab{}.
\newblock \bibinfo{booktitle}{{\em {IREX IX Part One: Performance of Iris
  Recognition Algorithms}}}.
\newblock \bibinfo{type}{{T}echnical {R}eport}. \bibinfo{institution}{NIST}.
\newblock
\showURL{%
\url{https://doi.org/10.6028/NIST.IR.8207}}
\newblock
\shownote{{Interagency Report 7629}.}


\bibitem[\protect\citeauthoryear{Raghavendra and Busch}{Raghavendra and
  Busch}{2014a}]%
        {Raghavendra_EUSIPCO_2014}
\bibfield{author}{\bibinfo{person}{Ramachandra Raghavendra} {and}
  \bibinfo{person}{Christoph Busch}.} \bibinfo{year}{2014}\natexlab{a}.
\newblock \showarticletitle{Presentation attack detection algorithm for face
  and iris biometrics}. In \bibinfo{booktitle}{{\em European Signal Processing
  Conf. (EUSIPCO)}}. \bibinfo{publisher}{IEEE}, \bibinfo{address}{Lisbon,
  Portugal}, \bibinfo{pages}{1387--1391}.
\newblock
\showISSN{2219-5491}


\bibitem[\protect\citeauthoryear{Raghavendra and Busch}{Raghavendra and
  Busch}{2014b}]%
        {Raghavendra_IJCB_2014}
\bibfield{author}{\bibinfo{person}{Ramachandra Raghavendra} {and}
  \bibinfo{person}{Christoph Busch}.} \bibinfo{year}{2014}\natexlab{b}.
\newblock \showarticletitle{Presentation attack detection on visible spectrum
  iris recognition by exploring inherent characteristics of Light Field
  Camera}. In \bibinfo{booktitle}{{\em {IEEE} Int. Joint Conf. on Biometrics
  (IJCB)}}. \bibinfo{publisher}{IEEE}, \bibinfo{address}{Clearwater, FL, USA},
  \bibinfo{pages}{1--8}.
\newblock
\showDOI{%
\url{https://doi.org/10.1109/BTAS.2014.6996226}}


\bibitem[\protect\citeauthoryear{Raghavendra and Busch}{Raghavendra and
  Busch}{2015}]%
        {Raghavendra_TIFS_2015}
\bibfield{author}{\bibinfo{person}{Ramachandra Raghavendra} {and}
  \bibinfo{person}{Christoph Busch}.} \bibinfo{year}{2015}\natexlab{}.
\newblock \showarticletitle{Robust Scheme for Iris Presentation Attack
  Detection Using Multiscale Binarized Statistical Image Features}.
\newblock \bibinfo{journal}{{\em {IEEE} Trans. Inf. Forens. Security\/}}
  \bibinfo{volume}{10}, \bibinfo{number}{4} (\bibinfo{date}{April}
  \bibinfo{year}{2015}), \bibinfo{pages}{703--715}.
\newblock
\showISSN{1556-6013}
\showDOI{%
\url{https://doi.org/10.1109/TIFS.2015.2400393}}


\bibitem[\protect\citeauthoryear{Raghavendra, Raja, and Busch}{Raghavendra
  et~al\mbox{.}}{2014}]%
        {Raghavendra_ESI_2014}
\bibfield{author}{\bibinfo{person}{Ramachandra Raghavendra},
  \bibinfo{person}{Kiran~B. Raja}, {and} \bibinfo{person}{Christoph Busch}.}
  \bibinfo{year}{2014}\natexlab{}.
\newblock \showarticletitle{Ensemble of Statistically Independent Filters for
  Robust Contact Lens Detection in Iris Images}. In \bibinfo{booktitle}{{\em
  Proceedings of the 2014 Indian Conf. on Computer Vision Graphics and Image
  Processing}} {\em (\bibinfo{series}{ICVGIP '14})}. \bibinfo{publisher}{ACM},
  \bibinfo{address}{New York, NY, USA}, Article \bibinfo{articleno}{24},
  \bibinfo{numpages}{7}~pages.
\newblock
\showISBNx{978-1-4503-3061-9}
\showDOI{%
\url{https://doi.org/10.1145/2683483.2683507}}


\bibitem[\protect\citeauthoryear{Raghavendra, Raja, and Busch}{Raghavendra
  et~al\mbox{.}}{2017}]%
        {Raghavendra_WACV_2017}
\bibfield{author}{\bibinfo{person}{Ramachandra Raghavendra},
  \bibinfo{person}{Kiran~B. Raja}, {and} \bibinfo{person}{Christoph Busch}.}
  \bibinfo{year}{2017}\natexlab{}.
\newblock \showarticletitle{ContlensNet: Robust Iris Contact Lens Detection
  Using Deep Convolutional Neural Networks}. In \bibinfo{booktitle}{{\em {IEEE}
  Winter Conf. on Applications of Computer Vision (WACV)}}.
  \bibinfo{publisher}{IEEE}, \bibinfo{address}{Santa Rosa, CA, USA},
  \bibinfo{pages}{1160--1167}.
\newblock
\showDOI{%
\url{https://doi.org/10.1109/WACV.2017.134}}


\bibitem[\protect\citeauthoryear{Raja, Raghavendra, Braun, and Busch}{Raja
  et~al\mbox{.}}{2016b}]%
        {Raja_BIOSIG_2016}
\bibfield{author}{\bibinfo{person}{Kiran~B. Raja}, \bibinfo{person}{R.
  Raghavendra}, \bibinfo{person}{Jean-Noel Braun}, {and}
  \bibinfo{person}{Christoph Busch}.} \bibinfo{year}{2016}\natexlab{b}.
\newblock \showarticletitle{Presentation attack detection using a generalizable
  statistical approach for periocular and iris systems}. In
  \bibinfo{booktitle}{{\em Int. Conf. of the Biometrics Special Interest Group
  (BIOSIG)}}, \bibfield{editor}{\bibinfo{person}{Arslan Br\"{o}mme},
  \bibinfo{person}{Christoph Busch}, \bibinfo{person}{Christian Rathgeb}, {and}
  \bibinfo{person}{Andreas Uhl}} (Eds.). \bibinfo{publisher}{Gesellschaft
  f\"{u}r Informatik e.V.}, \bibinfo{address}{Bonn}, \bibinfo{pages}{111--122}.
\newblock


\bibitem[\protect\citeauthoryear{Raja, Raghavendra, and Busch}{Raja
  et~al\mbox{.}}{2015a}]%
        {Raja_BTAS_2015}
\bibfield{author}{\bibinfo{person}{Kiran~B. Raja}, \bibinfo{person}{Ramachandra
  Raghavendra}, {and} \bibinfo{person}{Christoph Busch}.}
  \bibinfo{year}{2015}\natexlab{a}.
\newblock \showarticletitle{Presentation attack detection using Laplacian
  decomposed frequency response for visible spectrum and Near-Infra-Red iris
  systems}. In \bibinfo{booktitle}{{\em {IEEE} Int. Conf. on Biometrics: Theory
  Applications and Systems (BTAS)}}. \bibinfo{publisher}{IEEE},
  \bibinfo{address}{Arlington, VA, USA}, \bibinfo{pages}{1--8}.
\newblock
\showDOI{%
\url{https://doi.org/10.1109/BTAS.2015.7358790}}


\bibitem[\protect\citeauthoryear{Raja, Raghavendra, and Busch}{Raja
  et~al\mbox{.}}{2015b}]%
        {Raja_TIFS_2015}
\bibfield{author}{\bibinfo{person}{Kiran~B. Raja}, \bibinfo{person}{Ramachandra
  Raghavendra}, {and} \bibinfo{person}{Christoph Busch}.}
  \bibinfo{year}{2015}\natexlab{b}.
\newblock \showarticletitle{Video Presentation Attack Detection in Visible
  Spectrum Iris Recognition Using Magnified Phase Information}.
\newblock \bibinfo{journal}{{\em {IEEE} Trans. Inf. Forens. Security\/}}
  \bibinfo{volume}{10}, \bibinfo{number}{10} (\bibinfo{date}{October}
  \bibinfo{year}{2015}), \bibinfo{pages}{2048--2056}.
\newblock
\showISSN{1556-6013}
\showDOI{%
\url{https://doi.org/10.1109/TIFS.2015.2440188}}


\bibitem[\protect\citeauthoryear{Raja, Raghavendra, and Busch}{Raja
  et~al\mbox{.}}{2016a}]%
        {Raja_SIN_2016}
\bibfield{author}{\bibinfo{person}{Kiran~B. Raja}, \bibinfo{person}{Ramachandra
  Raghavendra}, {and} \bibinfo{person}{Christoph Busch}.}
  \bibinfo{year}{2016}\natexlab{a}.
\newblock \showarticletitle{Color Adaptive Quantized Patterns for Presentation
  Attack Detection in Ocular Biometric Systems}. In \bibinfo{booktitle}{{\em
  Int. Conf. on Security of Information and Networks}} {\em
  (\bibinfo{series}{SIN'16})}. \bibinfo{publisher}{ACM}, \bibinfo{address}{New
  York, NY, USA}, \bibinfo{pages}{9--15}.
\newblock
\showISBNx{978-1-4503-4764-8}
\showDOI{%
\url{https://doi.org/10.1145/2947626.2951959}}


\bibitem[\protect\citeauthoryear{Rathgeb and Busch}{Rathgeb and Busch}{2017}]%
        {Rathgeb_IJCB_2017}
\bibfield{author}{\bibinfo{person}{Christian Rathgeb} {and}
  \bibinfo{person}{Christoph Busch}.} \bibinfo{year}{2017}\natexlab{}.
\newblock \showarticletitle{On the feasibility of creating morphed iris-codes}.
  In \bibinfo{booktitle}{{\em {IEEE} Int. Joint Conf. on Biometrics (IJCB)}}.
  \bibinfo{publisher}{IEEE}, \bibinfo{address}{Denver, CO, USA},
  \bibinfo{pages}{152--157}.
\newblock
\showDOI{%
\url{https://doi.org/10.1109/BTAS.2017.8272693}}


\bibitem[\protect\citeauthoryear{Rathgeb and Uhl}{Rathgeb and Uhl}{2010}]%
        {Rathgeb_ICPR_2010}
\bibfield{author}{\bibinfo{person}{Christian Rathgeb} {and}
  \bibinfo{person}{Andreas Uhl}.} \bibinfo{year}{2010}\natexlab{}.
\newblock \showarticletitle{Attacking Iris Recognition: An Efficient
  Hill-Climbing Technique}. In \bibinfo{booktitle}{{\em Int. Conf. on Pattern
  Recognition (ICPR)}}. \bibinfo{publisher}{IEEE}, \bibinfo{address}{Istanbul,
  Turkey}, \bibinfo{pages}{1217--1220}.
\newblock
\showISSN{1051-4651}
\showDOI{%
\url{https://doi.org/10.1109/ICPR.2010.303}}


\bibitem[\protect\citeauthoryear{Rathgeb and Uhl}{Rathgeb and Uhl}{2011}]%
        {Rathgeb_EURASIP_2011}
\bibfield{author}{\bibinfo{person}{Christian Rathgeb} {and}
  \bibinfo{person}{Andreas Uhl}.} \bibinfo{year}{2011}\natexlab{}.
\newblock \showarticletitle{A survey on biometric cryptosystems and cancelable
  biometrics}.
\newblock \bibinfo{journal}{{\em EURASIP Journal on Information Security\/}}
  \bibinfo{volume}{2011}, \bibinfo{number}{1} (\bibinfo{date}{23 Sep}
  \bibinfo{year}{2011}), \bibinfo{pages}{3}.
\newblock
\showISSN{1687-417X}
\showDOI{%
\url{https://doi.org/10.1186/1687-417X-2011-3}}


\bibitem[\protect\citeauthoryear{Rigas and Komogortsev}{Rigas and
  Komogortsev}{2014}]%
        {Rigas_IJCB_2014}
\bibfield{author}{\bibinfo{person}{Ioannis Rigas} {and}
  \bibinfo{person}{Oleg~V. Komogortsev}.} \bibinfo{year}{2014}\natexlab{}.
\newblock \showarticletitle{Gaze estimation as a framework for iris liveness
  detection}. In \bibinfo{booktitle}{{\em {IEEE} Int. Joint Conf. on Biometrics
  (IJCB)}}. \bibinfo{publisher}{IEEE}, \bibinfo{address}{Clearwater, FL, USA},
  \bibinfo{pages}{1--8}.
\newblock
\showDOI{%
\url{https://doi.org/10.1109/BTAS.2014.6996282}}


\bibitem[\protect\citeauthoryear{Rigas and Komogortsev}{Rigas and
  Komogortsev}{2015}]%
        {Rigas_PRL_2015}
\bibfield{author}{\bibinfo{person}{Ioannis Rigas} {and}
  \bibinfo{person}{Oleg~V. Komogortsev}.} \bibinfo{year}{2015}\natexlab{}.
\newblock \showarticletitle{Eye movement-driven defense against iris
  print-attacks}.
\newblock \bibinfo{journal}{{\em Pattern Recognition Letters\/}}
  \bibinfo{volume}{68, Part 2} (\bibinfo{year}{2015}), \bibinfo{pages}{316 --
  326}.
\newblock
\showISSN{0167-8655}
\showDOI{%
\url{https://doi.org/10.1016/j.patrec.2015.06.011}}
\newblock
\shownote{Special Issue on ``Soft Biometrics''.}


\bibitem[\protect\citeauthoryear{Ruiz-Albacete, Tome-Gonzalez,
  Alonso-Fernandez, Galbally, Fierrez, and Ortega-Garcia}{Ruiz-Albacete
  et~al\mbox{.}}{2008}]%
        {Ruiz-Albacete_LNCS_2008}
\bibfield{author}{\bibinfo{person}{Virginia Ruiz-Albacete},
  \bibinfo{person}{Pedro Tome-Gonzalez}, \bibinfo{person}{Fernando
  Alonso-Fernandez}, \bibinfo{person}{Javier Galbally}, \bibinfo{person}{Julian
  Fierrez}, {and} \bibinfo{person}{Javier Ortega-Garcia}.}
  \bibinfo{year}{2008}\natexlab{}.
\newblock \showarticletitle{Direct Attacks Using Fake Images in Iris
  Verification}.
\newblock In \bibinfo{booktitle}{{\em Lecture Notes in Computer Science}}.
  \bibinfo{series}{Lecture Notes in Computer Science},
  Vol.~\bibinfo{volume}{5372}. \bibinfo{publisher}{Springer},
  \bibinfo{address}{Roskilde, Denmark}, \bibinfo{pages}{181--190}.
\newblock


\bibitem[\protect\citeauthoryear{Sansola}{Sansola}{2015}]%
        {Sansola_MastersThesis_2015}
\bibfield{author}{\bibinfo{person}{Alora Sansola}.}
  \bibinfo{year}{2015}\natexlab{}.
\newblock {\em \bibinfo{title}{Postmortem iris recognition and its application
  in human identification}}.
\newblock \bibinfo{thesistype}{Master's\ thesis}. \bibinfo{school}{Boston
  University}, \bibinfo{address}{Boston, MA, USA}.
\newblock


\bibitem[\protect\citeauthoryear{Sauerwein, Saul, Steadman, and
  Boehnen}{Sauerwein et~al\mbox{.}}{2017}]%
        {Sauerwein_JFO_2017}
\bibfield{author}{\bibinfo{person}{Kelly Sauerwein},
  \bibinfo{person}{Tiffany~B. Saul}, \bibinfo{person}{Dawnie~Wolfe Steadman},
  {and} \bibinfo{person}{Chris~B. Boehnen}.} \bibinfo{year}{2017}\natexlab{}.
\newblock \showarticletitle{The Effect of Decomposition on the Efficacy of
  Biometrics for Positive Identification}.
\newblock \bibinfo{journal}{{\em Journal of Forensic Sciences\/}}
  \bibinfo{volume}{62}, \bibinfo{number}{6} (\bibinfo{year}{2017}),
  \bibinfo{pages}{1599--1602}.
\newblock
\showISSN{1556-4029}
\showDOI{%
\url{https://doi.org/10.1111/1556-4029.13484}}


\bibitem[\protect\citeauthoryear{Sequeira, Chen, Ferryman, Wild,
  Alonso-Fernandez, Bigun, Raja, Raghavendra, Busch, de~Freitas~Pereira,
  Marcel, Behera, Gour, and Kanhangad}{Sequeira et~al\mbox{.}}{2017}]%
        {Sequeira_IJCB_2017}
\bibfield{author}{\bibinfo{person}{Ana~F. Sequeira}, \bibinfo{person}{Lulu
  Chen}, \bibinfo{person}{James Ferryman}, \bibinfo{person}{Peter Wild},
  \bibinfo{person}{Fernando Alonso-Fernandez}, \bibinfo{person}{Josef Bigun},
  \bibinfo{person}{Kiran~B. Raja}, \bibinfo{person}{Ramachandra Raghavendra},
  \bibinfo{person}{Christoph Busch}, \bibinfo{person}{Tiago de
  Freitas~Pereira}, \bibinfo{person}{Sebastien Marcel},
  \bibinfo{person}{Sushree~Sangeeta Behera}, \bibinfo{person}{Mahesh Gour},
  {and} \bibinfo{person}{Vivek Kanhangad}.} \bibinfo{year}{2017}\natexlab{}.
\newblock \showarticletitle{Cross-eyed 2017: Cross-spectral iris/periocular
  recognition competition}. In \bibinfo{booktitle}{{\em {IEEE} Int. Joint Conf.
  on Biometrics (IJCB)}}. \bibinfo{publisher}{IEEE}, \bibinfo{address}{Denver,
  CO, USA}, \bibinfo{pages}{725--732}.
\newblock
\showDOI{%
\url{https://doi.org/10.1109/BTAS.2017.8272762}}


\bibitem[\protect\citeauthoryear{Sequeira, Monteiro, Rebelo, and
  Oliveira}{Sequeira et~al\mbox{.}}{2014a}]%
        {Sequeira_VISAPPb_2014}
\bibfield{author}{\bibinfo{person}{Ana~F. Sequeira},
  \bibinfo{person}{Jo{\~{a}}o~C. Monteiro}, \bibinfo{person}{Ana Rebelo}, {and}
  \bibinfo{person}{H\'{e}lder~P. Oliveira}.} \bibinfo{year}{2014}\natexlab{a}.
\newblock \showarticletitle{{MobBIO: A multimodal database captured with a
  portable handheld device}}. In \bibinfo{booktitle}{{\em {IEEE} Int. Conf. on
  Computer Vision Theory and Applications (VISAPP)}}, Vol.~\bibinfo{volume}{3}.
  \bibinfo{publisher}{IEEE}, \bibinfo{address}{Lisbon, Portugal},
  \bibinfo{pages}{133--139}.
\newblock


\bibitem[\protect\citeauthoryear{Sequeira, Murari, and Cardoso}{Sequeira
  et~al\mbox{.}}{2014b}]%
        {Sequeira_VISAPPa_2014}
\bibfield{author}{\bibinfo{person}{Ana~F. Sequeira}, \bibinfo{person}{Juliano
  Murari}, {and} \bibinfo{person}{Jaime~S. Cardoso}.}
  \bibinfo{year}{2014}\natexlab{b}.
\newblock \showarticletitle{Iris liveness detection methods in mobile
  applications}. In \bibinfo{booktitle}{{\em {IEEE} Int. Conf. on Computer
  Vision Theory and Applications (VISAPP)}}, Vol.~\bibinfo{volume}{3}.
  \bibinfo{publisher}{IEEE}, \bibinfo{address}{Lisbon, Portugal},
  \bibinfo{pages}{22--33}.
\newblock


\bibitem[\protect\citeauthoryear{Sequeira, Murari, and Cardoso}{Sequeira
  et~al\mbox{.}}{2014c}]%
        {Sequeira_IJCNN_2014}
\bibfield{author}{\bibinfo{person}{Ana~F. Sequeira}, \bibinfo{person}{Juliano
  Murari}, {and} \bibinfo{person}{Jaime~S. Cardoso}.}
  \bibinfo{year}{2014}\natexlab{c}.
\newblock \showarticletitle{Iris liveness detection methods in the mobile
  biometrics scenario}. In \bibinfo{booktitle}{{\em {IEEE} Int. Joint Conf. on
  Neural Networks (IJCNN)}}. \bibinfo{publisher}{IEEE},
  \bibinfo{address}{Beijing, China}, \bibinfo{pages}{3002--3008}.
\newblock
\showISSN{2161-4393}
\showDOI{%
\url{https://doi.org/10.1109/IJCNN.2014.6889816}}


\bibitem[\protect\citeauthoryear{Sequeira, Oliveira, Monteiro, Monteiro, and
  Cardoso}{Sequeira et~al\mbox{.}}{2014d}]%
        {Sequeira_IJCB_2014}
\bibfield{author}{\bibinfo{person}{Ana~F. Sequeira},
  \bibinfo{person}{H\'{e}lder~P. Oliveira}, \bibinfo{person}{Jo{\~{a}}o~C.
  Monteiro}, \bibinfo{person}{Jo{\~{a}}o~P. Monteiro}, {and}
  \bibinfo{person}{Jaime~S. Cardoso}.} \bibinfo{year}{2014}\natexlab{d}.
\newblock \showarticletitle{{MobILive} 2014 - Mobile Iris Liveness Detection
  Competition}. In \bibinfo{booktitle}{{\em {IEEE} Int. Joint Conf. on
  Biometrics (IJCB)}}. \bibinfo{publisher}{IEEE}, \bibinfo{address}{Clearwater,
  FL, USA}, \bibinfo{pages}{1--6}.
\newblock
\showDOI{%
\url{https://doi.org/10.1109/BTAS.2014.6996290}}


\bibitem[\protect\citeauthoryear{Sequeira, Thavalengal, Ferryman, Corcoran, and
  Cardoso}{Sequeira et~al\mbox{.}}{2016}]%
        {Sequeira_TSP_2016}
\bibfield{author}{\bibinfo{person}{Ana~F. Sequeira}, \bibinfo{person}{Shejin
  Thavalengal}, \bibinfo{person}{James Ferryman}, \bibinfo{person}{Peter
  Corcoran}, {and} \bibinfo{person}{Jaime~S. Cardoso}.}
  \bibinfo{year}{2016}\natexlab{}.
\newblock \showarticletitle{A realistic evaluation of iris presentation attack
  detection}. In \bibinfo{booktitle}{{\em Int. Conf. on Telecommunications and
  Signal Processing (TSP)}}. \bibinfo{publisher}{IEEE},
  \bibinfo{address}{Vienna, Austria}, \bibinfo{pages}{660--664}.
\newblock
\showDOI{%
\url{https://doi.org/10.1109/TSP.2016.7760965}}


\bibitem[\protect\citeauthoryear{Shah and Ross}{Shah and Ross}{2006}]%
        {Shah_ICIP_2006}
\bibfield{author}{\bibinfo{person}{Samir Shah} {and} \bibinfo{person}{Arun
  Ross}.} \bibinfo{year}{2006}\natexlab{}.
\newblock \showarticletitle{Generating Synthetic Irises by Feature
  Agglomeration}. In \bibinfo{booktitle}{{\em 2006 Int. Conf. on Image
  Processing}}. \bibinfo{publisher}{IEEE}, \bibinfo{address}{Atlanta, GA, USA},
  \bibinfo{pages}{317--320}.
\newblock
\showISSN{1522-4880}
\showDOI{%
\url{https://doi.org/10.1109/ICIP.2006.313157}}


\bibitem[\protect\citeauthoryear{Shaydyuk and Cleland}{Shaydyuk and
  Cleland}{2016}]%
        {Shaydyuk_ICCST_2016}
\bibfield{author}{\bibinfo{person}{Nazariy~K. Shaydyuk} {and}
  \bibinfo{person}{Timothy Cleland}.} \bibinfo{year}{2016}\natexlab{}.
\newblock \showarticletitle{Biometric identification via retina scanning with
  liveness detection using speckle contrast imaging}. In
  \bibinfo{booktitle}{{\em {IEEE} Int. Carnahan Conf. on Security Technology
  (ICCST)}}. \bibinfo{publisher}{IEEE}, \bibinfo{address}{Orlando, FL, USA},
  \bibinfo{pages}{1--5}.
\newblock
\showDOI{%
\url{https://doi.org/10.1109/CCST.2016.7815706}}


\bibitem[\protect\citeauthoryear{Silva, Luz, Baeta, Pedrini, Falcao, and
  Menotti}{Silva et~al\mbox{.}}{2015}]%
        {Silva_SIBGRAPI_2015}
\bibfield{author}{\bibinfo{person}{Pedro Silva}, \bibinfo{person}{Eduardo Luz},
  \bibinfo{person}{Rafael Baeta}, \bibinfo{person}{Helio Pedrini},
  \bibinfo{person}{Alexandre~Xavier Falcao}, {and} \bibinfo{person}{David
  Menotti}.} \bibinfo{year}{2015}\natexlab{}.
\newblock \showarticletitle{An Approach to Iris Contact Lens Detection based on
  Deep Image Representations}. In \bibinfo{booktitle}{{\em Conf. on Graphics,
  Patterns and Images (SIBGRAPI)}}. IEEE, \bibinfo{publisher}{IEEE},
  \bibinfo{address}{Salvador, Brazil}, \bibinfo{pages}{157--164}.
\newblock
\showISSN{1530-1834}
\showDOI{%
\url{https://doi.org/10.1109/SIBGRAPI.2015.16}}


\bibitem[\protect\citeauthoryear{Singh, Mistry, Yadav, and Nigam}{Singh
  et~al\mbox{.}}{2018}]%
        {Singh_ISBA_2018}
\bibfield{author}{\bibinfo{person}{Avantika Singh}, \bibinfo{person}{Vishesh
  Mistry}, \bibinfo{person}{Dhananjay Yadav}, {and} \bibinfo{person}{Aditya
  Nigam}.} \bibinfo{year}{2018}\natexlab{}.
\newblock \showarticletitle{{GHCLNet: A Generalized Hierarchically tuned
  Contact Lens detection Network}}. In \bibinfo{booktitle}{{\em {IEEE} Int.
  Conf. on Identity, Security and Behavior Analysis (ISBA)}}.
  \bibinfo{publisher}{IEEE}, \bibinfo{address}{Singapore},
  \bibinfo{pages}{1--6}.
\newblock


\bibitem[\protect\citeauthoryear{Singh and Singh}{Singh and Singh}{2011}]%
        {Singh_ICT_2011}
\bibfield{author}{\bibinfo{person}{Yogendra~Narain Singh} {and}
  \bibinfo{person}{Sanjay~Kumar Singh}.} \bibinfo{year}{2011}\natexlab{}.
\newblock \showarticletitle{Vitality detection from biometrics:
  State-of-the-art}. In \bibinfo{booktitle}{{\em World Congress on Information
  and Communication Technologies}}. \bibinfo{publisher}{IEEE},
  \bibinfo{address}{Mumbai, India}, \bibinfo{pages}{106--111}.
\newblock
\showDOI{%
\url{https://doi.org/10.1109/WICT.2011.6141226}}


\bibitem[\protect\citeauthoryear{Smith, Yin, Feiner, and Nayar}{Smith
  et~al\mbox{.}}{2013a}]%
        {CAVE_DB_URL}
\bibfield{author}{\bibinfo{person}{Brian~A. Smith}, \bibinfo{person}{Qi Yin},
  \bibinfo{person}{Steven~K. Feiner}, {and} \bibinfo{person}{Shree~K. Nayar}.}
  \bibinfo{year}{2013}\natexlab{a}.
\newblock \bibinfo{title}{Columbia Gaze Data Set}.
\newblock   (\bibinfo{year}{2013}).
\newblock
\showURL{%
Retrieved August 8, 2017 from
  \url{http://www.cs.columbia.edu/CAVE/databases/columbia_gaze}}


\bibitem[\protect\citeauthoryear{Smith, Yin, Feiner, and Nayar}{Smith
  et~al\mbox{.}}{2013b}]%
        {Smith_UIST_2013}
\bibfield{author}{\bibinfo{person}{Brian~A. Smith}, \bibinfo{person}{Qi Yin},
  \bibinfo{person}{Steven~K. Feiner}, {and} \bibinfo{person}{Shree~K. Nayar}.}
  \bibinfo{year}{2013}\natexlab{b}.
\newblock \showarticletitle{Gaze Locking: Passive Eye Contact Detection for
  Human-object Interaction}. In \bibinfo{booktitle}{{\em Annual ACM Symp. on
  User Interface Software and Technology}} {\em (\bibinfo{series}{UIST '13})}.
  \bibinfo{publisher}{ACM}, \bibinfo{address}{New York, NY, USA},
  \bibinfo{pages}{271--280}.
\newblock
\showISBNx{978-1-4503-2268-3}
\showDOI{%
\url{https://doi.org/10.1145/2501988.2501994}}


\bibitem[\protect\citeauthoryear{{SOCIA Lab. - Soft Computing and Image
  Analysis Group}}{{SOCIA Lab. - Soft Computing and Image Analysis
  Group}}{2004}]%
        {UBIRIS_DB_URL}
\bibfield{author}{\bibinfo{person}{{SOCIA Lab. - Soft Computing and Image
  Analysis Group}}.} \bibinfo{year}{2004}\natexlab{}.
\newblock \bibinfo{title}{{Noisy Visible Wavelength Iris Image Databases
  (UBIRIS)}}.
\newblock   (\bibinfo{year}{2004}).
\newblock
\showURL{%
Retrieved January 19, 2018 from \url{http://iris.di.ubi.pt}}


\bibitem[\protect\citeauthoryear{S{\"{o}}llinger, Trung, and
  Uhl}{S{\"{o}}llinger et~al\mbox{.}}{2018}]%
        {Sollinger_IET_2017}
\bibfield{author}{\bibinfo{person}{Dominik S{\"{o}}llinger},
  \bibinfo{person}{Pauline Trung}, {and} \bibinfo{person}{Andreas Uhl}.}
  \bibinfo{year}{2018}\natexlab{}.
\newblock \showarticletitle{Non-reference image quality assessment and natural
  scene statistics to counter biometric sensor spoofing}.
\newblock \bibinfo{journal}{{\em IET Biometrics\/}} (\bibinfo{date}{January}
  \bibinfo{year}{2018}), \bibinfo{pages}{1--11}.
\newblock
\showISSN{2047-4938}
\showDOI{%
\url{https://doi.org/10.1049/iet-bmt.2017.0146}}


\bibitem[\protect\citeauthoryear{Sun and Tan}{Sun and Tan}{2014}]%
        {Sun_HoBAS_2014}
\bibfield{author}{\bibinfo{person}{Zhenan Sun} {and} \bibinfo{person}{Tieniu
  Tan}.} \bibinfo{year}{2014}\natexlab{}.
\newblock \showarticletitle{Iris Anti-spoofing}.
\newblock In \bibinfo{booktitle}{{\em Handbook of Biometric Anti-Spoofing:
  Trusted Biometrics under Spoofing Attacks}},
  \bibfield{editor}{\bibinfo{person}{S{\'e}bastien Marcel},
  \bibinfo{person}{Mark~S. Nixon}, {and} \bibinfo{person}{Stan~Z. Li}} (Eds.).
  \bibinfo{publisher}{Springer London}, \bibinfo{address}{London},
  \bibinfo{pages}{103--123}.
\newblock
\showISBNx{978-1-4471-6524-8}
\showDOI{%
\url{https://doi.org/10.1007/978-1-4471-6524-8_6}}


\bibitem[\protect\citeauthoryear{Sun, Zhang, Tan, and Wang}{Sun
  et~al\mbox{.}}{2014}]%
        {Sun_PAMI_2014}
\bibfield{author}{\bibinfo{person}{Zhenan Sun}, \bibinfo{person}{Hui Zhang},
  \bibinfo{person}{Tieniu Tan}, {and} \bibinfo{person}{Jianyu Wang}.}
  \bibinfo{year}{2014}\natexlab{}.
\newblock \showarticletitle{Iris Image Classification Based on Hierarchical
  Visual Codebook}.
\newblock \bibinfo{journal}{{\em {IEEE} Trans. Pattern Anal. Mach. Intell.\/}}
  \bibinfo{volume}{36}, \bibinfo{number}{6} (\bibinfo{date}{June}
  \bibinfo{year}{2014}), \bibinfo{pages}{1120--1133}.
\newblock
\showISSN{0162-8828}
\showDOI{%
\url{https://doi.org/10.1109/TPAMI.2013.234}}


\bibitem[\protect\citeauthoryear{Takano and Nakamura}{Takano and
  Nakamura}{2009}]%
        {Takano_SCE_2009}
\bibfield{author}{\bibinfo{person}{Hironobu Takano} {and}
  \bibinfo{person}{Kiyomi Nakamura}.} \bibinfo{year}{2009}\natexlab{}.
\newblock \showarticletitle{Rotation independent iris recognition by the
  rotation spreading neural network}. In \bibinfo{booktitle}{{\em {IEEE} Int.
  Symp. on Consumer Electronics}}. \bibinfo{publisher}{IEEE},
  \bibinfo{address}{Kyoto, Japan}, \bibinfo{pages}{651--654}.
\newblock
\showISSN{0747-668X}
\showDOI{%
\url{https://doi.org/10.1109/ISCE.2009.5156991}}


\bibitem[\protect\citeauthoryear{Thalheim, Krissler, and Ziegler}{Thalheim
  et~al\mbox{.}}{2002}]%
        {Thalheim_CT_2002}
\bibfield{author}{\bibinfo{person}{Lisa Thalheim}, \bibinfo{person}{Jan
  Krissler}, {and} \bibinfo{person}{Peter-Michael Ziegler}.}
  \bibinfo{year}{2002}\natexlab{}.
\newblock \bibinfo{title}{{Biometric Access Protection Devices and their
  Programs Put to the Test, Available online in c't Magazine, No. 11/2002, p.
  114}}.
\newblock \bibinfo{howpublished}{on-line}.   (\bibinfo{year}{2002}).
\newblock


\bibitem[\protect\citeauthoryear{Thavalengal and Corcoran}{Thavalengal and
  Corcoran}{2016}]%
        {Thavalengal_CEM_2016}
\bibfield{author}{\bibinfo{person}{Shejin Thavalengal} {and}
  \bibinfo{person}{Peter Corcoran}.} \bibinfo{year}{2016}\natexlab{}.
\newblock \showarticletitle{User Authentication on Smartphones: Focusing on
  iris biometrics}.
\newblock \bibinfo{journal}{{\em {IEEE} Consumer Electronics Magazine\/}}
  \bibinfo{volume}{5}, \bibinfo{number}{2} (\bibinfo{date}{April}
  \bibinfo{year}{2016}), \bibinfo{pages}{87--93}.
\newblock
\showISSN{2162-2248}
\showDOI{%
\url{https://doi.org/10.1109/MCE.2016.2522018}}


\bibitem[\protect\citeauthoryear{Thavalengal, Nedelcu, Bigioi, and
  Corcoran}{Thavalengal et~al\mbox{.}}{2016}]%
        {Thavalengal_TCE_2016}
\bibfield{author}{\bibinfo{person}{Shejin Thavalengal}, \bibinfo{person}{Tudor
  Nedelcu}, \bibinfo{person}{Petronel Bigioi}, {and} \bibinfo{person}{Peter
  Corcoran}.} \bibinfo{year}{2016}\natexlab{}.
\newblock \showarticletitle{Iris liveness detection for next generation
  smartphones}.
\newblock \bibinfo{journal}{{\em {IEEE} Trans. Cons. Elect.\/}}
  \bibinfo{volume}{62}, \bibinfo{number}{2} (\bibinfo{date}{May}
  \bibinfo{year}{2016}), \bibinfo{pages}{95--102}.
\newblock
\showISSN{0098-3063}
\showDOI{%
\url{https://doi.org/10.1109/TCE.2016.7514667}}


\bibitem[\protect\citeauthoryear{Thavalengal, Vranceanu, Condorovici, and
  Corcoran}{Thavalengal et~al\mbox{.}}{2014}]%
        {Thavalengal_IJCB_2014}
\bibfield{author}{\bibinfo{person}{Shejin Thavalengal},
  \bibinfo{person}{Ruxandra Vranceanu}, \bibinfo{person}{Razvan~G.
  Condorovici}, {and} \bibinfo{person}{Peter Corcoran}.}
  \bibinfo{year}{2014}\natexlab{}.
\newblock \showarticletitle{Iris pattern obfuscation in digital images}. In
  \bibinfo{booktitle}{{\em {IEEE} Int. Joint Conf. on Biometrics (IJCB)}}.
  \bibinfo{publisher}{IEEE}, \bibinfo{address}{Clearwater, FL, USA},
  \bibinfo{pages}{1--8}.
\newblock
\showDOI{%
\url{https://doi.org/10.1109/BTAS.2014.6996276}}


\bibitem[\protect\citeauthoryear{Tomeo-Reyes and Chandran}{Tomeo-Reyes and
  Chandran}{2013}]%
        {Tomeo-Reyes_ISIS_2013}
\bibfield{author}{\bibinfo{person}{Inmaculada Tomeo-Reyes} {and}
  \bibinfo{person}{Vinod Chandran}.} \bibinfo{year}{2013}\natexlab{}.
\newblock \showarticletitle{Iris based identity verification robust to sample
  presentation security attacks}.
\newblock \bibinfo{journal}{{\em Int. Journal of Information Science and
  Intelligent System\/}} \bibinfo{volume}{2}, \bibinfo{number}{1}
  (\bibinfo{date}{June} \bibinfo{year}{2013}), \bibinfo{pages}{27--41}.
\newblock
\showURL{%
\url{https://eprints.qut.edu.au/60842/}}


\bibitem[\protect\citeauthoryear{Toth and Galbally}{Toth and Galbally}{2015}]%
        {Toth_EB_2015}
\bibfield{author}{\bibinfo{person}{Anna~Bori Toth} {and}
  \bibinfo{person}{Javier Galbally}.} \bibinfo{year}{2015}\natexlab{}.
\newblock \showarticletitle{Anti-spoofing, Iris}.
\newblock In \bibinfo{booktitle}{{\em Encyclopedia of Biometrics}},
  \bibfield{editor}{\bibinfo{person}{Stan~Z. Li} {and} \bibinfo{person}{Anil
  Jain}} (Eds.). \bibinfo{publisher}{Springer US}, \bibinfo{address}{New York},
  \bibinfo{pages}{87--97}.
\newblock
\showISBNx{978-1-4899-7487-7}
\showURL{%
\url{http://www.springer.com/us/book/9781489974877}}


\bibitem[\protect\citeauthoryear{Toth}{Toth}{2009}]%
        {Toth_EB_2009}
\bibfield{author}{\bibinfo{person}{Bori Toth}.}
  \bibinfo{year}{2009}\natexlab{}.
\newblock \showarticletitle{Liveness Detection: Iris}.
\newblock In \bibinfo{booktitle}{{\em Encyclopedia of Biometrics}},
  \bibfield{editor}{\bibinfo{person}{Stan~Z. Li} {and} \bibinfo{person}{Anil
  Jain}} (Eds.). \bibinfo{publisher}{Springer US}, \bibinfo{address}{Boston,
  MA}, \bibinfo{pages}{931--938}.
\newblock
\showISBNx{978-0-387-73003-5}
\showDOI{%
\url{https://doi.org/10.1007/978-0-387-73003-5_179}}


\bibitem[\protect\citeauthoryear{Trokielewicz and Bartuzi}{Trokielewicz and
  Bartuzi}{2016}]%
        {Trokielewicz_JTIT_2016}
\bibfield{author}{\bibinfo{person}{Mateusz Trokielewicz} {and}
  \bibinfo{person}{Ewelina Bartuzi}.} \bibinfo{year}{2016}\natexlab{}.
\newblock \showarticletitle{Cross-spectral Iris Recognition for Mobile
  Applications using High-quality Color Images}.
\newblock \bibinfo{journal}{{\em Journal of Telecommunications and Information
  Technology\/}}  \bibinfo{volume}{3} (\bibinfo{year}{2016}),
  \bibinfo{pages}{91--97}.
\newblock


\bibitem[\protect\citeauthoryear{Trokielewicz, Czajka, and
  Maciejewicz}{Trokielewicz et~al\mbox{.}}{2016a}]%
        {Trokielewicz_BTAS_2016}
\bibfield{author}{\bibinfo{person}{Mateusz Trokielewicz}, \bibinfo{person}{Adam
  Czajka}, {and} \bibinfo{person}{Piotr Maciejewicz}.}
  \bibinfo{year}{2016}\natexlab{a}.
\newblock \showarticletitle{Human iris recognition in post-mortem subjects:
  Study and database}. In \bibinfo{booktitle}{{\em {IEEE} Int. Conf. on
  Biometrics: Theory Applications and Systems (BTAS)}}.
  \bibinfo{publisher}{IEEE}, \bibinfo{address}{Niagara Falls, NY, USA},
  \bibinfo{pages}{1--6}.
\newblock
\showDOI{%
\url{https://doi.org/10.1109/BTAS.2016.7791175}}


\bibitem[\protect\citeauthoryear{Trokielewicz, Czajka, and
  Maciejewicz}{Trokielewicz et~al\mbox{.}}{2016b}]%
        {Trokielewicz_ICB_2016}
\bibfield{author}{\bibinfo{person}{Mateusz Trokielewicz}, \bibinfo{person}{Adam
  Czajka}, {and} \bibinfo{person}{Piotr Maciejewicz}.}
  \bibinfo{year}{2016}\natexlab{b}.
\newblock \showarticletitle{Post-mortem human iris recognition}. In
  \bibinfo{booktitle}{{\em {IEEE} Int. Conf. on Biometrics (ICB)}}.
  \bibinfo{publisher}{IEEE}, \bibinfo{address}{Halmstad, Sweden},
  \bibinfo{pages}{1--6}.
\newblock
\showDOI{%
\url{https://doi.org/10.1109/ICB.2016.7550073}}


\bibitem[\protect\citeauthoryear{Venugopalan and Savvides}{Venugopalan and
  Savvides}{2011}]%
        {Venugopalan_TIFS_2011}
\bibfield{author}{\bibinfo{person}{Shreyas Venugopalan} {and}
  \bibinfo{person}{Marios Savvides}.} \bibinfo{year}{2011}\natexlab{}.
\newblock \showarticletitle{How to Generate Spoofed Irises From an Iris Code
  Template}.
\newblock \bibinfo{journal}{{\em {IEEE} Trans. Inf. Forens. Security\/}}
  \bibinfo{volume}{6}, \bibinfo{number}{2} (\bibinfo{date}{June}
  \bibinfo{year}{2011}), \bibinfo{pages}{385--395}.
\newblock
\showISSN{1556-6013}
\showDOI{%
\url{https://doi.org/10.1109/TIFS.2011.2108288}}


\bibitem[\protect\citeauthoryear{Villalbos-Castaldi and
  Suaste-G\'{o}mez}{Villalbos-Castaldi and Suaste-G\'{o}mez}{2014}]%
        {Villalbos-Castaldi_BF_2014}
\bibfield{author}{\bibinfo{person}{Fabiola~M. Villalbos-Castaldi} {and}
  \bibinfo{person}{Ernesto Suaste-G\'{o}mez}.} \bibinfo{year}{2014}\natexlab{}.
\newblock \showarticletitle{In the use of the spontaneous pupillary
  oscillations as a new biometric trait}. In \bibinfo{booktitle}{{\em Int.
  Workshop on Biometrics and Forensics}}. \bibinfo{publisher}{IEEE},
  \bibinfo{address}{Valletta, Malta}, \bibinfo{pages}{1--6}.
\newblock
\showDOI{%
\url{https://doi.org/10.1109/IWBF.2014.6914259}}


\bibitem[\protect\citeauthoryear{{Warsaw University of Technology}}{{Warsaw
  University of Technology}}{2013}]%
        {WARSAW_DBs_URL}
\bibfield{author}{\bibinfo{person}{{Warsaw University of Technology}}.}
  \bibinfo{year}{2013}\natexlab{}.
\newblock \bibinfo{title}{{Warsaw Datasets Webpage}}.
\newblock   (\bibinfo{year}{2013}).
\newblock
\showURL{%
Retrieved August 10, 2017 from \url{http://zbum.ia.pw.edu.pl/EN/node/46}}


\bibitem[\protect\citeauthoryear{Wei, Chen, and Ferryman}{Wei
  et~al\mbox{.}}{2013}]%
        {Wei_FRONTEX_2013}
\bibfield{author}{\bibinfo{person}{Hong Wei}, \bibinfo{person}{Lulu Chen},
  {and} \bibinfo{person}{James Ferryman}.} \bibinfo{year}{2013}\natexlab{}.
\newblock \showarticletitle{Biometrics in ABC: counter-spoofing research}. In
  \bibinfo{booktitle}{{\em FRONTEX 2nd Global Conf. on Future Developments of
  Automated Border Control}}. \bibinfo{publisher}{FRONTEX},
  \bibinfo{address}{Warsaw, Poland}, \bibinfo{pages}{1--4}.
\newblock
\showISBNx{978-92-95033-76-4}
\showDOI{%
\url{https://doi.org/10.2819/20688}}


\bibitem[\protect\citeauthoryear{Wei, Qiu, Sun, and Tan}{Wei
  et~al\mbox{.}}{2008a}]%
        {Wei_CPR_2008}
\bibfield{author}{\bibinfo{person}{Zhuoshi Wei}, \bibinfo{person}{Xianchao
  Qiu}, \bibinfo{person}{Zhenan Sun}, {and} \bibinfo{person}{Tieniu Tan}.}
  \bibinfo{year}{2008}\natexlab{a}.
\newblock \showarticletitle{Counterfeit iris detection based on texture
  analysis}. In \bibinfo{booktitle}{{\em Int. Conf. on Pattern Recognition
  (ICPR)}}. \bibinfo{publisher}{IEEE}, \bibinfo{address}{Tampa, FL, USA},
  \bibinfo{pages}{1--4}.
\newblock
\showISSN{1051-4651}
\showDOI{%
\url{https://doi.org/10.1109/ICPR.2008.4761673}}


\bibitem[\protect\citeauthoryear{Wei, Tan, and Sun}{Wei et~al\mbox{.}}{2008b}]%
        {Wei_ICPR_2008}
\bibfield{author}{\bibinfo{person}{Zhuoshi Wei}, \bibinfo{person}{Tieniu Tan},
  {and} \bibinfo{person}{Zhenan Sun}.} \bibinfo{year}{2008}\natexlab{b}.
\newblock \showarticletitle{Synthesis of large realistic iris databases using
  patch-based sampling}. In \bibinfo{booktitle}{{\em Int. Conf. on Pattern
  Recognition (ICPR)}}. \bibinfo{publisher}{IEEE}, \bibinfo{address}{Tampa, FL,
  USA}, \bibinfo{pages}{1--4}.
\newblock
\showISSN{1051-4651}
\showDOI{%
\url{https://doi.org/10.1109/ICPR.2008.4761674}}


\bibitem[\protect\citeauthoryear{Yadav, Kohli, Doyle, Singh, Vatsa, and
  Bowyer}{Yadav et~al\mbox{.}}{2014}]%
        {Yadav_TIFS_2014}
\bibfield{author}{\bibinfo{person}{Daksha Yadav}, \bibinfo{person}{Naman
  Kohli}, \bibinfo{person}{James~S. Doyle}, \bibinfo{person}{Richa Singh},
  \bibinfo{person}{Mayank Vatsa}, {and} \bibinfo{person}{Kevin~W. Bowyer}.}
  \bibinfo{year}{2014}\natexlab{}.
\newblock \showarticletitle{Unraveling the Effect of Textured Contact Lenses on
  Iris Recognition}.
\newblock \bibinfo{journal}{{\em {IEEE} Trans. Inf. Forens. Security\/}}
  \bibinfo{volume}{9}, \bibinfo{number}{5} (\bibinfo{date}{May}
  \bibinfo{year}{2014}), \bibinfo{pages}{851--862}.
\newblock
\showISSN{1556-6013}
\showDOI{%
\url{https://doi.org/10.1109/TIFS.2014.2313025}}


\bibitem[\protect\citeauthoryear{Yadav, Kohli, Vatsa, Singh, and Noore}{Yadav
  et~al\mbox{.}}{2017}]%
        {Yadav_IJCB_2017}
\bibfield{author}{\bibinfo{person}{Daksha Yadav}, \bibinfo{person}{Naman
  Kohli}, \bibinfo{person}{Mayank Vatsa}, \bibinfo{person}{Richa Singh}, {and}
  \bibinfo{person}{Afzel Noore}.} \bibinfo{year}{2017}\natexlab{}.
\newblock \showarticletitle{Unconstrained Visible Spectrum Iris with Textured
  Contact Lens Variations: Database and Benchmarking}. In
  \bibinfo{booktitle}{{\em {IEEE} Int. Joint Conf. on Biometrics (IJCB)}}.
  \bibinfo{publisher}{IEEE}, \bibinfo{address}{Denver, CO, USA},
  \bibinfo{pages}{1--6}.
\newblock


\bibitem[\protect\citeauthoryear{Yambay, Becker, Kohli, Yadav, Czajka, Bowyer,
  Schuckers, Singh, Vatsa, Noore, Gragnaniello, Sansone, Verdoliva, He, Ru, Li,
  Liu, Sun, and Tan}{Yambay et~al\mbox{.}}{2017}]%
        {Yambay_IJCB_2017}
\bibfield{author}{\bibinfo{person}{David Yambay}, \bibinfo{person}{Benedict
  Becker}, \bibinfo{person}{Naman Kohli}, \bibinfo{person}{Daksha Yadav},
  \bibinfo{person}{Adam Czajka}, \bibinfo{person}{Kevin~W. Bowyer},
  \bibinfo{person}{Stephanie Schuckers}, \bibinfo{person}{Richa Singh},
  \bibinfo{person}{Mayank Vatsa}, \bibinfo{person}{Afzel Noore},
  \bibinfo{person}{Diego Gragnaniello}, \bibinfo{person}{C. Sansone},
  \bibinfo{person}{L. Verdoliva}, \bibinfo{person}{Lingxiao He},
  \bibinfo{person}{Yiwei Ru}, \bibinfo{person}{Haiqing Li},
  \bibinfo{person}{Nianfeng Liu}, \bibinfo{person}{Zhenan Sun}, {and}
  \bibinfo{person}{Tieniu Tan}.} \bibinfo{year}{2017}\natexlab{}.
\newblock \showarticletitle{{LivDet Iris 2017} -- Iris Liveness Detection
  Competition 2017}. In \bibinfo{booktitle}{{\em {IEEE} Int. Joint Conf. on
  Biometrics (IJCB)}}. \bibinfo{publisher}{IEEE}, \bibinfo{address}{Denver, CO,
  USA}, \bibinfo{pages}{1--6}.
\newblock


\bibitem[\protect\citeauthoryear{Yambay, Doyle, Bowyer, Czajka, and
  Schuckers}{Yambay et~al\mbox{.}}{2014}]%
        {Yambay_IJCB_2014}
\bibfield{author}{\bibinfo{person}{David Yambay}, \bibinfo{person}{James~S.
  Doyle}, \bibinfo{person}{Kevin~W. Bowyer}, \bibinfo{person}{Adam Czajka},
  {and} \bibinfo{person}{Stephanie Schuckers}.}
  \bibinfo{year}{2014}\natexlab{}.
\newblock \showarticletitle{LivDet-iris 2013 - Iris Liveness Detection
  Competition 2013}. In \bibinfo{booktitle}{{\em {IEEE} Int. Joint Conf. on
  Biometrics (IJCB)}}. \bibinfo{publisher}{IEEE}, \bibinfo{address}{Clearwater,
  FL, USA}, \bibinfo{pages}{1--8}.
\newblock
\showDOI{%
\url{https://doi.org/10.1109/BTAS.2014.6996283}}


\bibitem[\protect\citeauthoryear{Yambay and Schuckers}{Yambay and
  Schuckers}{2018}]%
        {Yambay_PAD_Handbook_2018}
\bibfield{author}{\bibinfo{person}{David Yambay} {and}
  \bibinfo{person}{Stephanie Schuckers}.} \bibinfo{year}{2018}\natexlab{}.
\newblock \showarticletitle{{Review of Iris Presentation Attack Detection
  Competitions}}.
\newblock In \bibinfo{booktitle}{{\em Handbook of Biometric Anti-Spoofing (2nd
  Edition, to appear)}}, \bibfield{editor}{\bibinfo{person}{S\'{e}bastien
  Marcel}, \bibinfo{person}{Mark Nixon}, \bibinfo{person}{Julian Fierrez},
  {and} \bibinfo{person}{Nicholas Evans}} (Eds.). \bibinfo{publisher}{Springer
  Int. Publishing AG}, \bibinfo{pages}{1--16}.
\newblock


\bibitem[\protect\citeauthoryear{Yambay, Walczak, Schuckers, and Czajka}{Yambay
  et~al\mbox{.}}{2017}]%
        {Yambay_ISBA_2017}
\bibfield{author}{\bibinfo{person}{David Yambay}, \bibinfo{person}{Brian
  Walczak}, \bibinfo{person}{Stephanie Schuckers}, {and} \bibinfo{person}{Adam
  Czajka}.} \bibinfo{year}{2017}\natexlab{}.
\newblock \showarticletitle{LivDet-Iris 2015 - Iris Liveness Detection
  Competition 2015}. In \bibinfo{booktitle}{{\em {IEEE} Int. Conf. on Identity,
  Security and Behavior Analysis (ISBA)}}. \bibinfo{publisher}{IEEE},
  \bibinfo{address}{New Delhi, India}, \bibinfo{pages}{1--6}.
\newblock
\showDOI{%
\url{https://doi.org/10.1109/ISBA.2017.7947701}}


\bibitem[\protect\citeauthoryear{Zhang, Sun, and Tan}{Zhang
  et~al\mbox{.}}{2010}]%
        {Zhang_CPR_2010}
\bibfield{author}{\bibinfo{person}{Hui Zhang}, \bibinfo{person}{Zhenan Sun},
  {and} \bibinfo{person}{Tieniu Tan}.} \bibinfo{year}{2010}\natexlab{}.
\newblock \showarticletitle{Contact Lens Detection Based on Weighted LBP}. In
  \bibinfo{booktitle}{{\em Int. Conf. on Pattern Recognition (ICPR)}}.
  \bibinfo{publisher}{IEEE}, \bibinfo{address}{Istanbul, Turkey},
  \bibinfo{pages}{4279--4282}.
\newblock
\showISSN{1051-4651}
\showDOI{%
\url{https://doi.org/10.1109/ICPR.2010.1040}}


\bibitem[\protect\citeauthoryear{Zuo, Schmid, and Chen}{Zuo
  et~al\mbox{.}}{2007}]%
        {Zuo_TIFS_2007}
\bibfield{author}{\bibinfo{person}{Jinyu Zuo}, \bibinfo{person}{Natalia~A.
  Schmid}, {and} \bibinfo{person}{Xiaohan Chen}.}
  \bibinfo{year}{2007}\natexlab{}.
\newblock \showarticletitle{On Generation and Analysis of Synthetic Iris
  Images}.
\newblock \bibinfo{journal}{{\em {IEEE} Trans. Inf. Forens. Security\/}}
  \bibinfo{volume}{2}, \bibinfo{number}{1} (\bibinfo{date}{March}
  \bibinfo{year}{2007}), \bibinfo{pages}{77--90}.
\newblock
\showISSN{1556-6013}
\showDOI{%
\url{https://doi.org/10.1109/TIFS.2006.890305}}


\end{thebibliography}

\end{document}